\RequirePackage{fix-cm}
\documentclass[smallextended]{svjour3}       

\smartqed 
\usepackage{amsmath}
\usepackage{amssymb}
\usepackage{mathtools}
\usepackage{bm}  
\usepackage{graphicx}    
\usepackage{hyperref} 
\usepackage[numbers,sort&compress]{natbib}  
\usepackage{tabularx}
\usepackage{subcaption}
\begin{document}

\title{Screw-Theoretic PDE-Based Synthesis for Unified Representation of Multibody Flexible Manipulator Dynamics} 
\titlerunning{Screw-theoretic PDE synthesis...}

\subtitle{}

\author{S.~Yaqubi${}^*$ \and A.~Kitzinger \and A.~Müller \and H.~Gattringer \and J.~Mattila}

\institute{
S.~Yaqubi \at
Department of Automation Technology and Mechanical Engineering, Tampere University, Korkeakoulunkatu 6, 33720 Tampere, Finland \\
\email{sadeq.yaqubi@tuni.fi} \\
ORCID: \url{https://orcid.org/0000-0002-7093-5747}
\and
A.~Kitzinger \at
Institute of Robotics, Johannes Kepler University Linz,
4040~Linz, Austria \\
\email{alexander.kitzinger@jku.at} \\
ORCID: \url{https://orcid.org/0009-0007-4459-0654}
\and
A.~Müller \at
Institute of Robotics, Johannes Kepler University Linz,
4040~Linz, Austria \\
\email{a.mueller@jku.at} \\
ORCID: \url{https://orcid.org/0000-0003-2127-7335}
\and
H.~Gattringer \at
Institute of Robotics, Johannes Kepler University Linz,
4040~Linz, Austria \\
\email{hubert.gattringer@jku.at} \\
ORCID: \url{https://orcid.org/0000-0002-8846-9051}
\and
J. Mattila \at
Department of Automation Technology and Mechanical Engineering, Tampere University, Korkeakoulunkatu 6, 33720 Tampere, Finland \\
\email{jouni.mattila@tuni.fi} \\
ORCID: \url{https://orcid.org/0000-0003-1799-4323}
\and
Manuscript submitted to Springer for peer review. Copyright might be transferred at any moment without notice.
}

\maketitle
\markboth{}{ }

\noindent
${}^*$corresponding author.

\date{Received: date / Accepted: date}

\maketitle

\abstract{This paper addresses the open problem of partial differential equation (PDE)-based 
dynamic modeling for flexible multibody robotic systems by presenting a screw-theoretic synthesis methodology---developed within a unified Lie-algebraic framework---for serial flexible manipulators with an arbitrary number of links in three-dimensional motion. The proposed approach expresses all dynamic states — rigid-body motion, elastic deformation, and inter-link interaction forces — uniformly within the screw theoretic structure as body-fixed twists and their dual wrenches, treating all physical components consistently within the same geometric framework. Hence, the PDE structure of the deformation field is retained exactly, while the $\mathfrak{se}(3)$ representation admits the linear-algebraic formulation required for multibody assembly and formal well-posedness analysis. Building on a previously developed single-link screw-theoretic PDE model, joint constraints are enforced as screw-compatibility equations connecting the twist and wrench fields of adjacent links at their connection points, and the per-link models are assembled via a closed-form linear-algebraic stacking procedure. The resulting system matrix exhibits near-tridiagonal block structure, interaction wrenches appear explicitly as algebraic variables, and adding a link requires only appending block rows — making the synthesis automatable for arbitrary $n$. The assembled system is formulated as a semi-explicit index-1 differential-algebraic equation, and well-posedness is established by recasting it in abstract Cauchy form through modal projection. Presented solutions are validated experimentally on a two-link flexible manipulator in three-dimensional motion, confirming implementability and physical consistency.}



\begin{center}
  \textbf{Graphical Abstract} \\[1em]
  \includegraphics[width=\textwidth]{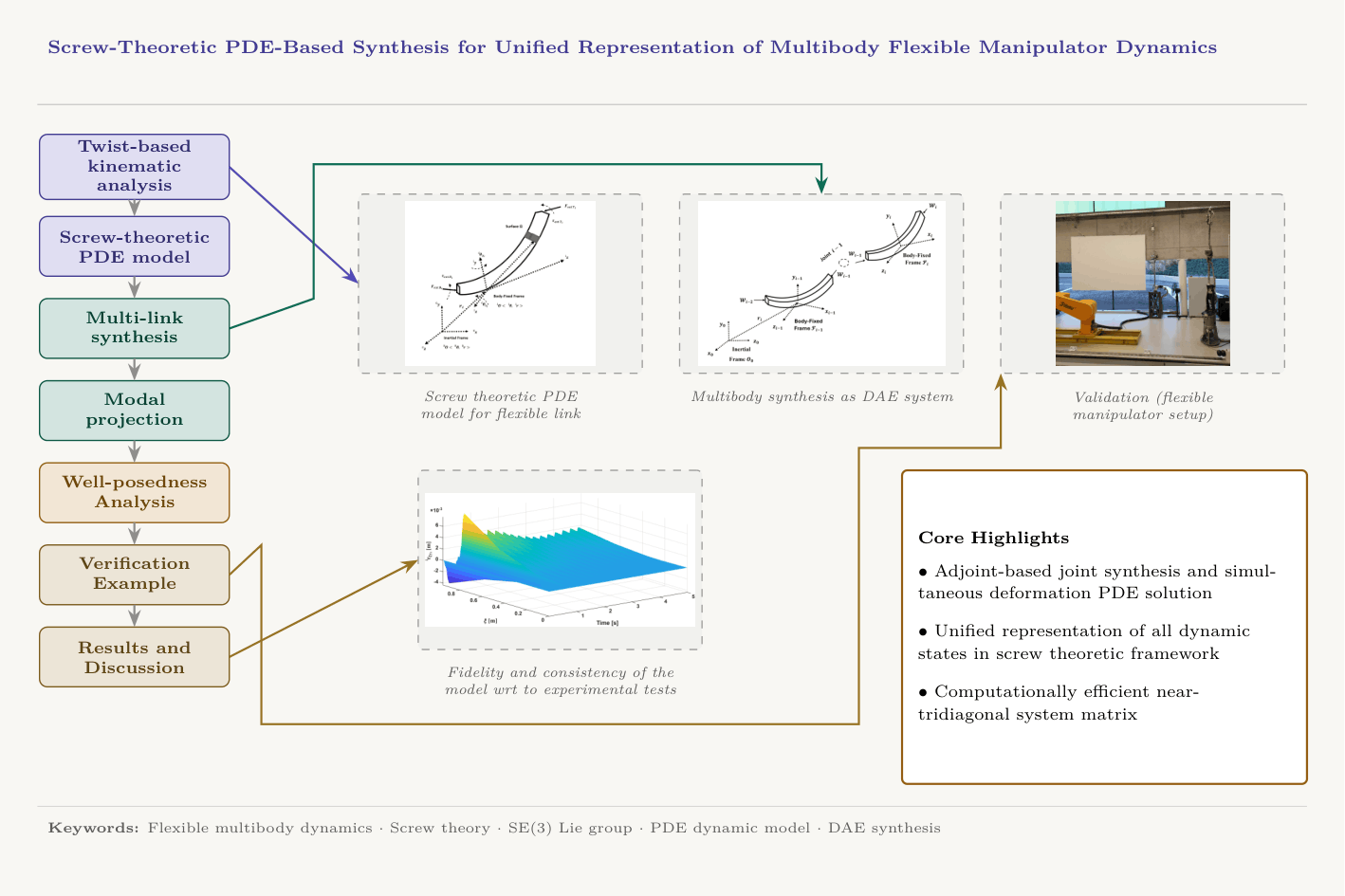}
\end{center}

\section{Introduction} \label{sec:1}

Flexible multibody robotic systems require modeling frameworks that capture two coupled phenomena: finite rigid-body motion of each link's frame, evolving on the Lie group $\mathrm{SE}(3)$, and distributed elastic deformation along each link, governed by an infinite-dimensional partial differential equation (PDE)~\cite{Boyer2021,Xu2023}. Existing approaches resolve this coupling either by discretizing the deformation into a finite-dimensional ordinary differential equation (ODE) or by linearizing the rigid-body motion~\cite{Walsh2015}: ODE approximations introduce truncation error whose effect on the true infinite-dimensional system cannot be rigorously bounded, while linearization restricts validity to small-motion regimes~\cite{Rao2007}. The present work pursues a third path: retaining the PDE structure exactly, expressing the nonlinear rigid-body dynamics via screw theory on $\mathfrak{se}(3)$, and establishing formal well-posedness of the resulting system.

For rigid multibody systems, it is well-established that projecting from $\mathrm{SE}(3)$ to its Lie algebra $\mathfrak{se}(3)$ via screw-theoretic twist and wrench coordinates enables modular composition of link dynamics~\cite{Featherstone2008,Mller2018,Zhu2010}: inter-link coupling reduces to linear Adjoint transformations, and the $n$-link forward dynamic map assembles algebraically rather than through nested nonlinear compositions. Extending this to flexible bodies is nontrivial: the deformation field must be represented as a continuous screw field along the body~\cite{Gao2024,Cibicik2021}, and its coupling to the rigid-body configuration means the forward dynamic map takes the nested form $\mathbf{f}_1(\mathbf{f}_2(\cdots(\mathbf{f}_n(\cdot))\cdots))$, where each $\mathbf{f}_i$ incorporates gravitational terms, trigonometric position dependence, generalized forces, and shape-function integrals~\cite{Li2024,Gao2023} — hindering algebraic assembly, automation for arbitrary $n$, and the linear operator structure required for well-posedness analysis. Expressing link velocities as body-fixed twists and internal forces as wrenches in $\mathfrak{se}(3)$ and $\mathfrak{se}^*(3)$ restores this structure: velocity-dependent generalized forces emerge from the adjoint action~\cite{Featherstone2008,Mller2018,Zhang2024}, the deformation field becomes a continuous screw field in body-fixed coordinates~\cite{Yaqubi2026,Cibicik2021}, and inter-link coupling reduces to structured Adjoint transformations~\cite{Featherstone2008,Zhu2010}.

Several well-established approaches exist for modeling multibody flexible robots. The Floating Frame of Reference (FFR) method~\cite{Shabana2013,Sugiyama2006} decomposes motion into a rigid reference frame and relative elastic deformation, with shape functions from modal analysis of boundary conditions, yielding a finite-dimensional ODE. The Absolute Nodal Coordinate Formulation~\cite{Shabana2015} and Geometrically Exact Beam Formulation~\cite{Trivedi2008,Demoures2015} — including recent Lie group variational integrator formulations based on Cosserat rod theory with strain-parameterized potential energy~\cite{Boyer2021,Herrmann2024,Chen2022} — capture large deformations with high geometric fidelity, but working directly on the configuration manifold complicates the governing equations and derivation of closed-form modal representations. Co-Rotational FEM~\cite{Wu2024} (COFEM) additionally relies on finite-element solutions, introducing computational demands that limit real-time applicability~\cite{Homaeinezhad2020}. Screw-theoretic extensions — including dual screw models with linearized elastic motion~\cite{Cibicik2021}, finite-element screw-based formulations~\cite{Grazioso2019}, Lie group formulation for geometrically exact beam~\cite{Herrmann2024}, and Lie algebra integration scheme for flexible multibody dynamics on $\mathrm{SE}(3)$~\cite{Chen2022} — demonstrate the potential of screw-theoretic representations, though fully exploiting this for PDE-based multibody synthesis remains an open research direction.

Despite these advances, retaining PDE structure within a Lie-algebraic framework has not been fully pursued. In such framework, the governing PDE would constitute a faithful representation of the true infinite-dimensional configuration space~\cite{Gao2024,Zhang2005}, benefiting applications such as high-fidelity simulation and model-based control synthesis~\cite{Xi2025}. Scaling such models to multibody systems is fundamentally harder than scaling ODE models: formulations on the nonlinear configuration manifold lack the global linear structure needed for analytical shape function derivation and algebraic multibody assembly~\cite{Trivedi2008,Shabana2015,Demoures2015}, whereas a Lie-algebraic representation naturally provides this: the linear-algebraic structure enable analytic separation-of-variables discretization, supports analytical boundary-condition-based shape functions, and renders well-posedness analysis tractable without linearization.

To bridge the gap between PDE representation and Lie algebra formulations, the present work develops a screw-theoretic PDE-based dynamic model for flexible multibody robotic systems, building on a previously developed single-link model~\cite{Yaqubi2026}. The following contributions are made. First, the multibody system is represented uniformly in $\mathfrak{se}(3)$: deformation states as continuous body-fixed twist fields, rigid-body motion as body-fixed twists, and joint constraints as screw-compatibility equations connecting the twist and wrench fields of adjacent links — a unified algebraic representation geometrically consistent with the $\mathrm{SE}(3)$ configuration structure, eliminating the representational heterogeneity of conventional approaches. Second, the per-link PDE models are assembled via a closed-form stacking procedure in which interaction wrenches appear explicitly as algebraic variables, the system matrix exhibits near-tridiagonal block structure reflecting physical chain connectivity, and adding a link requires only appending block rows — making synthesis automatable for arbitrary $n$ based on the arrangement of differential-algebraic equations (DAE). Third, the screw theoretic synthesis model is extended as a modal projection-based well-posedness analysis: the projected system admits a finite-dimensional Cauchy form whose regularity follows directly from the validity of energy terms, yielding formal guarantees of local existence, uniqueness, and continuous dependence on initial conditions — to the authors' knowledge the first such result for a flexible multibody PDE system of this generality. The model is validated experimentally on a two-link flexible manipulator, confirming implementability and physical consistency.

The rest of this paper is organized as follows. Section~\ref{sec:2} presents the problem statement, notation, and geometric definitions. The screw-theoretic PDE-based dynamic model for a single flexible link is reviewed in Section~\ref{sec:3}. The synthesis framework is developed in Section~\ref{sec:4}, constituting the core of the work. Separation-of-variables-based modal projection is investigated in Section~\ref{sec:5}, which enables the well-posedness analysis in Section~\ref{sec:6}. Numerical simulation and experimental validation are presented in Section~\ref{sec:7}.

\section{Problem statement} \label{sec:2}

\begin{figure}[!t]
\centering
\includegraphics[width=3.5in]{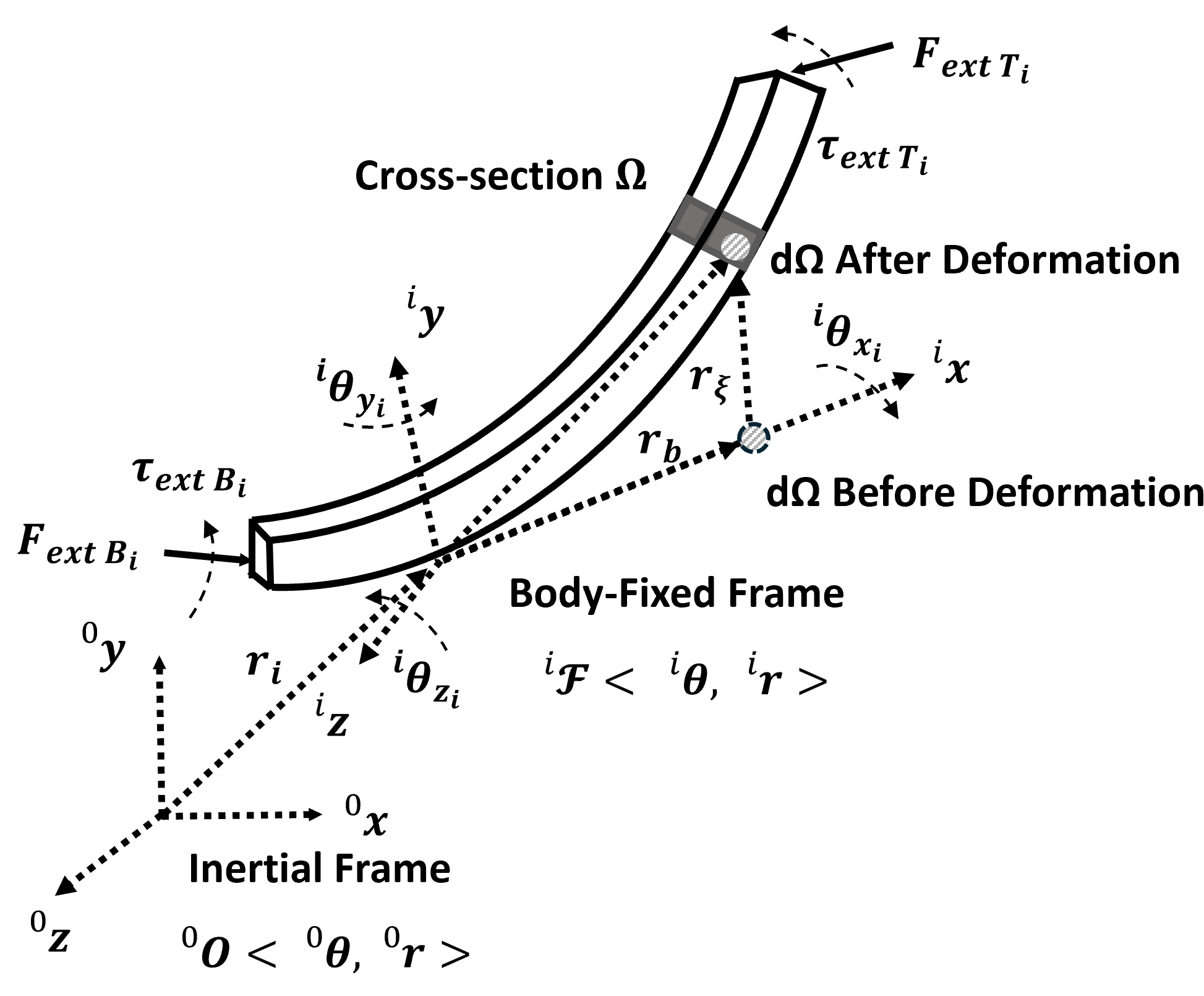}
\caption{Dynamic motion of flexible body with respect to inertial frame and body-fixed frame.}
\label{fig_1}
\end{figure}

\begin{figure*}[!t]
\centering
\includegraphics[width=\textwidth]{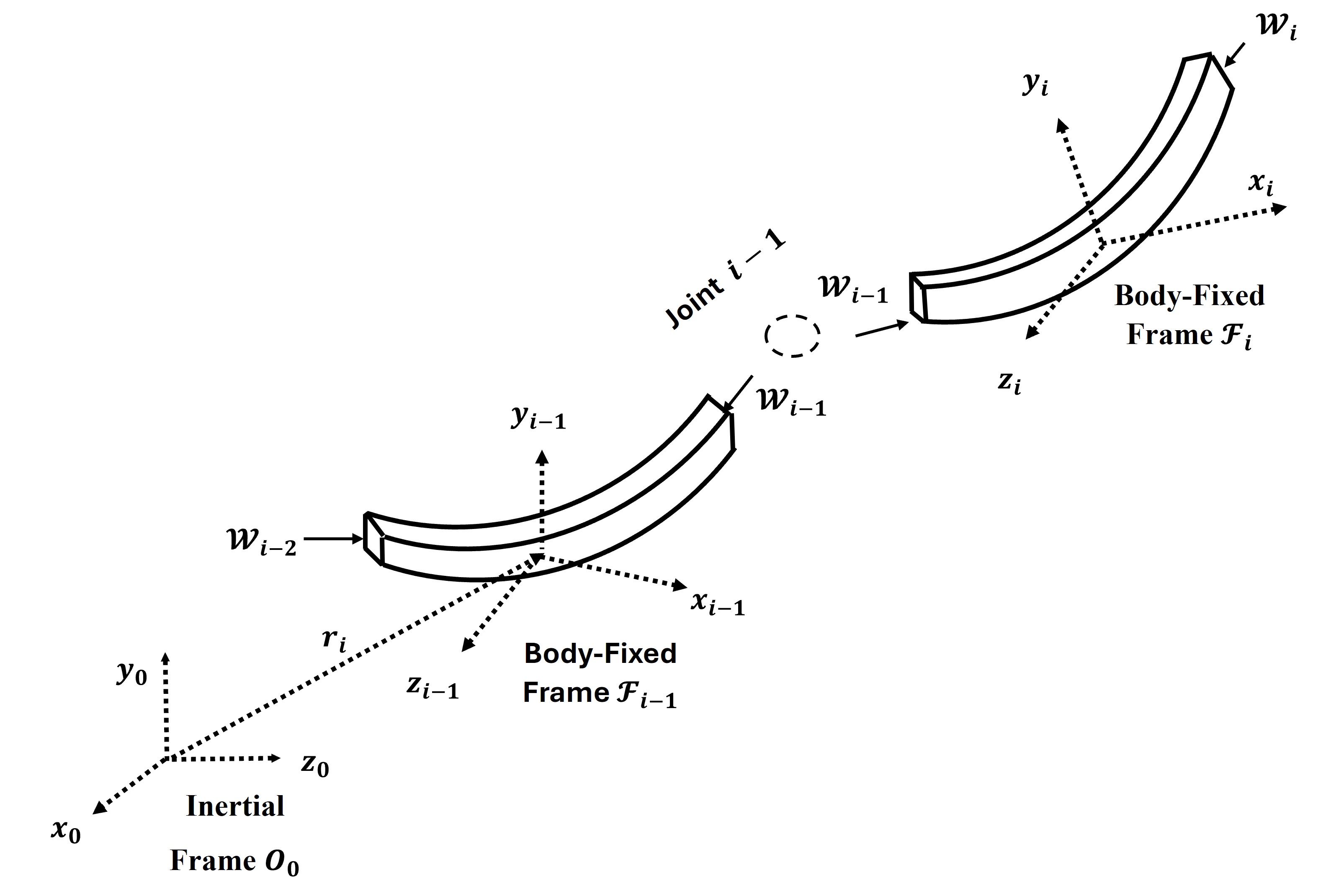}
\caption{Schematic of decomposed consecutive links in a multi-link flexible robot with holonomic constraints based on rotational joint connection.}
\label{fig_2}
\end{figure*}

This work considers the general dynamics of a multi-link serial flexible manipulator operating in three-dimensional space, where each link may be actuated by forces and torques applied at its endpoints, and is connected to the adjacent links via rotational joints. Such a formulation is particularly relevant for robotic systems, as it allows each link to be modeled individually while accounting for actuation inputs and interaction forces. The motion of a single flexible link is illustrated in Fig.~\ref{fig_1}, and the configuration of a multi-link robot with $n$ flexible links, including the decomposed joint connections between links, is shown in Fig.~\ref{fig_2}.

The present work builds on a previously developed screw-theoretic model for a single flexible link~\cite{Yaqubi2026}, shifting the focus from individual link modeling to systematic synthesis for multibody systems. The original model was derived via variational analysis of the link's Hamiltonian, accounting for the interaction between rigid-body motion and deformation by representing the flexible body as a continuum of infinitesimal elements. The motion of each element was expressed using a set of dual screws: one describing the motion of the body-fixed frame relative to the inertial frame, a second connecting the body-fixed frame to the undeformed geometry of the elements, and a third capturing the deformation of each element due to flexibility. All screws are represented in a rotating body-fixed frame, which may be attached to any point at the centerline of the flexible body. In this representation, the first and third screws are time-varying and require dynamic evaluation, while the second screw is time-invariant in the body-fixed frame and encodes the reference geometry of the link prior to deformation. The deformation of the body is modeled based on Euler--Bernoulli beam theory~\cite{Rao2007} under the assumptions of small deformation, isotropic material, and negligible shear deformation with additional consideration of longitudinal displacement. Altering the deformation model for more comprehensive deformation theory is permissible within the general screw theoretic kinematic and dynamic framework. 

Each flexible link is modeled as a collection of infinitesimal point 
elements $d\Omega_i$, each located within an infinitesimal cross-section 
element $\Omega_i$ of the beam. The cross-section is indexed by the 
spatial coordinate $\xi_i \in [a_i, c_i]$ measured along the beam axis 
in the body-fixed frame ${}^i\!\mathbf{\mathcal{F}}$, where $a_i$ and 
$c_i$ denote the distances of the beam endpoints from the body-fixed 
frame origin in the undeformed configuration, and $l_i = c_i - a_i$ is 
the beam length. The cross-sectional area is $A_i$, the mass density 
$\rho_i$, and the elastic modulus $E_i$.

The generalized coordinate vector ${}^{i}\mathbf{q}_i$, generalized velocity
${}^{i}\dot{\mathbf{q}}_i$, and the body-fixed twist ${}^{i}\mathbf{V}_i$, 
are introduced as

\begin{align}
{}^{i}\mathbf{q}_i &= 
\begin{bmatrix}
{}^{i}\boldsymbol{\theta}_i \\
{}^{i}\mathbf{r}_i
\end{bmatrix}
\label{eq:1} \\
{}^{i}\dot{\mathbf{q}}_i &= 
\begin{bmatrix}
{}^{i}\dot{\boldsymbol{\theta}}_i \\
{}^{i}\dot{\mathbf{r}}_i
\end{bmatrix}
\label{eq:2} \\
{}^{i}\mathbf{V}_i &= 
\begin{bmatrix}
{}^{i}\boldsymbol{\omega}_i \\
{}^{i}\mathbf{v}_i
\end{bmatrix}
= 
\begin{bmatrix}
{}^{i}\boldsymbol{\omega}_i \\
{}^{i}\dot{\mathbf{r}}_i + {}^{i}\tilde{\boldsymbol{\omega}}_i\,{}^{i}\mathbf{r}_i
\end{bmatrix}
\in \mathfrak{se}(3)
\label{eq:twist}
\end{align}


The generalized coordinate vector ${}^{i}\mathbf{q}_i = 
[{}^{i}\boldsymbol{\theta}_i^\top,\, {}^{i}\mathbf{r}_i^\top]^\top$ 
is a local parameterization of the Lie group $SE(3)$, where 
${}^{i}\mathbf{r}_i \in \mathbb{R}^3$ is the body-fixed origin 
position and ${}^{i}\boldsymbol{\theta}_i \in \mathbb{R}^3$ 
is a body-fixed angle parameterization of $SO(3)$. The rotation component 
${}^{i}\boldsymbol{\theta}_i$ parametrizes an element of $SO(3)$ via the 
exponential map $\mathbf{R}_{oi} = \exp({}^{i}\tilde{\boldsymbol{\theta}}_i) 
\in SO(3)$, where ${}^{0}\mathbf{r}_i = \mathbf{R}_{oi}\,{}^{i}\mathbf{r}_i$ is the position of the body-fixed frame expressed in the inertial frame.  
The angular velocity $\boldsymbol{\omega}_i$ relates to the coordinate 
derivative $\dot{\boldsymbol{\theta}}_i$ via 
$\boldsymbol{\omega}_i = \mathbf{J}(\boldsymbol{\theta}_i)\dot{\boldsymbol{\theta}}_i$, 
where $\mathbf{J}(\boldsymbol{\theta}_i)$ is the body-frame Jacobian of the 
exponential map~\cite{murray1994}. The skew-symmetric 
matrix corresponding to $\boldsymbol{\omega} \in \mathbb{R}^3$~\cite{murray1994} is expressed by $\tilde{\boldsymbol{\omega}} \in \mathfrak{so}(3)$.

The direction cosine matrix $\mathbf{R}_{oi}$ transforms vectors from 
body-fixed frame $\mathbf{\mathcal{F}}_i$ to inertial frame $o$, such that 
${}^0\mathbf{a} = \mathbf{R}_{oi}\,{}^i\mathbf{a}$ for any vector $\mathbf{a}$, 
and chains as $\mathbf{R}_{oi} = \mathbf{R}_{o(i-1)}\,\mathbf{R}_{(i-1)i}$. The twist transforms via the Adjoint representation 
${}^{0}\mathbf{V}_i = \mathbf{Ad}_{oi}\,{}^{i}\mathbf{V}_i \in 
\mathfrak{se}(3)$~\cite{murray1994}. For any two frames $a$ and $b$, the 
Adjoint $\mathbf{Ad}_{ab} \in \mathbb{R}^{6\times6}$ is defined as

\begin{align}
\mathbf{Ad}_{ba} =
\begin{bmatrix}
\mathbf{R}_{ba} & \mathbf{0} \\
\widetilde{{}^{b}\mathbf{r}_a}\,\mathbf{R}_{ba} & \mathbf{R}_{ba}
\end{bmatrix}
\label{eq:Ad}
\end{align}

where $\mathbf{R}_{ba}$ transforms vectors from frame $a$ to frame $b$, and 
${}^{b}\mathbf{r}_a$ is the position of frame $a$ expressed in frame $b$. 
The instance $\mathbf{Ad}_{oi}$ used above is obtained by setting $a = i$ 
and $b = o$.

In Fig.~\ref{fig_1}, $\boldsymbol{\mathcal{W}}_{extBi} = 
[\boldsymbol{\tau}_{extBi}^\top, \mathbf{F}_{extBi}^\top]^\top$ and 
$\boldsymbol{\mathcal{W}}_{extTi} = [\boldsymbol{\tau}_{extTi}^\top, 
\mathbf{F}_{extTi}^\top]^\top$ are the wrenches applied at the base and tip 
of the link respectively. Each wrench is an element of $\mathfrak{se}^*(3)$, 
the dual space of $\mathfrak{se}(3)$, and pairs with the body-fixed twist 
$\mathbf{V}_i \in \mathfrak{se}(3)$ via the power product 
$\boldsymbol{\mathcal{W}}^\top \mathbf{V} = \boldsymbol{\tau}^\top\boldsymbol{\omega} 
+ \mathbf{F}^\top\mathbf{v}$, 
representing instantaneous power delivered to the body.

Following Euler--Bernoulli beam theory, cross-sections are assumed 
to remain planar and perpendicular to the principal axes after 
deformation. Under this assumption, the reference coordinates ${}^{i}\mathbf{q}_{b_i}$, elastic displacement ${}^{i}\mathbf{q}_{\xi_i}$, and their coordinate derivatives are

\begin{align}
{}^{i}\mathbf{q}_{b_i} &=
\begin{bmatrix} \mathbf{0}_3 \\ {}^{i}\mathbf{r}_{b_i} \end{bmatrix}, \quad
{}^{i}\dot{\mathbf{q}}_{b_i} = \mathbf{0}, \quad
{}^{i}\mathbf{r}_{b_i} =
\begin{bmatrix} \xi_i \\ {}^{i}r_{y_{b_i}} \\ {}^{i}r_{z_{b_i}} \end{bmatrix}
\label{eq:5} \\[6pt]
{}^{i}\mathbf{q}_{\xi_i} &=
\begin{bmatrix} \mathbf{0}_3 \\ {}^{i}\mathbf{r}_{\xi_i} \end{bmatrix}, \quad
{}^{i}\dot{\mathbf{q}}_{\xi_i} =
\begin{bmatrix} \mathbf{0}_3 \\ {}^{i}\dot{\mathbf{r}}_{\xi_i} \end{bmatrix}
\label{eq:8}
\end{align}

The zero rotational components reflect the Euler--Bernoulli assumption that 
cross-sections undergo no independent rotation beyond that of the body-fixed 
frame and translational deformation of beam element. ${}^{i}\dot{\mathbf{q}}_{b_i} = \mathbf{0}$ confirms that the 
reference geometry is time-invariant in the body-fixed frame, carrying no 
dynamic degrees of freedom. The corresponding twists ${}^{i}\mathbf{V}_{b_i}$ 
and ${}^{i}\mathbf{V}_{\xi_i} \in \mathfrak{se}(3)$ incorporate the 
rotational contribution $\tilde{\boldsymbol{\omega}}_i\mathbf{r}$ 
as in~(\ref*{eq:twist}), and are explicitly derived in 
Appendix~\ref{app:kinematics} following the transport theorem.

The position of element $d\Omega_i$ relative to the body-fixed and inertial 
frames follows as

\begin{align}
\mathbf{r}_{i {b_i}} &= \mathbf{r}_{\xi_i} + \mathbf{r}_{b_i} \label{eq:10} \\
\mathbf{r}_{ob_i} &= \mathbf{r}_i + \mathbf{r}_{i {b_i}} \label{eq:11}
\end{align}

Connection between links is implemented via holonomic joint constraints as 
depicted in Fig.~\ref{fig_2}. The interaction wrenches 
$\boldsymbol{\mathcal{W}}_{i} = [\boldsymbol{\tau}_{i}^\top, 
\mathbf{F}_{i}^\top]^\top \in \mathfrak{se}^*(3)$ represent forces and moments 
between adjacent links through the joint constraints.

The time derivative in the inertial frame is denoted $\frac{d(\cdot)}{dt}$, 
the body-fixed time derivative as $\dot{(\cdot)}$, and the spatial derivative 
with respect to $\xi_i$ as $(\cdot)'$. The following notational simplifications denote terms expressed in the 
body-fixed frame throughout: 
$\mathbf{r}_i = {}^{i}\mathbf{r}_i$, 
$\mathbf{r}_{\xi_i} = {}^{i}\mathbf{r}_{\xi_i}$, 
$\mathbf{r}_{b_i} = {}^{i}\mathbf{r}_{b_i}$, 
$\dot{\mathbf{r}}_i = {}^{i}\dot{\mathbf{r}}_i$, 
$\dot{\mathbf{r}}_{\xi_i} = {}^{i}\dot{\mathbf{r}}_{\xi_i}$,
$\mathbf{v}_i = {}^{i}\mathbf{v}_i$,
$\mathbf{v}_{\xi_i} = {}^{i}\mathbf{v}_{\xi_i}$,
$\boldsymbol{\omega}_i = {}^{i}\boldsymbol{\omega}_i$.
Superscript $i$ is similarly removed from coordinates $\mathbf{q}$, 
derivatives $\dot{\mathbf{q}}$, and twists $\mathbf{V}$ when expressed in 
body-fixed frame $\mathbf{\mathcal{F}}_i$. 

\section{Screw-theoretic PDE dynamic model of the flexible link} \label{sec:3}

Based on variational analysis of the Hamiltonian of flexible link $i$ on 
$SE(3)$, the equations of motion were derived in a preceding 
study~\cite{Yaqubi2026} using the screw-theoretic framework introduced in 
Section~\ref{sec:2}. The derivation proceeds geometrically on the Lie group 
$SE(3)$, yielding an inertia operator $\mathbf{M}^*_i$ that is symmetric and 
positive definite by construction, and Coriolis and centrifugal terms that 
arise from the adjoint action $\mathrm{ad}_{\mathbf{V}_i}$ on the body-fixed 
twist. The model expresses the three-dimensional motion of the link in inertial space in 
terms of the body-fixed twist $\mathbf{V}_i \in \mathfrak{se}(3)$ and 
externally applied wrenches $\boldsymbol{\mathcal{W}}^*_i \in \mathfrak{se}^*(3)$, 
presenting translational and rotational dynamics in a unified and dynamically 
consistent form. Furthermore, the deformation field $\mathbf{r}_{\xi_i}(\xi_i,t)$ 
is governed by a PDE in both space and time, preserving the infinite-dimensional 
nature of the flexibility dynamics rather than truncating it to a 
finite-dimensional ODE system a priori --- a key distinction from assumed-modes 
and finite element approaches where spatial discretization precedes dynamic 
modeling. This structure enables extension to multibody robotic systems via 
chaining of the link dynamics and constraint-based incorporation of interaction 
wrenches, which is the subject of Section~\ref{sec:4}.

The link dynamics model features a set of coupled PDEs comprising the governing 
dynamics~(\ref*{eq:26}), the deformation PDE~(\ref*{eq:34}), and boundary conditions~(\ref*{eq:35}--\ref*{eq:38}).

The dynamic equation is

\begin{align}
\mathbf{M}^*_i\,\dot{\mathbf{V}}_i + \mathbf{D}^*_i(\dot{\mathbf{V}}_{\xi_i}) 
+ \mathbf{H}^*_i = \boldsymbol{\mathcal{W}}^*_i \label{eq:26}
\end{align}

where $\dot{\mathbf{V}}_i$ is the body-fixed derivative of the twist, defined in Appendix~\ref{app:kinematics}~(\ref*{eq:Vdot_coord}), and $\dot{\mathbf{V}}_{\xi_i}$ 
is its elastic counterpart. The corresponding terms in the model including inertia $\mathbf{M}^*_i$, distributed inertia $\mathbf{D}^*_i(\dot{\mathbf{V}}_{\xi_i})$, generalized forces $\mathbf{H}^*_i$, and effective wrench $\boldsymbol{\mathcal{W}}^*_i$ are obtained as

\begin{gather}
\mathbf{M}^*_i =
\begin{bmatrix}
\mathbf{I}_{b_i} & 
\rho_i A_i\int_{a_i}^{c_i}\tilde{\mathbf{r}}_{b_i}\,d\xi \\[6pt]
-\rho_i A_i\int_{a_i}^{c_i}\tilde{\mathbf{r}}_{b_i}\,d\xi &
m_i \mathbf{I}_3
\end{bmatrix} \label{eq:27} \\[10pt]
\mathbf{D}^*_i(\dot{\mathbf{V}}_{\xi_i}) =
\begin{bmatrix}
\rho_i A_i\int_{a_i}^{c_i}\tilde{\mathbf{r}}_{b_i}\dot{\mathbf{v}}_{\xi_i}\,d\xi \\[6pt]
\rho_i A_i\int_{a_i}^{c_i}\dot{\mathbf{v}}_{\xi_i}\,d\xi
\end{bmatrix} \label{eq:28} \\[10pt]
\mathbf{H}^*_i = \begin{bmatrix}
\begin{split}
&\rho_i A_i\int_{a_i}^{c_i} \tilde{\mathbf{r}}_{b_i}\tilde{\boldsymbol{\omega}}_i\mathbf{v}_{ib_i}\,d\xi + \rho_i A_i\int_{a_i}^{c_i}\widetilde{\mathbf{v}_{ib_i}}\mathbf{v}_{ob_i}\,d\xi\\
&+ m_i\tilde{\mathbf{r}}_i\mathbf{R}_{oi}^\top\mathbf{g} + \rho_i A_i\int_{a_i}^{c_i}\tilde{\mathbf{r}}_{b_i} \mathbf{R}_{oi}^\top\mathbf{g}\,d\xi \\
&+ \rho_i A_i \int_{a_i}^{c_i}\tilde{\mathbf{r}}_{\xi_i} \mathbf{R}_{oi}^\top\mathbf{g}\,d\xi + \int_{a_i}^{c_i}\tilde{\mathbf{r}}_{\xi_i}' \mathbf{I}_{v1_i}\mathbf{r}_{\xi_i}'\,d\xi \\
&+ \int_{a_i}^{c_i}\tilde{\mathbf{r}}_{\xi_i}'' \mathbf{I}_{v2_i}\mathbf{r}_{\xi_i}''\,d\xi 
\end{split} \\[24pt]
\rho_i A_i\int_{a_i}^{c_i} \tilde{\boldsymbol{\omega}}_i\mathbf{v}_{ib_i}\,d\xi + m_i\mathbf{R}_{oi}^\top\mathbf{g}
\end{bmatrix} \label{eq:29}
\end{gather}

\begin{align}
\boldsymbol{\mathcal{W}}^*_i = \begin{bmatrix}
\boldsymbol{\tau}_{extB_i} - \boldsymbol{\tau}_{extT_i} 
- \tilde{\mathbf{r}}_{\xi(a_i)}\mathbf{F}_{extB_i} 
+ \tilde{\mathbf{r}}_{\xi(c_i)}\mathbf{F}_{extT_i} \\[4pt]
\mathbf{F}_{extB_i} - \mathbf{F}_{extT_i}
\end{bmatrix} \label{eq:30}
\end{align}

where $\mathbf{v}_{i {b_i}} = \mathbf{v}_{\xi_i} + \mathbf{v}_{b_i}$
is the translational velocity of element $\Omega_i$ with respect to the frame $\mathbf{\mathcal{F}}_i$ expressed in the 
body-fixed frame.

Mass of the beam $i$ is indicated by $m_i = \rho_i A_i l_i$. The term 
$\mathbf{g}$ is the gravitational acceleration 
vector. The moment of inertia $\mathbf{I}_{b_i}$ of the distributed mass about 
the body-fixed frame origin, appearing in~(\ref*{eq:27}), is

\begin{align}
\mathbf{I}_{b_i} = -\rho_i A_i \int_{a_i}^{c_i}\tilde{\mathbf{r}}_{b_i}^2\,d\xi
\label{eq:Iibi}
\end{align}

The cross-section stiffness matrices $\mathbf{I}_{v1_i}$ and $\mathbf{I}_{v2_i}$ are

\begin{align}
\mathbf{I}_{v1_i} &= 
\begin{bmatrix}
E_iA_i & 0 & 0 \\
0 & 0 & 0 \\
0 & 0 & 0
\end{bmatrix}
\label{eq:32} \\
\mathbf{I}_{v2_i} &=
\begin{bmatrix}
0 & 0 & 0 \\
0 & E_i I_{z_i} & 0 \\
0 & 0 & E_i I_{y_i}
\end{bmatrix}
\label{eq:33}
\end{align}

where $I_{y_i}$ and $I_{z_i}$ are the second moments of area of the beam cross-section.

The deformation PDE corresponding to~(\ref*{eq:26}) is

\begin{align}
&\dot{\mathbf{v}}_i + \dot{\mathbf{v}}_{\xi_i}
- \tilde{\mathbf{r}}_{i b_i}\dot{\boldsymbol{\omega}}_i
+ \tilde{\boldsymbol{\omega}}_i\mathbf{v}_{i b_i}
+ \mathbf{R}_{oi}^\top\mathbf{g} \nonumber \\
&\quad + \frac{1}{\rho_i A_i}\mathbf{I}_{v2_i}\mathbf{r}_{\xi_i}''''
- \frac{1}{\rho_i A_i}\mathbf{I}_{v1_i}\mathbf{r}_{\xi_i}'' = 0
\label{eq:34}
\end{align}

The boundary conditions acting on link $i$ are

\begin{align}
\mathbf{I}_{v2_i}\,\mathbf{r}_{\xi_i}^{'''}(a_i) - \mathbf{I}_{v1_i}\,\mathbf{r}_{\xi_i}^{'}(a_i)
&= \mathbf{F}_{extB_i} \label{eq:35} \\
\mathbf{I}_{v2_i}\,\mathbf{r}_{\xi_i}^{'''}(c_i) - \mathbf{I}_{v1_i}\,\mathbf{r}_{\xi_i}^{'}(c_i)
&= \mathbf{F}_{extT_i} \label{eq:36} \\
-\mathbf{I}_{v2_i}\,\mathbf{r}_{\xi_i}^{''}(a_i) &= \mathbf{H}\boldsymbol{\tau}_{extB_i} \label{eq:37} \\
-\mathbf{I}_{v2_i}\,\mathbf{r}_{\xi_i}^{''}(c_i) &= \mathbf{H}\boldsymbol{\tau}_{extT_i} \label{eq:38}
\end{align}

The matrix $H$ connects the bending moment from the deformation PDE to the corresponding external torque

\begin{align}
\mathbf{H} = \begin{bmatrix} 
0 & 0 & 0 \\
0 & 0 & 1 \\
0 & 1 & 0 
\end{bmatrix} \label{eq:39}
\end{align}

The terms $\mathbf{I}_{v1_i}\mathbf{r}_{\xi_i}''$ and 
$\mathbf{I}_{v2_i}\mathbf{r}_{\xi_i}''''$ represent the elastic restoring force 
per unit length arising from axial and bending stiffness respectively. The 
boundary conditions~(\ref*{eq:35})--(\ref*{eq:38}) are the natural 
boundary conditions of the variational principle, equating the boundary values 
of the internal elastic forces and moments to the externally applied wrenches 
$\boldsymbol{\mathcal{W}}_{extB_i}$ and $\boldsymbol{\mathcal{W}}_{extT_i} 
\in \mathfrak{se}^*(3)$.

Dynamic equation (\ref*{eq:26}) together with the deformation PDE~(\ref*{eq:34}) 
constitute a coupled PDE system in the spatial variable $\xi_i \in [a_i, c_i]$ 
and time $t$. The dynamic equation~(\ref*{eq:26}) governs the motion of the 
body-fixed frame and is an ODE in time, while~(\ref*{eq:34}) is a 
fourth-order PDE in $\xi_i$ and second-order in $t$, governing the distributed 
deformation field $\mathbf{r}_{\xi_i}(\xi_i, t)$ along the beam axis. The two equations are coupled through $\mathbf{D}^*_i$, which introduces 
elastic acceleration $\dot{\mathbf{v}}_{\xi_i}$ into the rigid body 
equation.

This PDE system forms a well-posed dynamic model for an individual flexible 
link $i$~\cite{Yaqubi2026}. In Section~\ref{sec:4}, synthesis of this model for an interconnected 
multi-link flexible manipulator is presented, resulting in a well-posed 
dynamic model for the entire system.

\section{Screw-theoretic synthesis for multi-link flexible manipulator} \label{sec:4}

To extend the dynamic model of Section~\ref{sec:3} to a multi-link system, 
adjacent links are connected via holonomic joint constraints. The velocity of 
the tip of link $i-1$ and the base of link $i$, expressed as elements of 
$\mathfrak{se}(3)$, are

\begin{align}
\mathbf{V}_{T_{i-1}} &= \mathbf{V}_{i-1} + \mathbf{V}_{\xi_{i-1}}(c_{i-1}) 
+ \mathbf{V}_{b_{i-1}}(c_{i-1}) \label{eq:46} \\
\mathbf{V}_{B_{i}} &= \mathbf{V}_{i} + \mathbf{V}_{\xi_{i}}(a_i) 
+ \mathbf{V}_{b_i}(a_i) \label{eq:47}
\end{align}

where $\mathbf{V}_{b_i}(\xi_i) = [\mathbf{0}^\top, (\tilde{\boldsymbol{\omega}}_i
\mathbf{r}_{b_i}(\xi_i))^\top]^\top \in \mathfrak{se}(3)$ is the twist of the 
reference geometry for $\xi_i$, as derived in 
Appendix~\ref{app:kinematics} ~(\ref*{eq:24}). 

The velocity-level holonomic constraint for rotational connecting joint equates the projected inertial representations of these endpoint twists as follows.

\begin{align}
\mathbf{P}_{i}\Big[\mathbf{Ad}_{o(i-1)}\,\mathbf{V}_{T_{i-1}} 
- \mathbf{Ad}_{oi}\,\mathbf{V}_{B_i}\Big] = \mathbf{0}
\label{eq:45}
\end{align}

where $\mathbf{Ad}_{oi}$ is the Adjoint defined in ~(\ref*{eq:Ad}), and 
$\mathbf{P}_{i}$ is a projection matrix that enforces the kinematic 
restrictions of the joint in the inertial frame

\begin{align}
\mathbf{P}_{i} =
\begin{bmatrix}
\mathbf{I}_{3}-{}^{o}\mathbf{A}_i{}^{o}\mathbf{A}_i^\top & \mathbf{0}_{3\times3} \\
\mathbf{0}_{3\times3} & \mathbf{I}_{3}
\end{bmatrix}
\label{eq:43}
\end{align}

with ${}^{o}\mathbf{A}_i \in \mathbb{R}^{3\times m_i}$ a matrix whose columns 
contain the allowed rotational joint axes expressed in the inertial frame, 
obtained from the body-fixed geometry as ${}^{o}\mathbf{A}_i = 
\mathbf{R}_{oi}\,{}^{i}\mathbf{A}_i$ where ${}^{i}\mathbf{A}_i$ is 
time-invariant in the body-fixed frame of link $i$.

After decomposing the multi-link system into individual links and joints as 
depicted in Fig.~\ref{fig_2}, the dynamic equation for link $i$ incorporating 
interaction wrenches $\boldsymbol{\mathcal{W}}_i \in \mathfrak{se}^*(3)$ is

\begin{align}
\mathbf{M}^*_i\,\dot{\mathbf{V}}_i + \mathbf{D}^*_i(\dot{\mathbf{V}}_{\xi_i}) 
+ \mathbf{H}^*_i = \boldsymbol{\mathcal{W}}^*_i + \boldsymbol{\mathcal{W}}_{i} 
- {}^{i}\boldsymbol{\mathcal{W}}_{{i-1}} \label{eq:48}
\end{align}

The interaction wrench from the preceding link transforms to frame 
${}^i\!\mathcal{F}$ as a co-vector under the co-Adjoint

\begin{align}
{}^i\boldsymbol{\mathcal{W}}_{{i-1}} = \mathbf{Ad}_{io}^{-\top}
\mathbf{Ad}_{o(i-1)}^{-\top}\boldsymbol{\mathcal{W}}_{{i-1}}
\label{eq:49}
\end{align}

since wrenches $\boldsymbol{\mathcal{W}} \in \mathfrak{se}^*(3)$ transform 
under $\mathbf{Ad}^{-\top}$ instead of $\mathbf{Ad}$. Substituting~(\ref*{eq:49}) in~(\ref*{eq:48}) leads to

\begin{align}
\mathbf{M}^*_i\,\dot{\mathbf{V}}_i + \mathbf{D}^*_i(\dot{\mathbf{V}}_{\xi_i}) 
+ \mathbf{H}^*_i - \boldsymbol{\mathcal{W}}_{i} + \nonumber \\
 \mathbf{Ad}_{io}^{-\top}\mathbf{Ad}_{o(i-1)}^{-\top}\boldsymbol{\mathcal{W}}_{{i-1}} = \boldsymbol{\mathcal{W}}^*_i \label{eq:50}
\end{align}

To incorporate the constraint~(\ref*{eq:45}) into the stacked 
system, its inertial time-derivative is taken as follows.

\begin{align}
\mathbf{P}_{i}\Big[\mathbf{Ad}_{o(i-1)}\,\frac{d\mathbf{V}_{T_{i-1}}}{dt}
- \mathbf{Ad}_{oi}\,\frac{d\mathbf{V}_{B_i}}{dt}\Big] = \mathbf{0}
\label{eq:accconst1}
\end{align}

The time-derivatives of the endpoint twists, expanded via the 
Lie derivative on $\mathfrak{se}(3)$, are

\begin{align}
\frac{d\mathbf{V}_{T_{i-1}}}{dt}
&= \dot{\mathbf{V}}_{i-1}
+ \dot{\mathbf{V}}_{(i-1)\xi(c_{i-1})}
+ \mathrm{ad}_{\mathbf{V}_{i-1}}\mathbf{V}_{(i-1)\xi(c_{i-1})}
\label{eq:accconst1b} \\
\frac{d\mathbf{V}_{B_i}}{dt}
&= \dot{\mathbf{V}}_{i}
+ \dot{\mathbf{V}}_{i\xi(a_i)}
+ \mathrm{ad}_{\mathbf{V}_{i}}\mathbf{V}_{i\xi(a_i)}
\label{eq:accconst2}
\end{align}

where the small adjoint $\mathrm{ad}_{\mathbf{V}_i}$, the matrix representation 
of the Lie bracket on $\mathfrak{se}(3)$, is defined in 
Appendix~\ref{app:kinematics}~(\ref*{eq:25}).

Substituting~(\ref*{eq:accconst1b})--(\ref*{eq:accconst2}) 
into~(\ref*{eq:accconst1}) leads to

\begin{align}
\mathbf{P}_{i} \Big[
&\mathbf{Ad}_{o(i-1)}\dot{\mathbf{V}}_{i-1}
- \mathbf{Ad}_{oi}\dot{\mathbf{V}}_{i} \nonumber \\
&+ \mathbf{Ad}_{o(i-1)}\dot{\mathbf{V}}_{(i-1)\xi(c_{i-1})}
- \mathbf{Ad}_{oi}\dot{\mathbf{V}}_{i\xi(a_i)}
+ \mathbf{H}_{pi}\Big] = \mathbf{0}
\label{eq:accconst4}
\end{align}

\begin{align}
\mathbf{H}_{pi} =\,&
\mathbf{Ad}_{o(i-1)}\mathrm{ad}_{\mathbf{V}_{i-1}}\mathbf{V}_{(i-1)\xi(c_{i-1})} \nonumber\\
&- \mathbf{Ad}_{oi}\mathrm{ad}_{\mathbf{V}_{i}}\mathbf{V}_{i\xi(a_i)}
\label{eq:accconst5}
\end{align}

Stacking the governing dynamics~(\ref*{eq:50}) and 
constraint~(\ref*{eq:accconst4}), the governing algebraic-partial differential 
equations for link $i$ are

\begin{align}
\mathbf{M}^*_{ci}\,
\begin{bmatrix}
\boldsymbol{\mathcal{W}}_{{i-1}} \\
\dot{\mathbf{V}}_{i-1} \\
\boldsymbol{\mathcal{W}}_{i} \\
\dot{\mathbf{V}}_{i}
\end{bmatrix}
+ \mathbf{D}^*_{ci} + \mathbf{H}^*_{ci} = \boldsymbol{\mathcal{W}}^*_{ci} \label{eq:58}
\end{align}

where

\begin{align}
\mathbf{M}^*_{ci} = 
\scalebox{0.68}{$
\begin{bmatrix}
0 & \mathbf{P}_{i}\mathbf{Ad}_{o(i-1)} &
0 & -\mathbf{P}_{i}\mathbf{Ad}_{oi} \\[2pt]
\mathbf{Ad}_{io}^{-\top}\mathbf{Ad}_{o(i-1)}^{-\top} & \mathbf{M}^*_i &
-\mathbf{I} & 0
\end{bmatrix}$}
\label{eq:59}
\end{align}

\begin{equation}
\mathbf{D}^*_{ci} =
\begin{bmatrix}
\mathbf{P}_{i}\mathbf{Ad}_{o(i-1)}\,\dot{\mathbf{V}}_{(i-1)\xi(c_{i-1})} 
- \mathbf{P}_{i}\mathbf{Ad}_{oi}\,\dot{\mathbf{V}}_{i\xi(a_i)} \\[4pt]
\mathbf{D}^*_i(\dot{\mathbf{V}}_{\xi_i})
\end{bmatrix}
\label{eq:60}
\end{equation}

\begin{equation}
\mathbf{H}^*_{ci} =
\begin{bmatrix}
\mathbf{H}_{pi} \\[4pt]
\mathbf{H}^*_i
\end{bmatrix}
\label{eq:61}
\end{equation}

\begin{equation}
\boldsymbol{\mathcal{W}}^*_{ci} =
\begin{bmatrix}
\mathbf{0} \\[4pt]
\boldsymbol{\mathcal{W}}^*_i
\end{bmatrix}
\label{eq:62}
\end{equation}

For $i = 1$, there is no preceding link state, so the 
constraint~(\ref*{eq:accconst4}) reduces to

\begin{align}
\mathbf{P}_{1} \Big[
- \mathbf{Ad}_{o1}\dot{\mathbf{V}}_{1}
- \mathbf{Ad}_{o1}\dot{\mathbf{V}}_{1\xi(a_1)}
+ \mathbf{H}_{p1}\Big] = \mathbf{0}
\label{eq:63}
\end{align}

and the dynamics~(\ref*{eq:50}) for the first link, with $\mathbf{Ad}_{00} = 
\mathbf{I}$, reduces to

\begin{align}
\mathbf{M}^*_1\,\dot{\mathbf{V}}_1 + \mathbf{D}^*_1(\dot{\mathbf{V}}_{\xi_1}) 
+ \mathbf{H}^*_1 - \boldsymbol{\mathcal{W}}_{1} 
+ \boldsymbol{\mathcal{W}}_{0} = \boldsymbol{\mathcal{W}}^*_1 \label{eq:65}
\end{align}

The governing dynamic model of the full multi-link flexible robotic system is 
obtained by stacking~(\ref*{eq:50}) and~(\ref*{eq:accconst4}) for 
$i = 2,\ldots,n$, together with~(\ref*{eq:63}) and~(\ref*{eq:65}) for the 
base link. The resulting differential-algebraic system is obtained as follows.

\begin{align}
\mathbf{M}_s\,\boldsymbol{\Psi} + \mathbf{D}_s(\dot{\mathbf{V}}_\xi) 
+ \mathbf{h}_s = \boldsymbol{\mathcal{W}}_s
\label{eq:66}
\end{align}

The augmented state vector $\boldsymbol{\Psi}$ comprises the body-fixed twist 
derivatives $\dot{\mathbf{V}}_i$ for $i = 1,\ldots,n$ and the interaction 
wrenches $\boldsymbol{\mathcal{W}}_i$.

\begin{align}
\boldsymbol{\Psi} =
\begin{bmatrix}
\boldsymbol{\mathcal{W}}_{0} \\
\dot{\mathbf{V}}_1 \\
\boldsymbol{\mathcal{W}}_{1} \\
\dot{\mathbf{V}}_2 \\
\vdots \\
\boldsymbol{\mathcal{W}}_{n-1} \\
\dot{\mathbf{V}}_n
\end{bmatrix}
\in \mathbb{R}^{12n}, \qquad
\dot{\mathbf{V}}_i,\,\boldsymbol{\mathcal{W}}_{j} \in \mathbb{R}^6
\label{eq:67}
\end{align}

The augmented inertia matrix is

\begin{equation}
\mathbf{M}_s =
\begin{bmatrix}
\begin{smallmatrix}
0 & \mathbf{P}_{1}\mathbf{Ad}_{o1} & 0 & 0 & \cdots & 0 \\
\mathbf{I} & \mathbf{M}^*_1 & -\mathbf{I} & 0 & \cdots & 0 \\
0 & -\mathbf{P}_{2}\mathbf{Ad}_{o1} & 0 &
\mathbf{P}_{2}\mathbf{Ad}_{o2} & \cdots & 0 \\
0 & \mathbf{Ad}_{2o}^{-\top}\mathbf{Ad}_{o1}^{-\top} & \mathbf{M}^*_2 & -\mathbf{I} & \cdots & 0 \\
\vdots & \vdots & \vdots & \vdots & \ddots & \vdots \\
0 & 0 & 0 & 0 & \cdots & \mathbf{M}^*_n
\end{smallmatrix}
\end{bmatrix}
\label{eq:68}
\end{equation}

The distributed flexibility operator $\mathbf{D}_s$ stacks the $\mathbf{D}^*_i$ 
terms from the dynamic equations and the endpoint twist derivative terms from 
the constraints.

\begin{align}
\mathbf{D}_s(\dot{\mathbf{V}}_\xi) =
\begin{bmatrix}
-\mathbf{P}_{1}\mathbf{Ad}_{o1}\,\dot{\mathbf{V}}_{1\xi(a_1)} \\
\mathbf{D}^*_1(\dot{\mathbf{V}}_{\xi_1}) \\
\mathbf{P}_{2}\mathbf{Ad}_{o1}\,\dot{\mathbf{V}}_{1\xi(c_1)}
-\mathbf{P}_{2}\mathbf{Ad}_{o2}\,\dot{\mathbf{V}}_{2\xi(a_2)} \\
\mathbf{D}^*_2(\dot{\mathbf{V}}_{\xi_2}) \\
\vdots \\
\mathbf{D}^*_n(\dot{\mathbf{V}}_{\xi_n})
\end{bmatrix}
\label{eq:69}
\end{align}

The augmented nonlinear vector and stacked external wrench vector are:

\begin{align}
\mathbf{h}_s =
\begin{bmatrix}
\mathbf{H}_{p1} \\
\mathbf{H}^*_1 \\
\mathbf{H}_{p2} \\
\mathbf{H}^*_2 \\
\vdots \\
\mathbf{H}^*_n
\end{bmatrix}
\label{eq:70}
\end{align}

\begin{align}
\boldsymbol{\mathcal{W}}_s =
\begin{bmatrix}
\mathbf{0} \\
\mathbf{{\mathcal{W}}}^*_1 \\
\mathbf{0} \\
\mathbf{{\mathcal{W}}}^*_2 \\
\vdots \\
\mathbf{{\mathcal{W}}}^*_n
\end{bmatrix}
\label{eq:71}
\end{align}

The differential-algebraic PDE system~(\ref*{eq:66}), solved in parallel with the deformation PDE~(\ref*{eq:34}), constitutes the complete screw-theoretic multibody model. In its current form, the system is not directly amenable to numerical solution: the dynamic equation~(\ref*{eq:66}) involves distributed integrals over the deformation field in the term $\mathbf{D}_s(\dot{\mathbf{V}}_\xi)$, while the deformation PDE~(\ref*{eq:34}) governs pointwise displacements of individual element twists ${\mathbf{v}}_{\xi_i}$, and the coupling between these two representations prevents reduction to a finite-dimensional algebraic form. Section~\ref{sec:5} addresses this by applying modal projection to the PDE system, yielding a finite-dimensional representation in which all differential-algebraic states — body-fixed twists, modal coordinates, and interaction wrenches — are represented explicitly.

\section{Modal projection of the PDE system} \label{sec:5} 

The complete model derived in Section~\ref{sec:4} consists of the 
stacked differential-algebraic system~(\ref*{eq:66}), the deformation 
PDE~(\ref*{eq:34}) for $i = 1,\dots,n$, and the boundary 
conditions~(\ref*{eq:35}--\ref*{eq:38}) for $i = 1,\dots,n$. As 
established in Section~\ref{sec:3}, this system is not directly amenable 
to numerical solution due to the coupling between distributed integrals 
in~(\ref*{eq:66}) and the pointwise deformation field in~(\ref*{eq:34}). 
Modal projection is applied to obtain a finite-dimensional representation 
in which all system states — body-fixed twists, modal coordinates, and 
interaction wrenches — appear explicitly.

To integrate the screw model using standard numerical solutions, the following coordinate equivalent terms are substituted in the dynamic model.

\begin{align}
\mathbf{v}_i &= \dot{\mathbf{r}}_i + \tilde{\boldsymbol{\omega}}_i\mathbf{r}_i, \quad
\mathbf{v}_{b_i} = \tilde{\boldsymbol{\omega}}_i\mathbf{r}_{b_i}, \quad
\mathbf{v}_{\xi_i} = \dot{\mathbf{r}}_{\xi_i} + \tilde{\boldsymbol{\omega}}_i\mathbf{r}_{\xi_i} \notag \\
\dot{\mathbf{v}}_{\xi_i} &= \ddot{\mathbf{r}}_{\xi_i} 
- \tilde{\mathbf{r}}_{\xi_i} \dot{\boldsymbol{\omega}}_i 
+ \tilde{\boldsymbol{\omega}}_i\dot{\mathbf{r}}_{\xi_i}, \quad
\tilde{\boldsymbol{\omega}}_i\mathbf{v}_{ib_i} = 
\tilde{\boldsymbol{\omega}}_i\dot{\mathbf{r}}_{\xi_i}
+ \tilde{\boldsymbol{\omega}}_i^2\mathbf{r}_{ib_i}
\label{eq:twist_coords}
\end{align}
 
The displacement field is represented via separation of variables as:

\begin{align}
\mathbf{r}_{\xi_i} &= \boldsymbol{\phi}_i \, \boldsymbol{\eta}_i \label{eq:72} \\
\boldsymbol{\phi}_i &= \begin{bmatrix} \boldsymbol{\phi}_{i_1}, & \cdots, & \boldsymbol{\phi}_{i_r} \end{bmatrix} \label{eq:73} \\
\boldsymbol{\eta}_i &= \begin{bmatrix} \eta_{i_1}\\ \vdots \\ \eta_{i_r} \end{bmatrix} \label{eq:74} \\
\boldsymbol{\phi}_{i_p} &= \begin{bmatrix}
\boldsymbol{\phi}_{x_{i_p}}^\top & 0 & 0 \\
0 & \boldsymbol{\phi}_{y_{i_p}}^\top & 0 \\
0 & 0 & \boldsymbol{\phi}_{z_{i_p}}^\top
\end{bmatrix}, \quad p = 1,\ldots,r \label{eq:75}\\
\boldsymbol{\eta}_{i_p} &= \begin{bmatrix}
\eta_{x_{i_p}} \\
\eta_{y_{i_p}} \\
\eta_{z_{i_p}}
\end{bmatrix}, \quad p = 1,\ldots,r  \label{eq:76}
\end{align}

The discretized expression of displacement is constituted of $\boldsymbol{\phi}_i$
denoting the spatial basis functions and $\boldsymbol{\eta}_i$ representing the
temporal modal coordinates. Basis functions $\boldsymbol{\phi}_{x_{i_p}}$,
$\boldsymbol{\phi}_{y_{i_p}}$ and $\boldsymbol{\phi}_{z_{i_p}}$ for
$p=1,\cdots,r$ are uniquely determined based on boundary
conditions~(\ref*{eq:35}--\ref*{eq:38}) and must satisfy the orthogonality
condition $\int_{a_i}^{c_i} \phi_{kp}\,\phi_{kq}\,d\xi_i = \delta_{pq}$ for
$k \in \{x_i, y_i, z_i\}$, where $\delta_{pq}$ is the Kronecker delta. The
integer $r$ describes the number of allocated spatial modes. For convenience,
the same number of activated modes $r$ is assigned to all flexible links
$i=1,\cdots,n$. Matrix dimensions are $\boldsymbol{\phi}_i \in \mathbb{R}^{3
\times 3r}$ and $\boldsymbol{\eta}_i \in \mathbb{R}^{3r \times 1}$. All
elements of $\boldsymbol{\phi}_i$ belong to the function space below.

\begin{align}
\phi_{kp}(\xi_i) &\in L^2(a_i,c_i) \times H^2(a_i,c_i) \label{eq:77} \\[4pt]
L^2(a_i,c_i) &= 
\left\{
f \;\middle|\;
\int_{a_i}^{c_i} |f(\xi_i)|^{2}\, d\xi_i < \infty
\right\} \label{eq:78} \\[4pt]
H^{s}(a_i,c_i) &= 
\left\{
f \;\middle|\;
\frac{\partial^{k} f}{\partial \xi_i^{k}} \in L^2(a_i,c_i),\;
k=0,\dots,s
\right\}
\label{eq:79}
\end{align}

Since $\mathbf{v}_\xi$ is a body-fixed twist component, its 
modal expansion includes the frame rotation transport 
term as

\begin{align}
\mathbf{v}_\xi = \dot{\mathbf{r}}_\xi 
+ \tilde{\boldsymbol{\omega}}_i\mathbf{r}_\xi
= \boldsymbol{\phi}\dot{\boldsymbol{\eta}} 
+ \tilde{\boldsymbol{\omega}}_i\boldsymbol{\phi}\boldsymbol{\eta}
\label{eq:vxi_modal}
\end{align}

The body-fixed time derivative $\dot{\mathbf{v}}_\xi$ used 
in $\mathbf{D}_s(\dot{\mathbf{v}}_\xi)$~(\ref*{eq:69}) 
expands as

\begin{align}
\dot{\mathbf{v}}_\xi 
= \ddot{\mathbf{r}}_\xi 
- \tilde{\mathbf{r}}_\xi\dot{\boldsymbol{\omega}}_i 
+ \tilde{\boldsymbol{\omega}}_i\dot{\mathbf{r}}_\xi
= \boldsymbol{\phi}\ddot{\boldsymbol{\eta}}
- \widetilde{\boldsymbol{\phi}\boldsymbol{\eta}}\,\dot{\boldsymbol{\omega}}_i
+ \tilde{\boldsymbol{\omega}}_i\boldsymbol{\phi}\dot{\boldsymbol{\eta}}
\label{eq:vxidot_modal}
\end{align}

containing inertia $\boldsymbol{\phi}\ddot{\boldsymbol{\eta}}$, 
angular acceleration coupling 
$-\widetilde{\boldsymbol{\phi}\boldsymbol{\eta}}\dot{\boldsymbol{\omega}}_i$, 
and Coriolis contribution $\tilde{\boldsymbol{\omega}}_i\boldsymbol{\phi}\dot{\boldsymbol{\eta}}$ 
terms. 

Substituting the discretized expressions, (\ref*{eq:twist_coords}), (\ref*{eq:vxi_modal}), and (\ref*{eq:vxidot_modal}) in displacement 
equation (\ref*{eq:34}) results in

\begin{align}
&\ddot{\mathbf{r}}_i + \boldsymbol{\phi}_i\ddot{\boldsymbol{\eta}}_i
- \tilde{\mathbf{r}}_{i {b_i}}\dot{\boldsymbol{\omega}}_i
+ 2\tilde{\boldsymbol{\omega}}_i\dot{\mathbf{r}}_{i {b_i}}
+ \tilde{\boldsymbol{\omega}}_i^2\mathbf{r}_{i {b_i}}
+ \mathbf{R}_{oi}^\top\mathbf{g} \nonumber \\
&\quad + \frac{1}{\rho_i A_i}\mathbf{I}_{v2_i}\boldsymbol{\phi}_i''''\boldsymbol{\eta}_i
- \frac{1}{\rho_i A_i}\mathbf{I}_{v1_i}\boldsymbol{\phi}_i''\boldsymbol{\eta}_i = 0
\label{eq:81}
\end{align}

Multiplying by $\boldsymbol{\phi}_{ip}^\top$ for $p = 1, \dots, r$ and
integrating over $\xi_i\in [a_i, b_i]$ leads to

\begin{align}
    & \int_{a_i}^{c_i} \boldsymbol{\phi}_{ip}^\top \, d\xi_i \, \ddot{\mathbf{r}}_i
    + \int_{a_i}^{c_i} \boldsymbol{\phi}_{ip}^\top \boldsymbol{\phi}_{ip} \, d\xi_i \, \ddot{\boldsymbol{\eta}}_i
    + \int_{a_i}^{c_i} \left(-\boldsymbol{\phi}_{ip}^\top \tilde{\mathbf{r}}_{i {b_i}}\right) d\xi_i \, \dot{\boldsymbol{\omega}}_i \notag \\
    & + \frac{1}{\rho_i A_i} \mathbf{I}_{v2_i} \int_{a_i}^{c_i} \boldsymbol{\phi}_{ip}^{''''\top} \boldsymbol{\phi}_{ip} \, d\xi_i \, \boldsymbol{\eta}_i
    - \frac{1}{\rho_i A_i} \mathbf{I}_{v1_i} \int_{a_i}^{c_i} \boldsymbol{\phi}_{ip}^{''\top} \boldsymbol{\phi}_{ip} \, d\xi_i \, \boldsymbol{\eta}_i \notag \\
    & + \int_{a_i}^{c_i} \boldsymbol{\phi}_{ip}^\top \left(
    2\tilde{\boldsymbol{\omega}}_i\dot{\mathbf{r}}_{i {b_i}}
    + \tilde{\boldsymbol{\omega}}_i^2\mathbf{r}_{i {b_i}}
    + \mathbf{R}_{oi}^\top\mathbf{g}
    \right) d\xi_i = 0, \quad p = 1, \dots, r
    \label{eq:82}
\end{align}

The integrated terms in~(\ref*{eq:82}) are stacked for $p = 1, \dots, r$ to
obtain a well-posed set of equations corresponding to the displacement of
flexible link $i$.

\begin{align}
\mathbf{M}_{D\phi,i} \, \boldsymbol{\Psi}_{Z_i} + \mathbf{M}_{W\phi,i} \ddot{\boldsymbol{\eta}}_i + \mathbf{h}_{i\phi} &= \mathbf{0} \label{eq:83}
\end{align}

\begin{align}
\boldsymbol{\Psi}_{Z_i} = \begin{bmatrix}
\boldsymbol{\mathcal{W}}_{i-1}\\
\dot{\mathbf{Z}}_{i}
\end{bmatrix}\label{eq:84}
\end{align}

\begin{align}
\mathbf{Z}_i = \begin{bmatrix}\boldsymbol{\omega}_i\\
\dot{\mathbf{r}}_i
\end{bmatrix} \in \mathbb{R}^6
\label{eq:zi_def}
\end{align}

\begin{align}
\mathbf{M}_{D\phi,i} = \begin{bmatrix}
\mathbf{0} & \mathbf{0} &
\displaystyle\int_{a_i}^{c_i} \left(-\boldsymbol{\phi}_{i1}^\top \tilde{\mathbf{r}}_{i {b_i}}\right) d\xi_i &
\displaystyle\int_{a_i}^{c_i} \boldsymbol{\phi}_{i1}^\top \, d\xi_i \\[6pt]
\vdots & \vdots & \vdots & \vdots \\[2pt]
\mathbf{0} & \mathbf{0} &
\displaystyle\int_{a_i}^{c_i} \left(-\boldsymbol{\phi}_{ir}^\top \tilde{\mathbf{r}}_{i {b_i}}\right) d\xi_i &
\displaystyle\int_{a_i}^{c_i} \boldsymbol{\phi}_{ir}^\top \, d\xi_i
\end{bmatrix}\label{eq:85}
\end{align}

\begin{align}
\mathbf{M}_{W\phi,i} = \begin{bmatrix}
\displaystyle\int_{a_i}^{c_i} \boldsymbol{\phi}_{i1}^\top \boldsymbol{\phi}_{i1} \, d\xi_i & \cdots & \mathbf{0} \\[6pt]
\vdots & \ddots & \vdots \\[2pt]
\mathbf{0} & \cdots & \displaystyle\int_{a_i}^{c_i} \boldsymbol{\phi}_{ir}^\top \boldsymbol{\phi}_{ir} \, d\xi_i
\end{bmatrix}\label{eq:86}
\end{align}

\begin{equation}
\label{eq:88}
\begin{aligned}
\mathbf{h}_{i\phi} &= \frac{1}{\rho_i A_i}\mathbf{I}_{v2_i} \int_{a_i}^{c_i} \boldsymbol{\phi}_{ip}^{''''\top} \boldsymbol{\phi}_{ip} \, d\xi_i \, \boldsymbol{\eta}_i \\
&\quad - \frac{1}{\rho_i A_i}\mathbf{I}_{v1_i} \int_{a_i}^{c_i} \boldsymbol{\phi}_{ip}^{''\top} \boldsymbol{\phi}_{ip} \, d\xi_i \, \boldsymbol{\eta}_i \\
&\quad + \int_{a_i}^{c_i} \boldsymbol{\phi}_{ip}^\top \Big( 2\tilde{\boldsymbol{\omega}}_i\dot{\mathbf{r}}_{i {b_i}} + \tilde{\boldsymbol{\omega}}_i^2\mathbf{r}_{i {b_i}} \\
&\qquad + \mathbf{R}_{oi}^\top\mathbf{g} \Big) \, d\xi_i, \quad p = 1,\dots,r
\end{aligned}
\end{equation}

To obtain the displacement model for the entire multibody system,
(\ref*{eq:83}) is stacked for $i=1,\dots,n$, where
$\boldsymbol{\Psi}_Z = [\boldsymbol{\Psi}_{Z_1}^\top,\ldots,
\boldsymbol{\Psi}_{Z_n}^\top]^\top$, yielding
\begin{align}
\mathbf{M}_{D\phi}\boldsymbol{\Psi}_Z
+ \mathbf{M}_{W\phi}\ddot{\underline{\boldsymbol{\eta}}}
+ \mathbf{h}_\phi = \mathbf{0}
\label{eq:89}
\end{align}
where $\mathbf{M}_{D\phi} = \mathrm{diag}(\mathbf{M}_{D\phi,1},
\ldots,\mathbf{M}_{D\phi,n})$,
$\mathbf{M}_{W\phi} = \mathrm{diag}(\mathbf{M}_{W\phi,1},
\ldots,\mathbf{M}_{W\phi,n})$,
$\ddot{\underline{\boldsymbol{\eta}}} =
[\ddot{\boldsymbol{\eta}}_1^\top,\ldots,
\ddot{\boldsymbol{\eta}}_n^\top]^\top$,
and $\mathbf{h}_\phi =
[\mathbf{h}_{1\phi}^\top,\ldots,\mathbf{h}_{n\phi}^\top]^\top$.

However, simply substituting the discretized displacement into the
displacement model does not yield the complete discretized model for
the entire system. The distributed flexibility term
$\mathbf{D}_s(\dot{\mathbf{V}}_\xi)$ in~(\ref*{eq:66}) must also be
discretized, and the twist-based state vector $\boldsymbol{\Psi}$
must be re-expressed in terms of $\boldsymbol{\Psi}_Z$. Both
substitutions are carried out simultaneously by applying the modal
expansion~(\ref*{eq:72}) and the coordinate
relation $\mathbf{V}_i = \mathbf{Z}_i + \boldsymbol{\gamma}_i$,
$\boldsymbol{\gamma}_i =
[\mathbf{0}_3^\top,\,(\tilde{\boldsymbol{\omega}}_i\mathbf{r}_i)^\top]^\top$,
to~(\ref*{eq:66}). This yields
\begin{align}
\mathbf{D}_s(\dot{\mathbf{V}}_\xi) &=
\boldsymbol{M}_D\,\ddot{\underline{\boldsymbol{\eta}}}
+ \boldsymbol{M}_{D\Psi}\boldsymbol{\Psi}_Z
+ \mathbf{h}_{M_D}
\label{eq:95}\\
\mathbf{M}_s\boldsymbol{\Psi} &=
\mathbf{M}_s(\mathbf{I}+\mathbf{T}_\gamma)\boldsymbol{\Psi}_Z
+ \mathbf{h}_{M_s}
\label{eq:MsPsi}
\end{align}
where $\boldsymbol{\Psi}$ is the twist-based state vector
from~(\ref*{eq:67}), and $\mathbf{T}_\gamma =
\mathrm{diag}(\mathbf{0}_6,\mathbf{T}_{\gamma_1},\ldots,
\mathbf{0}_6,\mathbf{T}_{\gamma_n})$ with
$\mathbf{T}_{\gamma_i} =
\bigl[\begin{smallmatrix}\mathbf{0}_3&\mathbf{0}_3\\
-\tilde{\mathbf{r}}_i&\mathbf{0}_3\end{smallmatrix}\bigr]$
absorbs the acceleration-level correction from $\dot{\boldsymbol{\gamma}}_i$,
and
\begin{align}
\mathbf{h}_{M_s} = \mathbf{M}_s\boldsymbol{\Gamma}^*, \qquad
\boldsymbol{\Gamma}^* =
\begin{bmatrix}\mathbf{0}_6\\\boldsymbol{\gamma}_1^*\\
\vdots\\\mathbf{0}_6\\\boldsymbol{\gamma}_n^*\end{bmatrix}, \qquad
\boldsymbol{\gamma}_i^* =
\begin{bmatrix}\mathbf{0}_3\\\tilde{\boldsymbol{\omega}}_i\dot{\mathbf{r}}_i
\end{bmatrix}
\label{eq:HMs}
\end{align}
collects the Coriolis and centripetal contributions arising from the 
twist-to-$\mathbf{Z}$ coordinate transformation. The full form of matrices $\boldsymbol{M}_D$,
$\boldsymbol{M}_{D\Psi}$, and $\mathbf{h}_{M_D}$ after indicated substitutions are given
in the Appendix~\ref{app:modal}~(\ref*{eq:B1}--\ref*{eq:B3}). Substituting~(\ref*{eq:95})
and~(\ref*{eq:MsPsi}) into~(\ref*{eq:66}) and
defining $\mathbf{M}_s^* = \mathbf{M}_s(\mathbf{I}+\mathbf{T}_\gamma)$
gives the discretized dynamic equation
\begin{align}
\mathbf{M}_{\mathrm{dyn}}\boldsymbol{\Psi}_Z
+ \boldsymbol{M}_D\ddot{\underline{\boldsymbol{\eta}}}
+ \mathbf{h}_{\mathrm{dyn}} &= \boldsymbol{\mathcal{W}}_s,
\notag\\
\mathbf{M}_{\mathrm{dyn}} = \mathbf{M}_s^* + \mathbf{M}_{D\Psi},
&\qquad
\mathbf{h}_{\mathrm{dyn}} = \mathbf{h}_{M_s} + \mathbf{h}_{M_D}
\label{eq:97}
\end{align}
Stacking~(\ref*{eq:97}) over the dynamic rows and~(\ref*{eq:89})
over the displacement rows gives the complete discretized model
\begin{align}
\mathbf{M}_{\mathrm{sys}}\,\mathbf{q}_{\mathrm{sys}}
+ \mathbf{H}_{\mathrm{sys}} = \boldsymbol{\mathcal{W}}_{\mathrm{sys}}
\label{eq:98}
\end{align}
with
\begin{equation}
\label{eq:100}
\begin{split}
\mathbf{q}_{\mathrm{sys}} = \begin{bmatrix} \boldsymbol{\Psi}_Z \\ \ddot{\underline{\boldsymbol{\eta}}} \end{bmatrix}, \quad 
&\mathbf{M}_{\mathrm{sys}} = \begin{bmatrix} \mathbf{M}_{\mathrm{dyn}} & \boldsymbol{M}_D \\ \mathbf{M}_{D\phi} & \mathbf{M}_{W\phi} \end{bmatrix} \\
\mathbf{H}_{\mathrm{sys}} = \begin{bmatrix} \mathbf{h}_{\mathrm{dyn}} \\ \mathbf{h}_\phi \end{bmatrix}, \quad 
&\boldsymbol{\mathcal{W}}_{\mathrm{sys}} = \begin{bmatrix} \boldsymbol{\mathcal{W}}_s \\ \mathbf{0} \end{bmatrix}
\end{split}
\end{equation}

The modal form~(\ref*{eq:98}) provides a uniform finite-dimensional
representation of the complete multibody system in terms of the
acceleration-level unknowns $\mathbf{q}_{\mathrm{sys}}$. Once
$\mathbf{Z}_i$ is obtained by integration of $\dot{\mathbf{Z}}_i$,
the generalized coordinates $\mathbf{q}_i$ follow via the Jacobian
relation
\begin{align}
\dot{\mathbf{q}}_i = \mathbf{J}(\mathbf{q}_i)\mathbf{Z}_i =
\begin{bmatrix}
\mathbf{J}(\boldsymbol{\theta}_i) & \mathbf{0} \\
\mathbf{0} & \mathbf{I}_3
\end{bmatrix}
\begin{bmatrix}\boldsymbol{\omega}_i\\ \dot{\mathbf{r}}_i\end{bmatrix}
\label{eq:jacobian}
\end{align}
and subsequent integration, and the deformation field
$\mathbf{r}_{\xi_i}(\xi,t)$ is reconstructed via~(\ref*{eq:72}).
Solution of~(\ref*{eq:98}) and well-posedness of the modal projection is established in
Section~\ref{sec:6}.

\section{Well-posedness of DAE system}\label{sec:6}

The modal form~(\ref*{eq:98}) is a differential-algebraic system
coupling body-frame accelerations $\dot{\mathbf{Z}}_i$, interaction
wrenches $\boldsymbol{\mathcal{W}}_i$, and modal accelerations
$\ddot{\underline{\boldsymbol{\eta}}}$. Well-posedness is established
by recasting~(\ref*{eq:98}) in abstract Cauchy form and applying the
Picard--Lindelöf theorem~\cite{hairer2006}.

Several structural properties of $\mathbf{M}_{\mathrm{sys}}$ are
first established. The basis functions $\boldsymbol{\phi}_i$ are
linearly independent by the orthogonality condition. $\mathbf{I}_{b_i}$
is positive definite from~(\ref*{eq:Iibi}), so $\mathbf{M}_i^*$ is
full-rank from~(\ref*{eq:27}). The constraint rows of $\mathbf{M}_s$
are linearly independent from the dynamic rows. The block
$\boldsymbol{M}_D$ is generally nonsymmetric; however,
from~(\ref*{eq:85}) and~(\ref*{eq:B1}), for $k = 1,\dots,n$
\begin{align}
\mathbf{M}_{D\phi}(i,j) =
\begin{cases}
\boldsymbol{M}_D(j,i), & j = 4k+3 \text{ or } j = 4k+4, \\
\mathbf{0}, & j = 4k+1 \text{ or } j = 4k+2
\end{cases}
\label{eq:103}
\end{align}
so all wrench columns of $\mathbf{M}_{D\phi}$ are zero.
$\mathbf{M}_{W\phi}$ is symmetric positive definite by construction
from~(\ref*{eq:86}).

\textit{Invertibility of $\mathbf{M}_{\mathrm{sys}}$.}
Within $\boldsymbol{\Psi}_Z$, each block
$\boldsymbol{\Psi}_{Z_i} = [\boldsymbol{\mathcal{W}}_{i-1}^\top,
\dot{\mathbf{Z}}_i^\top]^\top$ interleaves wrench components
$\boldsymbol{\mathcal{W}}_{i-1}$, denoted $\boldsymbol{\Psi}_Z^{(alg)}$,
and body-frame acceleration components $\dot{\mathbf{Z}}_i$, denoted
$\boldsymbol{\Psi}_Z^{(diff)}$, corresponding to algebraic rows
($4k+1$, $4k+2$) and differential rows ($4k+3$, $4k+4$) of
$\mathbf{M}_{\mathrm{dyn}}$ respectively.

\textit{Step 1 — differential and modal states.}
From~(\ref*{eq:85}), $\mathbf{M}_{D\phi}$ has zero columns
at wrench slots. The kinetic energy quadratic form is evaluated over
the differential-modal subspace ($\boldsymbol{\Psi}_Z^{(alg)} =
\mathbf{0}$), over which the cross terms satisfy

\begin{equation}
\label{eq:cross}
\begin{aligned}
\boldsymbol{\Psi}_Z^{(diff)\top} \boldsymbol{M}_D\,\dot{\underline{\boldsymbol{\eta}}} 
&+ \dot{\underline{\boldsymbol{\eta}}}^\top \mathbf{M}_{D\phi}\,\boldsymbol{\Psi}_Z^{(diff)} \\
&= 2\,\dot{\underline{\boldsymbol{\eta}}}^\top \mathbf{M}_{D\phi}\,\boldsymbol{\Psi}_Z^{(diff)}
\end{aligned}
\end{equation}

where the equality follows from $\boldsymbol{M}_D^\top\big|_{diff} = \mathbf{M}_{D\phi}\big|_{diff}$ by eq.~(\ref*{eq:103}), so that $\boldsymbol{\Psi}_Z^{(diff)\top}\boldsymbol{M}_D\,\dot{\underline{\boldsymbol{\eta}}} = \dot{\underline{\boldsymbol{\eta}}}^\top\mathbf{M}_{D\phi}\,\boldsymbol{\Psi}_Z^{(diff)}$ and the two cross terms are equal. Hence, the quadratic form restricted to differential and modal states is
\begin{equation}
\label{eq:2T}
\begin{aligned}
2T &= \boldsymbol{\Psi}_Z^{(diff)\top} \mathbf{M}_{\mathrm{dyn}}^{(diff)} \boldsymbol{\Psi}_Z^{(diff)} \\
&\quad + 2\dot{\underline{\boldsymbol{\eta}}}^\top \mathbf{M}_{D\phi}\boldsymbol{\Psi}_Z^{(diff)} + \dot{\underline{\boldsymbol{\eta}}}^\top \mathbf{M}_{W\phi}\dot{\underline{\boldsymbol{\eta}}} > 0
\end{aligned}
\end{equation}
which is the kinetic energy of the flexible multibody system after
modal substitution of~(\ref*{eq:72}), strictly positive for any
nonzero $[\boldsymbol{\Psi}_Z^{(diff)\top},\,
\dot{\underline{\boldsymbol{\eta}}}^\top]^\top$ by positive
definiteness of $\mathbf{I}_{b_i}$ and $L^2$-orthogonality
of $\boldsymbol{\phi}_i$ from~(\ref*{eq:77}). Hence the submatrix
\begin{align}
\mathbf{M}_{\mathrm{sys}}^{(diff)} =
\begin{bmatrix}
\mathbf{M}_{\mathrm{dyn}}^{(diff)} & \boldsymbol{M}_D^{(diff)} \\
\mathbf{M}_{D\phi} & \mathbf{M}_{W\phi}
\end{bmatrix} \succ 0
\label{eq:Msys_diff}
\end{align}
is positive definite, hence invertible.

\textit{Step 2 — algebraic rows.}
The rotational algebraic rows ($4k+2$) of $\mathbf{M}_{\mathrm{dyn}}$
coincide with those of $\mathbf{M}_s^*$ by~(\ref*{eq:B2}) and are
linearly independent by full rank of $\mathbf{M}_i^*$
from~(\ref*{eq:27}). The translational algebraic rows ($4k+1$) carry
$\pm\mathbf{R}_{oi}\in SO(3)$ from $\mathbf{M}_{D\Psi}$, which are
full rank and act on distinct $\dot{\boldsymbol{\omega}}_i$ column
blocks across links. Since wrench variables
$\boldsymbol{\mathcal{W}}_{i-1}$ do not appear in
$\mathbf{M}_{\mathrm{sys}}^{(diff)}$, the algebraic rows are linearly
independent from the differential-modal rows. Together with Step~1,
all rows of $\mathbf{M}_{\mathrm{sys}}$ are linearly independent and
$\mathbf{M}_{\mathrm{sys}}$ is invertible. In other words, while $\mathbf{M}_{\mathrm{sys}}$ as presented in (\ref{eq:100}) interleaves algebraic and differential states for recursive consistency, it is equivalent via a permutation of rows and columns to a partitioned system where the differential-modal block is decoupled from the algebraic constraint manifold.

The unique solution of~(\ref*{eq:98}) is therefore
\begin{align}
\mathbf{q}_{\mathrm{sys}}^* =
\mathbf{M}_{\mathrm{sys}}^{-1}
\!\left(\boldsymbol{\mathcal{W}}_{\mathrm{sys}}
- \mathbf{H}_{\mathrm{sys}}\right)
\label{eq:DAESOL}
\end{align}
from which $\dot{\mathbf{Z}}_i^*$ and
$\ddot{\underline{\boldsymbol{\eta}}}^*$ are extracted from
$\boldsymbol{\Psi}_Z^*$ at the differential slots ($4k+3$, $4k+4$),
and interaction wrenches $\boldsymbol{\mathcal{W}}_i^*$ at the
algebraic slots ($4k+1$, $4k+2$). The differential states
\begin{align}
\mathbf{x}_c =
[\underline{\mathbf{q}}^\top,\,\underline{\boldsymbol{\eta}}^\top,\,
\underline{\mathbf{Z}}^\top,\,\dot{\underline{\boldsymbol{\eta}}}^\top]^\top
\label{eq:109}
\end{align}
evolve according to
\begin{equation}
\label{eq:110}
\begin{aligned}
\dot{\mathbf{x}}_c &= \mathbf{A}_c(\mathbf{x}_c)\mathbf{x}_c + \mathbf{g}_c, \\
\mathbf{x}_c(0) &= [\underline{\mathbf{q}}_0^\top,\, \underline{\boldsymbol{\eta}}_0^\top,\, \underline{\mathbf{Z}}_0^\top,\, \dot{\underline{\boldsymbol{\eta}}}_0^\top]^\top
\end{aligned}
\end{equation}
\begin{equation}
\label{eq:111}
\begin{aligned}
\mathbf{A}_c(\mathbf{x}_c) = \begin{bmatrix}
\mathbf{0} & \mathbf{0} & \mathbf{J}(\underline{\mathbf{q}}) & \mathbf{0} \\
\mathbf{0} & \mathbf{0} & \mathbf{0} & \mathbf{I} \\
\mathbf{0} & \mathbf{0} & \mathbf{0} & \mathbf{0} \\
\mathbf{0} & \mathbf{0} & \mathbf{0} & \mathbf{0}
\end{bmatrix}, \quad
\mathbf{g}_c = \begin{bmatrix}
\mathbf{0} \\ \mathbf{0} \\ \dot{\underline{\mathbf{Z}}}^* \\ \ddot{\underline{\boldsymbol{\eta}}}^*
\end{bmatrix}
\end{aligned}
\end{equation}
where $\mathbf{J}(\underline{\mathbf{q}}) =
\mathrm{diag}(\mathbf{J}(\mathbf{q}_1),\ldots,\mathbf{J}(\mathbf{q}_n))$
is the configuration-dependent Jacobian from~(\ref*{eq:jacobian}).
Since $\mathbf{A}_c$ depends smoothly on $\mathbf{x}_c$ through
$\mathbf{J}(\underline{\mathbf{q}})$ and $\mathbf{g}_c$ consists of
smooth functions of $\mathbf{x}_c$, the right-hand side
of~(\ref*{eq:110}) is locally Lipschitz. By the Picard--Lindelöf
theorem, a unique solution $\mathbf{x}_c(t)$ exists locally in time
and depends continuously on $\mathbf{x}_c(0)$.

Numerical integration of~(\ref*{eq:110}) proceeds stepwise, with
interaction wrenches $\boldsymbol{\mathcal{W}}_i$, body-fixed twists
$\mathbf{V}_i = \mathbf{Z}_i + \boldsymbol{\gamma}_i$, and deformation
field $\mathbf{r}_{\xi_i}(\xi,t)$ recovered at each step via the
post-processing chain summarized in Table~\ref{tab:1}.  Each row applies identically to link $i$ for $i = 1,\dots,n$, reflecting the $O(1)$ per-link assembly cost of the formulation: adding a link appends one block to 
$\mathbf{M}_{\mathrm{sys}}$ and one element $i$ to row descriptions of this table, without modifying existing entries.

\begin{table}[!t]
\renewcommand{\arraystretch}{1.1}
\setlength{\arrayrulewidth}{0.8pt}
\setlength{\tabcolsep}{4pt}
\centering
\begin{tabular}{|p{7.3cm}|}
\hline
\textbf{// initial configuration} \\
\hline
\textbf{i.} input: mechanical and geometric properties \\
\textbf{ii.} input: initial inertial-frame states
${}^{o}\mathbf{q}_i(0)$,
${}^{o}\mathbf{q}_{\xi_i}(0)$,
${}^{o}\mathbf{V}_i(0)$,
${}^{o}\mathbf{V}_{\xi_i}(0)$ \\
\textbf{iii.} input: initial Adjoint $\mathbf{Ad}_{oi}(0)$ from initial configuration \\
\textbf{iv.} transform to body-fixed frame:
${}^{i}\mathbf{q}_i(0)$,
${}^{i}\mathbf{q}_{\xi_i}(0)$,
${}^{i}\mathbf{V}_i(0)$,
${}^{i}\mathbf{V}_{\xi_i}(0)$ \\
\textbf{v.} calculate: $\boldsymbol{\phi}_i$ based on
Eqs.~(\ref*{eq:35}--\ref*{eq:38}) \\
\hline
\textbf{// sample $k$ ($k \geq 1$)} \\
\hline
\textbf{vi.} input:
${}^{i}\mathbf{q}_{i}(k\!-\!1)$,
${}^{i}\mathbf{q}_{\xi_i}(k\!-\!1)$,
${}^{i}\mathbf{V}_{i}(k\!-\!1)$,
${}^{i}\mathbf{V}_{\xi_i}(k\!-\!1)$,
$\mathbf{Ad}_{oi}(k\!-\!1)$ \\
\textbf{vii.} solve: discretized model~(\ref*{eq:98}) to obtain
$\dot{\underline{\mathbf{Z}}}^*(k\!-\!1)$,
$\ddot{\underline{\boldsymbol{\eta}}}^*(k\!-\!1)$,
$\underline{\boldsymbol{\mathcal{W}}}^*(k\!-\!1)$ \\
\textbf{viii.} integrate: update body-fixed states
${}^{i}\mathbf{q}_{i}(k)$,
${}^{i}\mathbf{q}_{\xi_i}(k)$,
${}^{i}\mathbf{V}_{i}(k)$,
${}^{i}\mathbf{V}_{\xi_i}(k)$
via $\dot{\mathbf{q}}_i = \mathbf{J}_i(\mathbf{q}_i)\mathbf{Z}_i$
and $\boldsymbol{\eta}_i$, $\dot{\boldsymbol{\eta}}_i$ from (vii) \\
\textbf{ix.} update: $\mathbf{Ad}_{oi}(k)$ from ${}^{i}\mathbf{q}_i(k)$ \\
\textbf{x.} transform to inertial frame:
${}^{o}\mathbf{q}_{i}(k)$,
${}^{o}\mathbf{q}_{\xi_i}(k)$,
${}^{o}\mathbf{V}_{i}(k)$,
${}^{o}\mathbf{V}_{\xi_i}(k)$
using $\mathbf{Ad}_{oi}(k)$ \\
\hline
\textbf{xi.} $k = k + 1$, goto \textbf{(vi)} \\
\hline
\end{tabular}
\caption{\textbf{Separation-of-variables solution algorithm for the
PDE-based flexible multibody dynamic model.}}
\label{tab:1}
\end{table}

\section{Results and discussion} \label{sec:7}

 The modeling framework developed in this paper is general and applicable to different subtypes of multibody flexible systems with motion in three-dimensional inertial space. Various systems may be investigated by expansion, reduction, and modification of the proposed framework for specific scenarios, or implementing different constraints. In this section, to demonstrate physical validity of the framework, a representative implementation example is presented, regarding a main direct application of the work, which is dynamic analysis of serial two-link flexible manipulator with revolute joints operating in three-dimensional space. In this example, the synthesis model is numerically implemented and the results are verified based on measurements of the corresponding experimental setup, Elastic Lightweight Laboratory Arm (ELLA), developed in Institute of Robotics, Johannes Kepler University Linz~\cite{Kitzinger2026}. The experimental setup consists of two flexible links, three actuated revolute joints, three permanent magnet DC motors, and 
an endpoint mass as payload. The manipulator operates in 
three-dimensional space, with motion permitted about the 
rotation axes of the revolute joints. Two revolute joints 
actuating the base link about its $y$- and $z$-axes are modeled 
as a single two-degree-of-freedom joint with configuration space 
$SO(2)\times SO(2)$, and the remaining revolute joint actuating 
the second link about its body-fixed $z$-axis is modeled as a 
single degree-of-freedom joint with configuration space $SO(2)$. The assigned configuration of joints implies that in the joint constraint projection matrix $\mathbf{P}$, the term $\mathbf{I}_{3} - {}^{o}\mathbf{A}_{1}\,{}^{o}\mathbf{A}_{1}^{\top}$ 
is equal to $\mathrm{diag}(0, 1, 1)$ and $\mathbf{I}_{3} - {}^{o}\mathbf{A}_{2}\,{}^{o}\mathbf{A}_{2}^{\top}$ is equal to $\mathrm{diag}(0, 0, 1)$. 

\begin{figure*}[htbp]
\centering

\subfloat[]{
\includegraphics[width=0.45\textwidth]{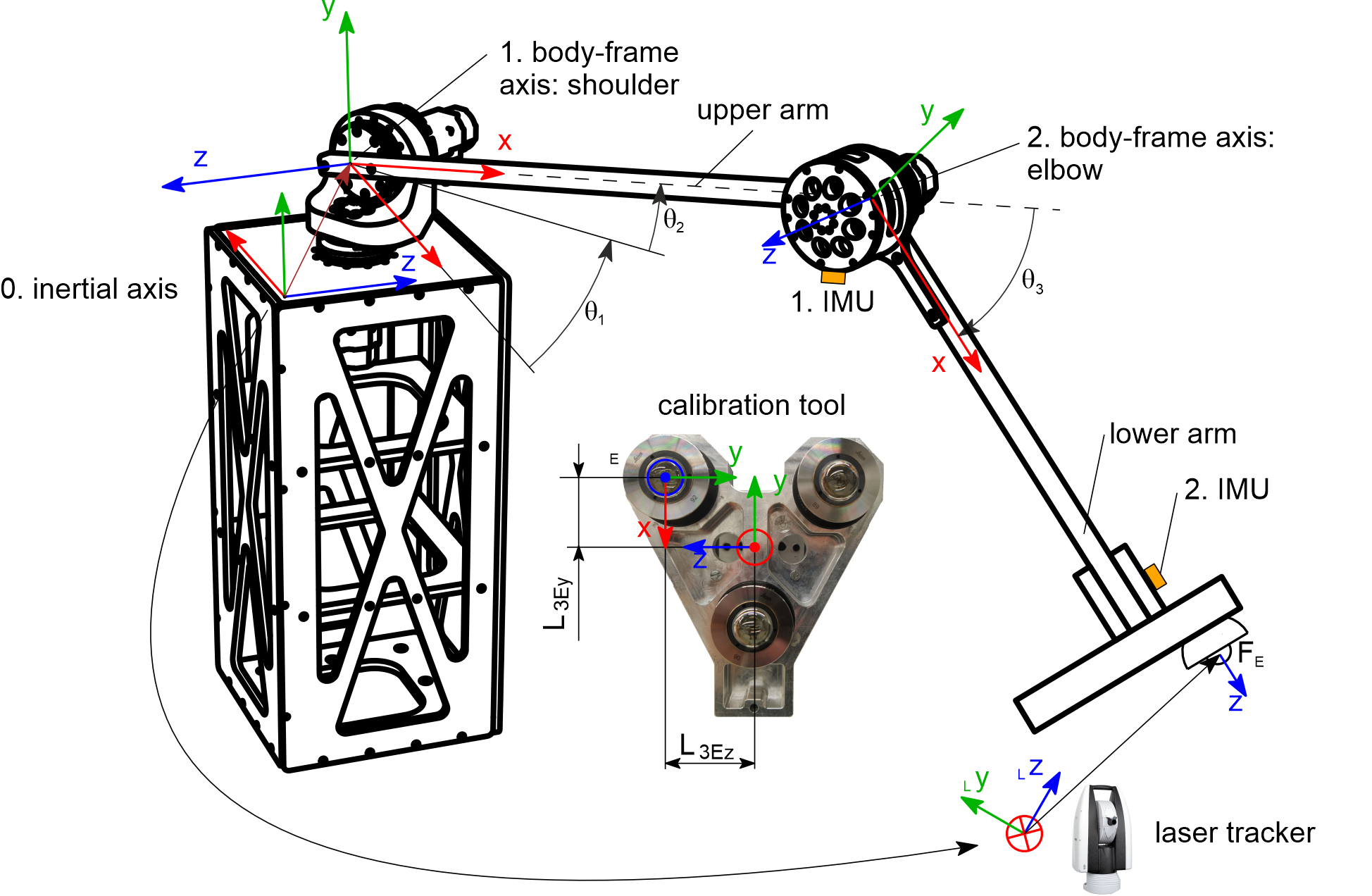}
\label{fig_expa}}
\hfill
\subfloat[]{
\includegraphics[width=0.45\textwidth]{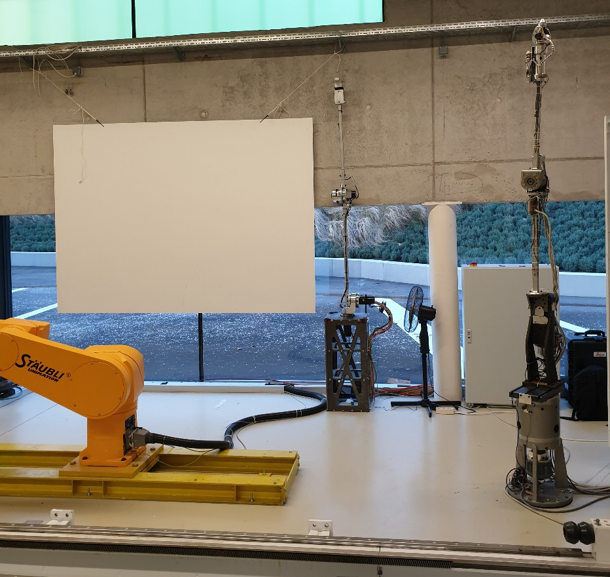}
\label{figexp}}

\caption{Elastic Lightweight Laboratory Arm setup,  Institute of Robotics, Johannes Kepler University Linz.  (a) Schematic description of the setup (b) Experimental setup in laboratory environment.}
\label{fig_exp}
\end{figure*}

The experimental setup used in this verification example is shown in 
Fig.~\ref{figexp}.

The implementation of the model is conducted considering the mechanism configuration described in Fig.~\ref{fig_2}, and the modeling and synthesis framework presented in Sections~\ref{sec:3}-\ref{sec:6}. 
In this implementation example, the inertial frame is located at the base of the rigid link and shares the same origin as the body-fixed frame ${}^1\!\mathbf{\mathcal{F}}$. The body-fixed frame ${}^2\!\mathbf{\mathcal{F}}$ is placed at the connecting joint between the two links. This placement of body-fixed frames was selected for convenience compatibility with standard robotics conventions. It is notable that in the presented modeling framework, the body-fixed frames may be arbitrarily placed at any point in the centerline of flexible body, at the cost of alternate measurement requirements. The mechanical properties of the beam are assigned as $E I_z = 3.67 \times 10^3 \, [\mathrm{N} \cdot \mathrm{m}^2], \, E I_y = 2.12 \times 10^3 \, [\mathrm{N} \cdot \mathrm{m}^2], \, \text{and} \, E A = 4.38 \times 10^7 \, [\mathrm{N}]$. Length of the flexible link $i=1.00 [\mathrm{m}]$ is $l_1=0.95 [\mathrm{m}]$, with uniform surface area $A = 2.09 \times 10^{-4} [\mathrm{m}^2]$ and the length of the second link is $l_2 = 0.93 [\mathrm{m}]$. Motor masses are $3.3 [kg]$, and payload of $2.1 [kg]$ is included at the endpoint of second link.

Synchronized with the experimental 
measurements, the simulation begins with the first flexible link in an undeformed 
configuration as $^{0}\mathbf{r}_{\xi_1}(a_i, t{=}0) = \mathbf{0}$, while the 
second link has initial endpoint displacement 
$^{0}\mathbf{r}_{\xi_2}(b_i, t{=}0) = \begin{bmatrix} -0.02,\ 0.0065,\ -0.0083 \end{bmatrix}^\top\ [\mathrm{m}]$. 
Inputs to the system are generated based on a PD control law---exerted in the form 
of joint torques---as

\begin{align}
\boldsymbol{\tau}_{\mathrm{ext}B_i} &= \mathrm{diag}(k_{p_1}, k_{p_2}, k_{p_2}) 
(\boldsymbol{\theta}_{di} - \boldsymbol{\theta}_i) \notag\\ 
&\quad + \mathrm{diag}(k_{d_1}, k_{d_2}, k_{d_2}) 
(\boldsymbol{\omega}_{di} - \boldsymbol{\omega}_i)
\end{align}

In this implementation, the desired trajectories $\theta_{di}$ and $\omega_{di}$ 
are assigned for endpoint tracking of system response for controllable states 
$\theta_{y_1}$, $\theta_{z_1}$, and $\theta_{z_2}$, given a setpoint for each 
state of $\theta_{y_1}(t_f) = -0.4928$, $\theta_{z_1}(t_f) = 0.6922$, and 
$\theta_{z_2}(t_f) = 0.4270$, and optimization-based trajectory generation~\cite{Staufer2012}. 
The desired angular velocity trajectory is generated based on the time-derivative of the desired joint angle trajectory and its configuration Jacobian.

Shape functions in the separation-of-variables implementation are assigned in harmonic form for a homogeneous clamped-free beam~\cite{Rao2007}, in accordance with Sturm-Liouville completeness theorem~\cite{zettl2005}. The axial, lateral, and out-of-plane bending eigenfunction are respectively given by

\begin{align}
    \phi_x(\xi_i) &= a_x \frac{\pi}{2L} \cos\left(\frac{\pi}{2L}\xi_i\right) 
    \label{eq:phi_x} \\
    \phi_y(\xi_i) &= a_y \beta_y \Big[ \sinh(\beta_y \xi_i) + \sin(\beta_y \xi_i) \notag\\ 
    &\quad - r_{\phi_y}\left(\cosh(\beta_y \xi_i) - \cos(\beta_y \xi_i)\right) \Big] 
    \label{eq:phi_y} \\
    \phi_z(\xi_i) &= a_z \beta_z \Big[ \sinh(\beta_z \xi_i) + \sin(\beta_z \xi_i) \notag\\ 
    &\quad - r_{\phi_z}\left(\cosh(\beta_z \xi_i) - \cos(\beta_z \xi_i)\right) \Big]
    \label{eq:phi_z}
\end{align}

\noindent where $\beta_y$ and $\beta_z$ are the modal wavenumbers for bending in the 
$y$ and $z$ planes respectively. $a_x$, $a_y$, and $a_z$ are 
normalization constants. $r_{\phi_y}$ and 
$r_{\phi_z}$ are the mode shape ratios from the clamped-free boundary conditions:

\begin{equation}
    r_{\phi} = \frac{\sinh(\beta L) - \sin(\beta L)}{\cosh(\beta L) + \cos(\beta L)}
\end{equation}

\noindent The wavenumber $\beta_n$ relates to the $n$-th natural frequency $f_n$ by

\begin{equation}
    \beta_n^2 = 2\pi f_n \sqrt{\frac{\rho A}{EI}}
\end{equation}

\noindent

All measurements potentially include errors, unstructured uncertainty, and inevitable incompatibility of specific details between setup configurations and the model--for example measurement calibrations, joint offsets, hidden stiffnesses and damping effects, transfer function uncertainty, and mechatronic characteristics--which are not removed as inputs or initial conditions of the model and are implemented as is. For error elimination, adaptive closed-loop systems may be designed and implemented in future studies.

The numerical solution of PDE model is obtained using the implementation algorithm detailed in Table.~\ref{tab:1}, which is based on the dynamic model expressed in Section~\ref{sec:3} and the separation of variables-based discretization detailed in Section~\ref{sec:4}. On this basis, system states in body-fixed frame —and correspondingly in inertial frame—are calculated at each sample. A new body-fixed frame is then established based on the dynamic response of the model at the investigated instance, and the relation between inertial and body-fixed frame is established for the subsequent simulation sample. This process is iterated for the duration of the numerical simulations to obtain system response in the simulation interval. 

The modal ODE system~(\ref{eq:98}) is integrated 
using the forward Euler method with fixed time step 
$T_s = 10^{-3}$ [s]. The numerical calculation stability for the clamped-free 
flexible beam~\cite{Rao2007} with $n$ retained modes requires

\begin{align}
T_s \leq \frac{2}{\omega_{\max}} = 
\frac{2l^2}{(1.8751)^2 n^2}\sqrt{\frac{\rho A}{EI}}
\label{eq:stability}
\end{align}

For the flexible links parameters ($E = 2.1\times10^{11}$ 
[Pa], $A = 2.09 \times 10^{-4} [\mathrm{m}^2]$, $\rho = 7800$ [kg/m$^3$], $n = 1$), the bound evaluates to $T_s \leq 10.04\times10^{-3}$ [s] based on 
the strongest bending direction ($EI_y = 
3.67\times10^3$ [Nm$^2$]) and shortest link ($l_2 = 0.93$ [m]), which is the critical constraint. The simulation time step $T_s = 10^{-3}$ [s] 
satisfies this bound with a factor of 10.04 safety 
margin, confirming numerical stability. That correct 
and stable results are obtained with a simple 
first-order explicit integrator — without requiring 
implicit or higher-order methods — 
directly reflects the well-posedness and numerical 
regularity of the proposed formulation, as the 
model produces a 
well-conditioned ODE system whose solution does not 
require specialized numerical treatment for stable 
integration.

\begin{figure*}[htbp]
\centering

\subfloat[]{
\includegraphics[width=0.48\textwidth]{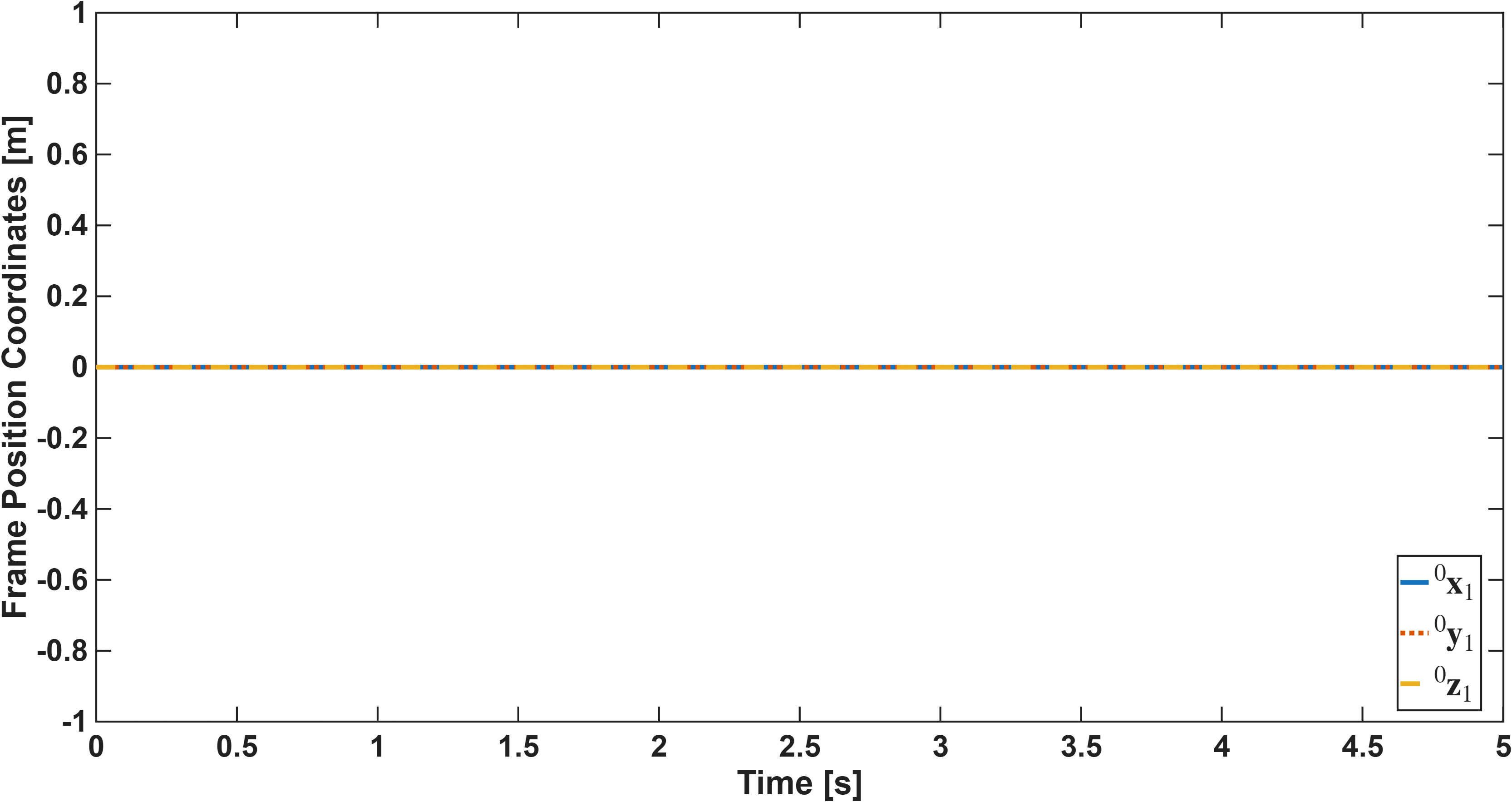}
\label{fig:4a}}
\hfill
\subfloat[]{
\includegraphics[width=0.48\textwidth]{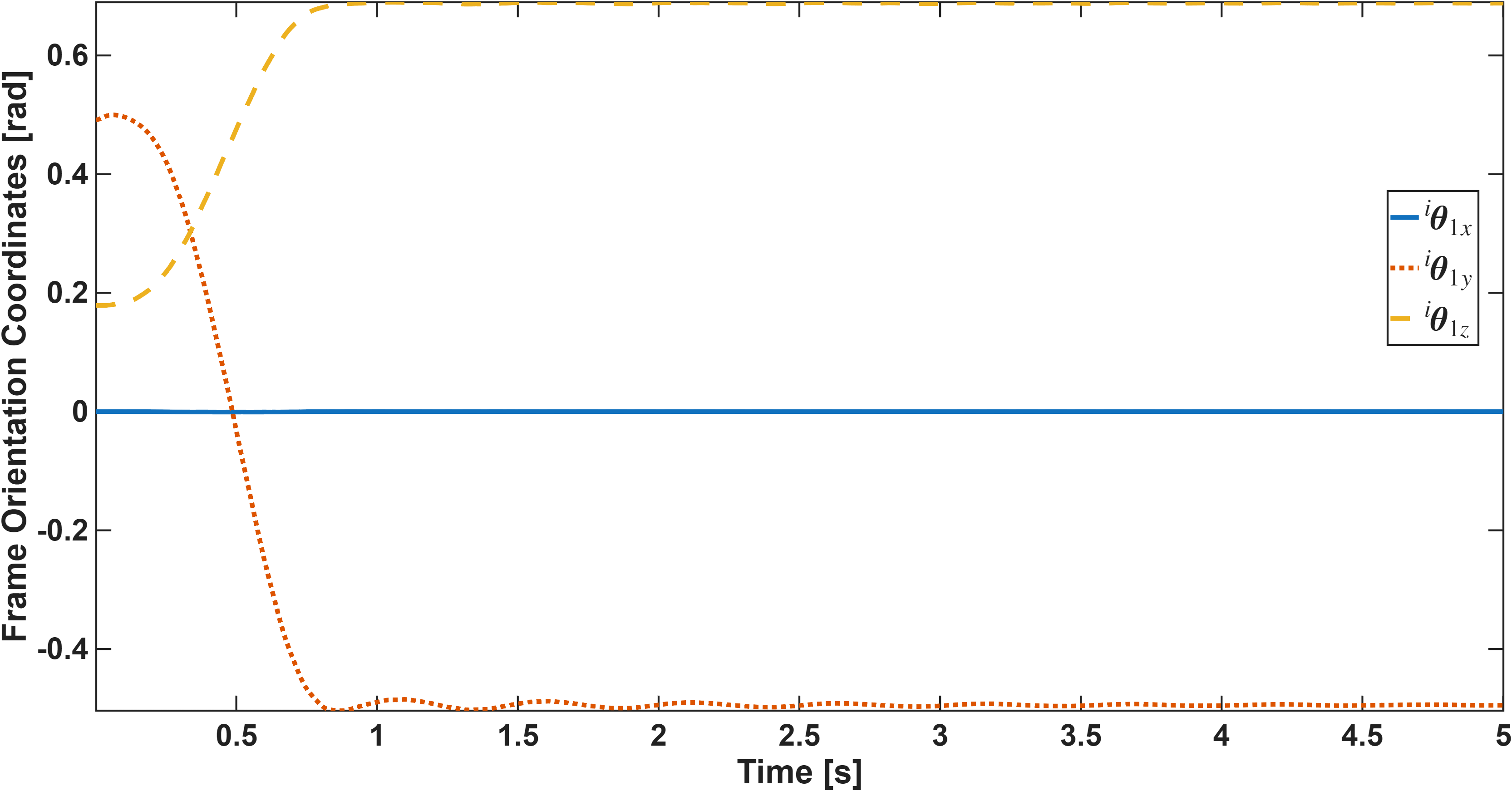}
\label{fig:4b}}

\vspace{1em}

\subfloat[]{
\includegraphics[width=0.48\textwidth]{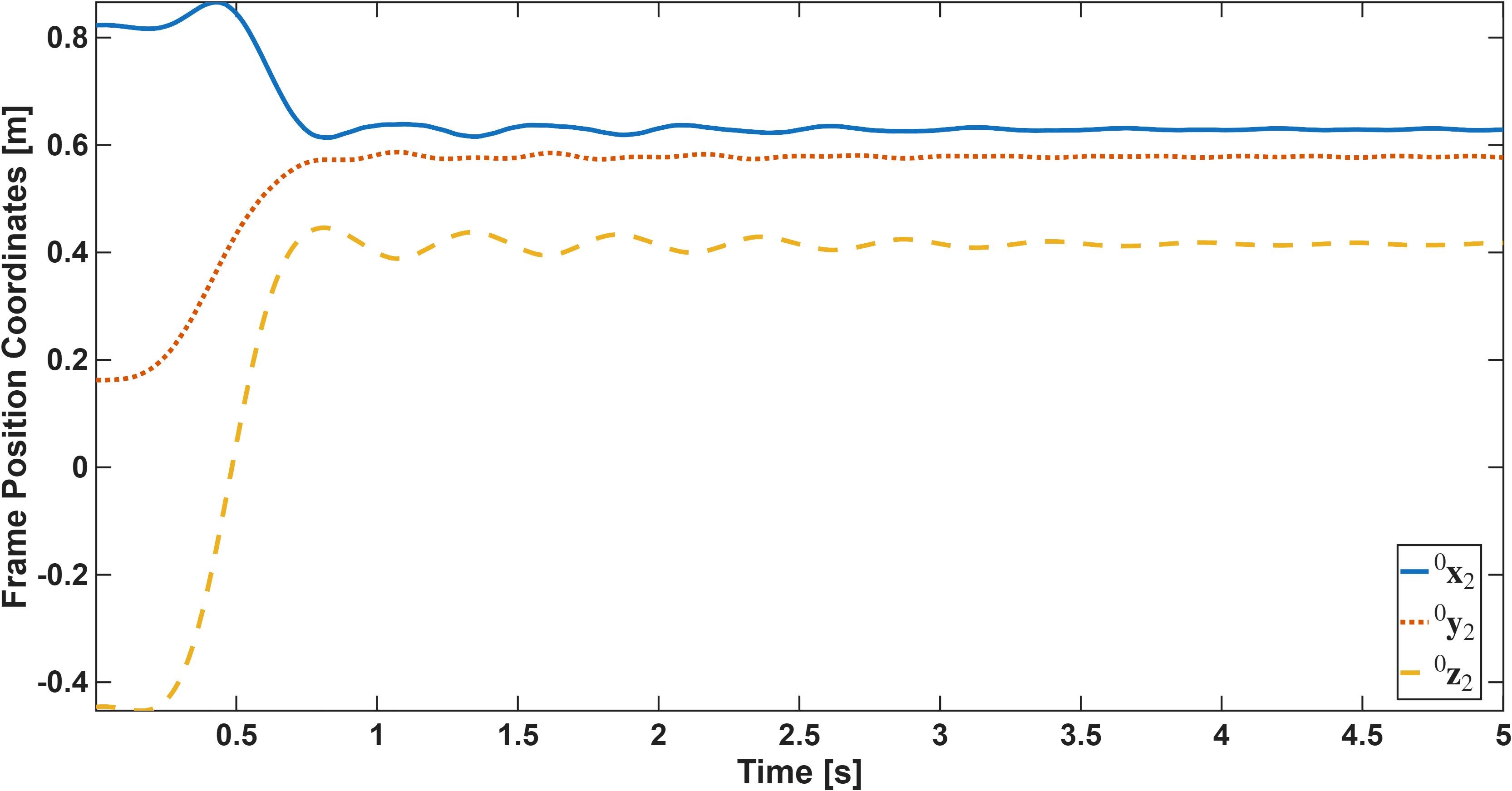}
\label{fig:4c}}
\hfill
\subfloat[]{
\includegraphics[width=0.48\textwidth]{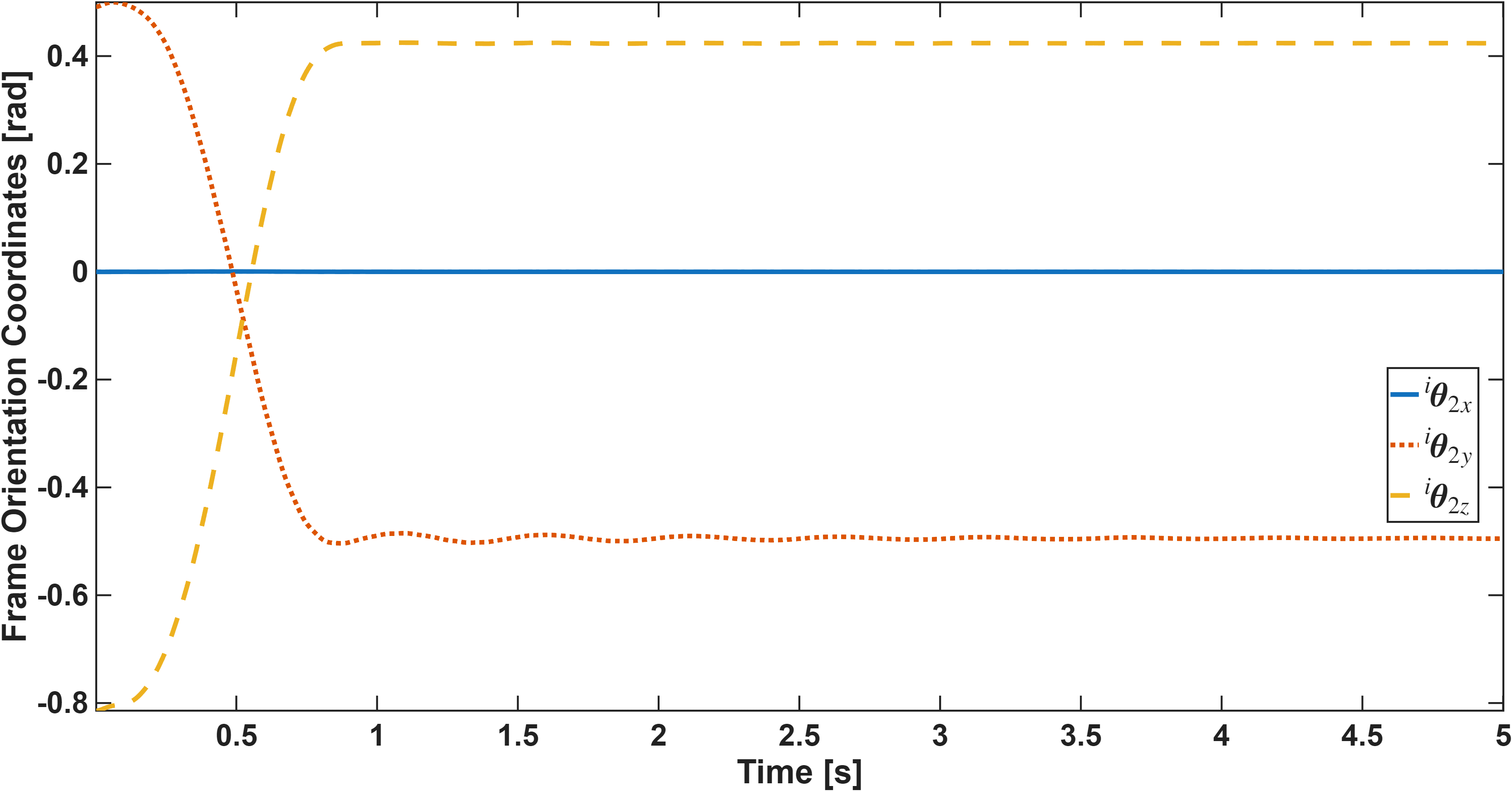}
\label{fig:4d}}

\caption{Coordinates vector components for each link: 
(a) $^{i}\mathbf{r}_{1}$, (b) $^{i}\theta_{1}$, (c) $^{i}\mathbf{r}_{2}$, (d) $^{i}\theta_{2}$.}
\label{fig_4}
\end{figure*}

\begin{figure*}[htbp]
\centering

\subfloat[]{
\includegraphics[width=0.48\textwidth]{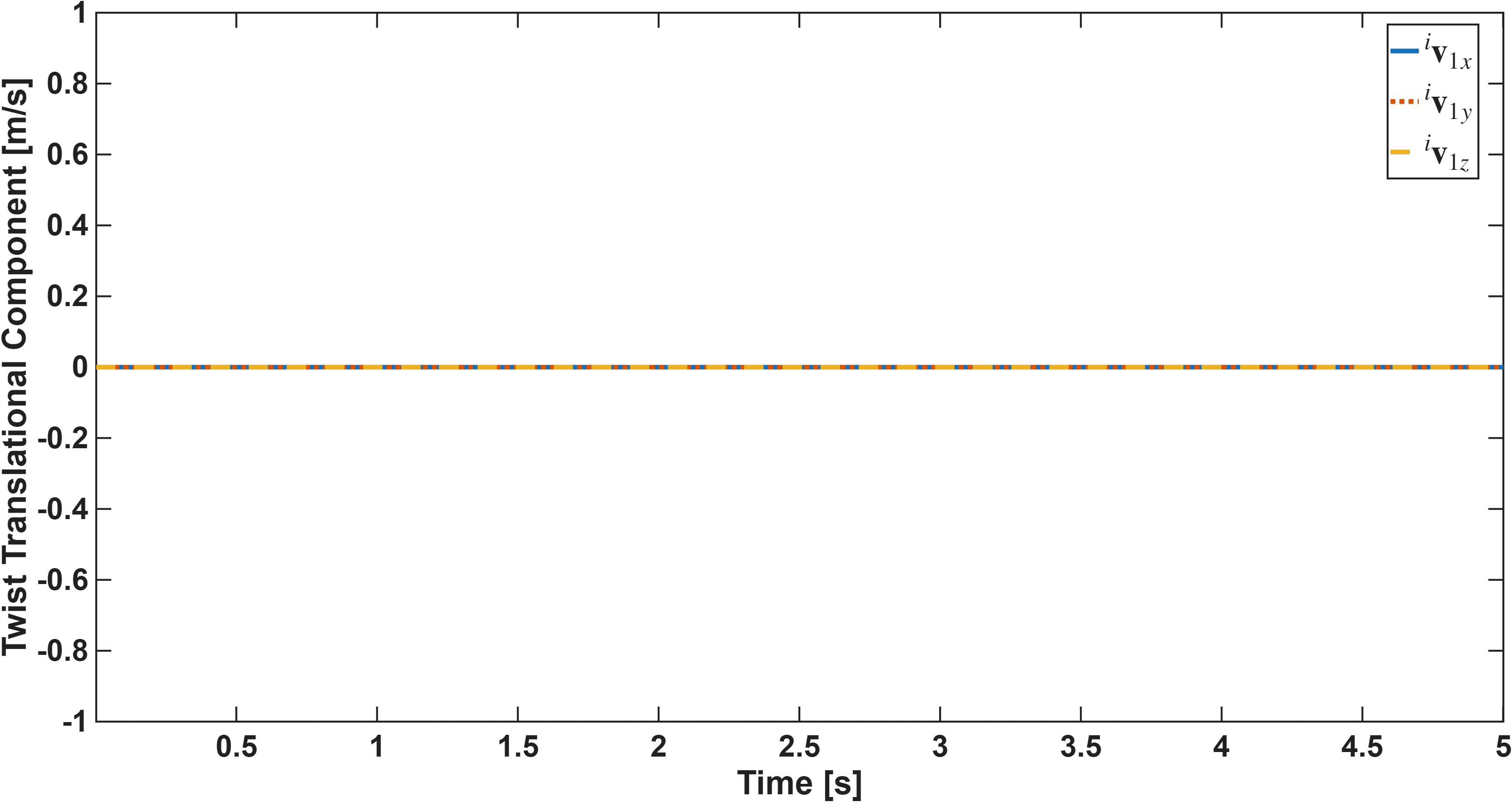}
\label{fig:6a}}
\hfill
\subfloat[]{
\includegraphics[width=0.48\textwidth]{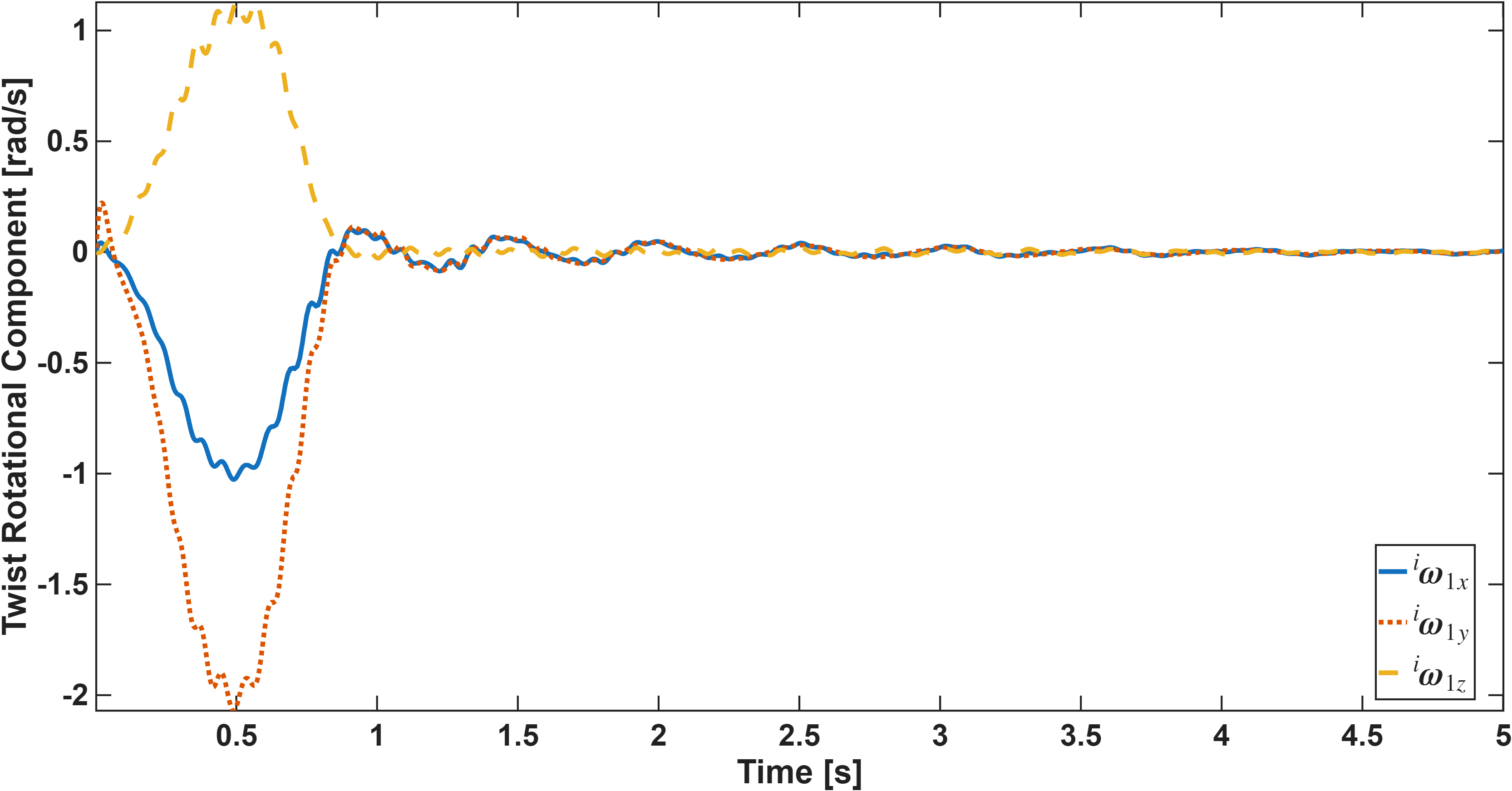}
\label{fig:6b}}

\vspace{1em}

\subfloat[]{
\includegraphics[width=0.48\textwidth]{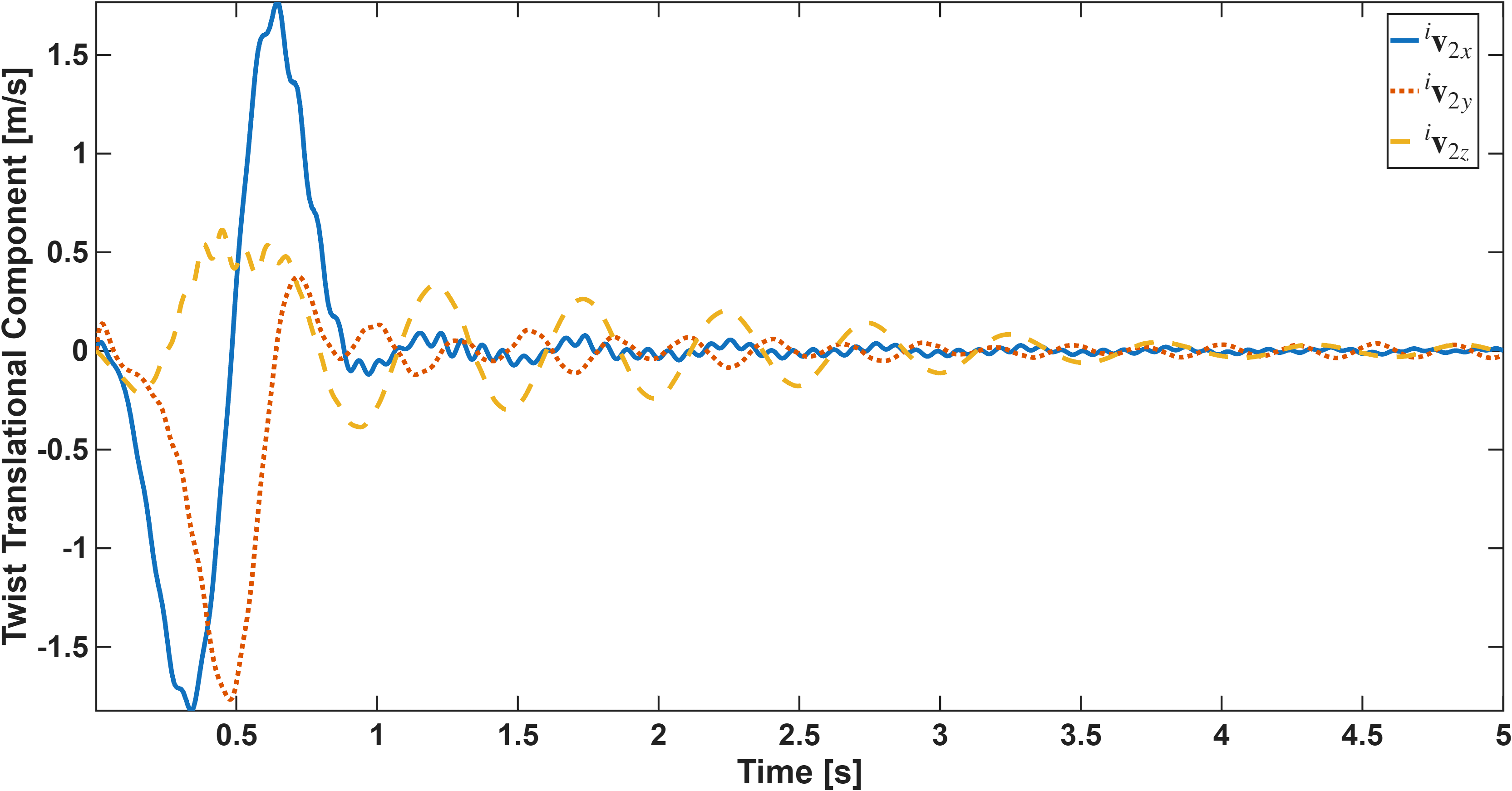}
\label{fig:6c}}
\hfill
\subfloat[]{
\includegraphics[width=0.48\textwidth]{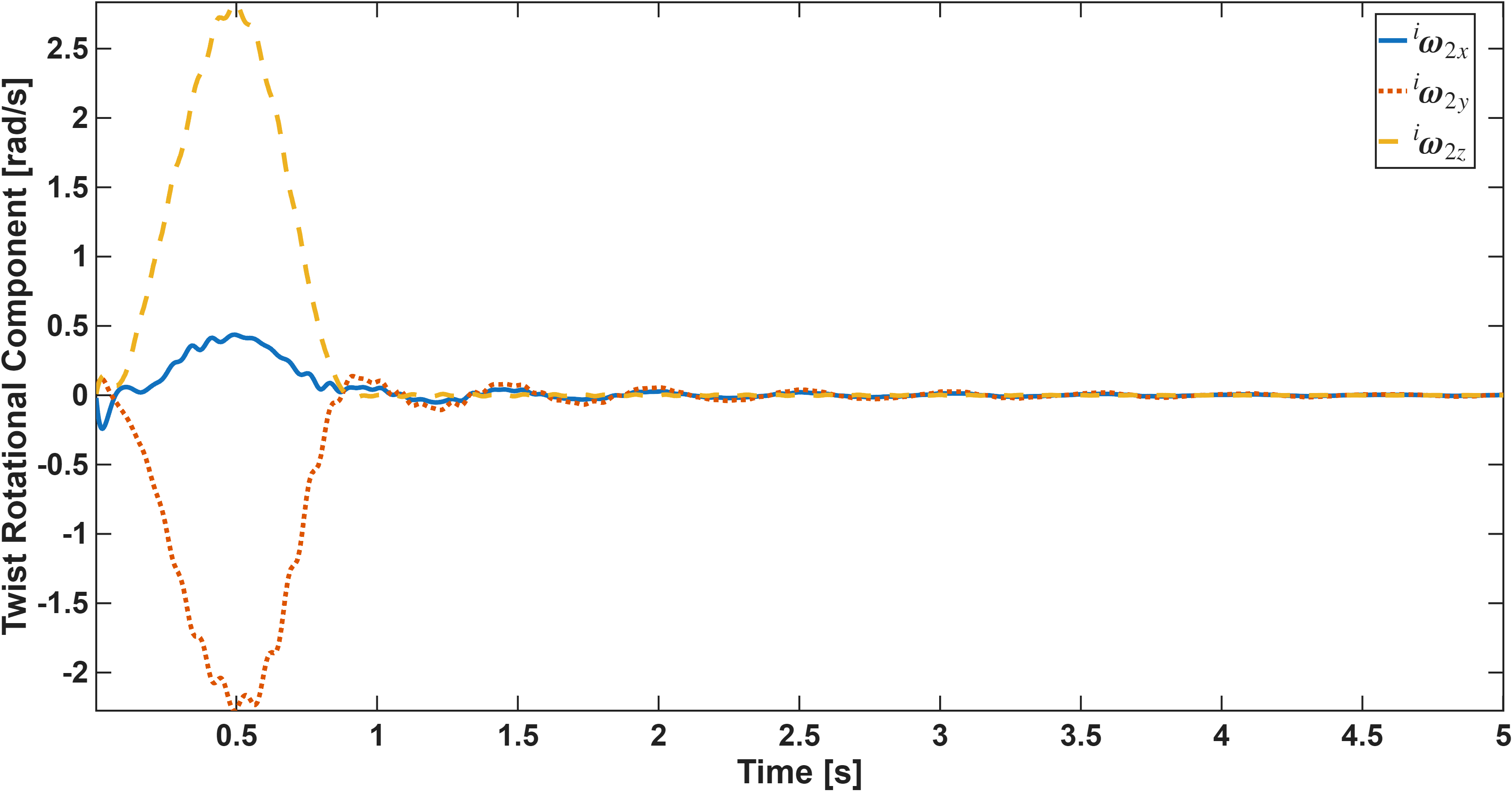}
\label{fig:6d}}

\caption{Twist components for each link: 
(a) $^{i}\mathbf{v}_{1}$, (b) $^{i}\omega_{1}$, (c) $^{i}\mathbf{v}_{2}$, (d) $^{i}\omega_{2}$.}
\label{fig_6}
\end{figure*}

\begin{figure*}[htbp]
\centering

\subfloat[]{
\includegraphics[width=0.47\textwidth]{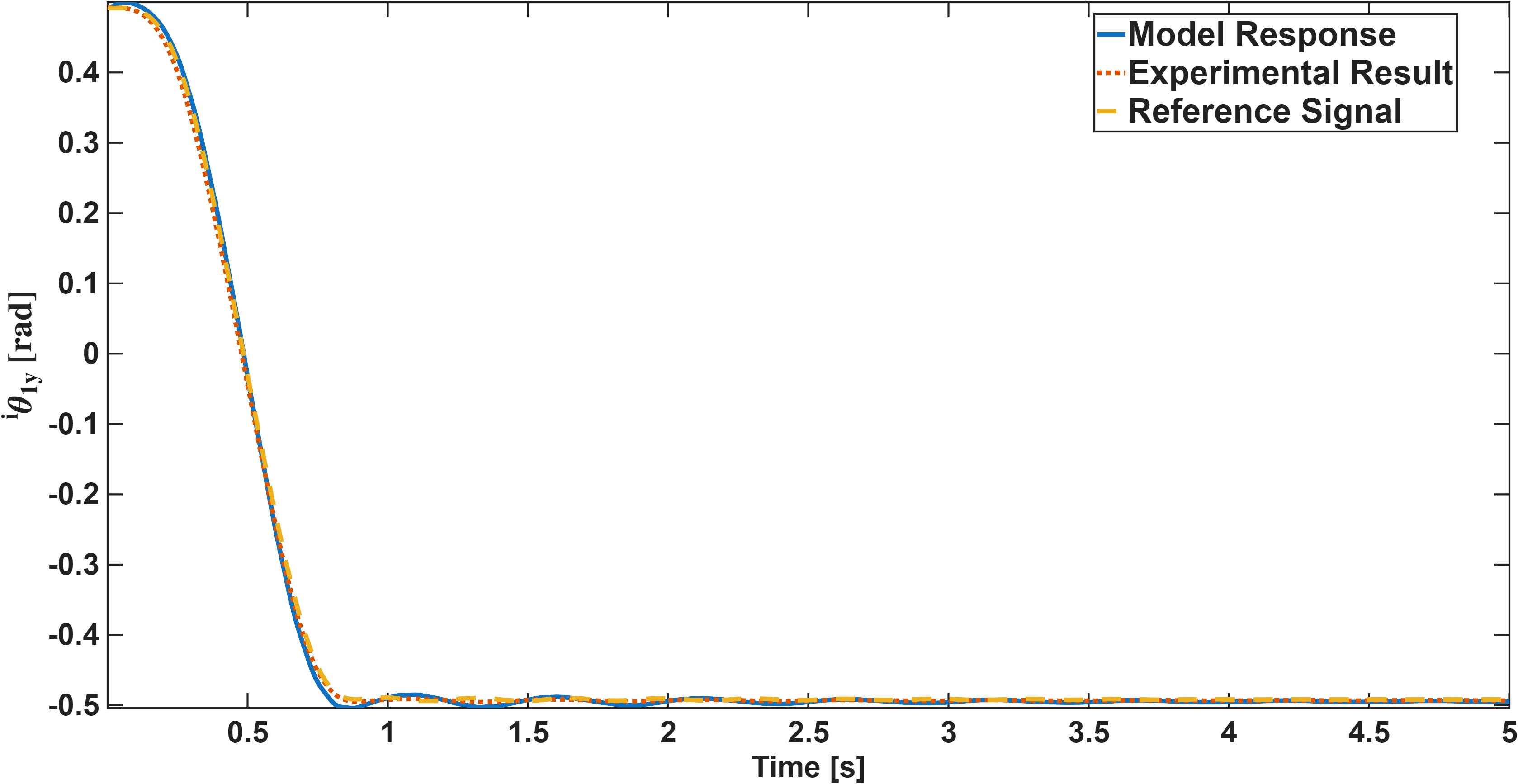}
\label{fig:3a}}
\hfill
\subfloat[]{
\includegraphics[width=0.47\textwidth]{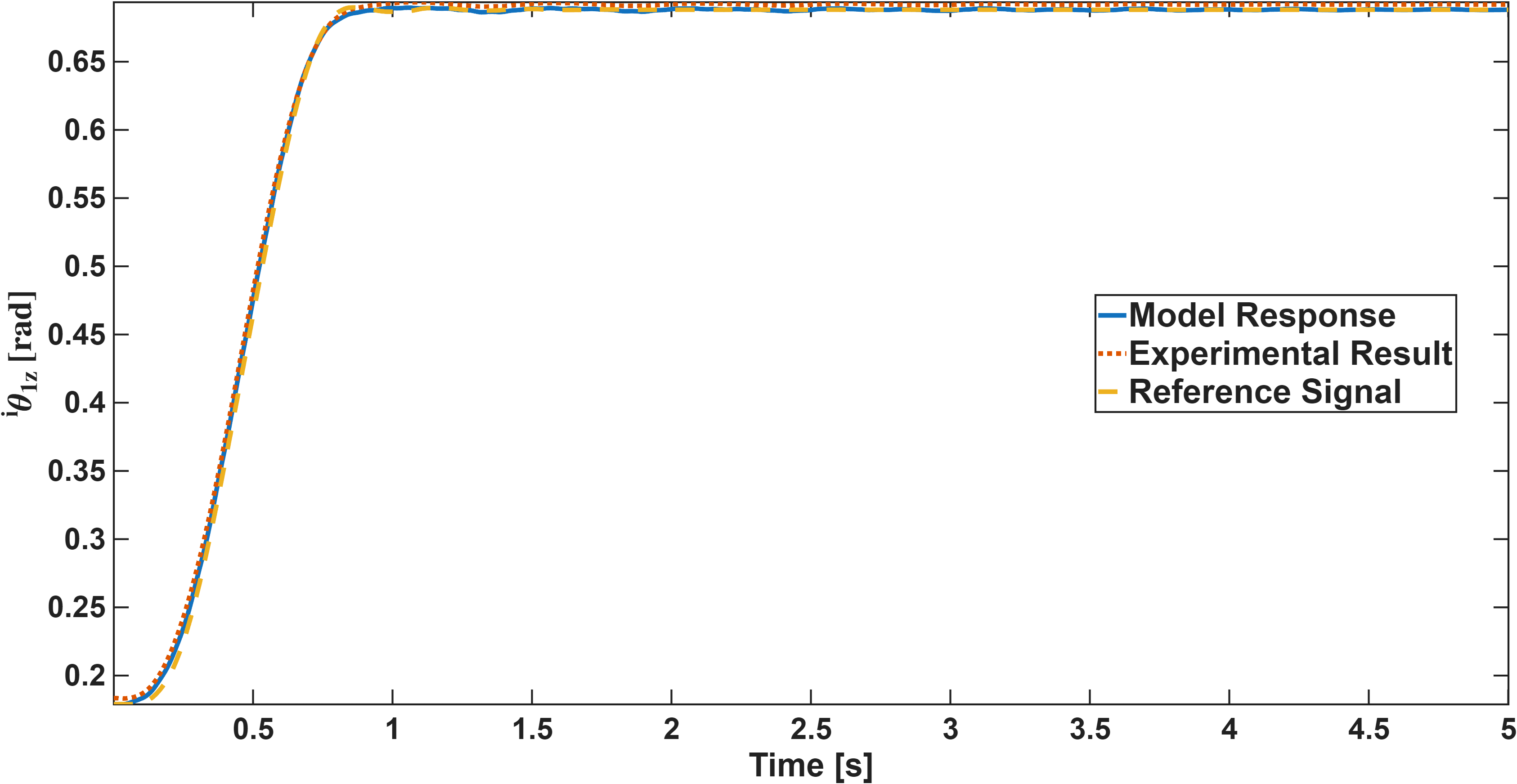}
\label{fig:3b}}

\vspace{1em}

\subfloat[]{
\includegraphics[width=0.47\textwidth]{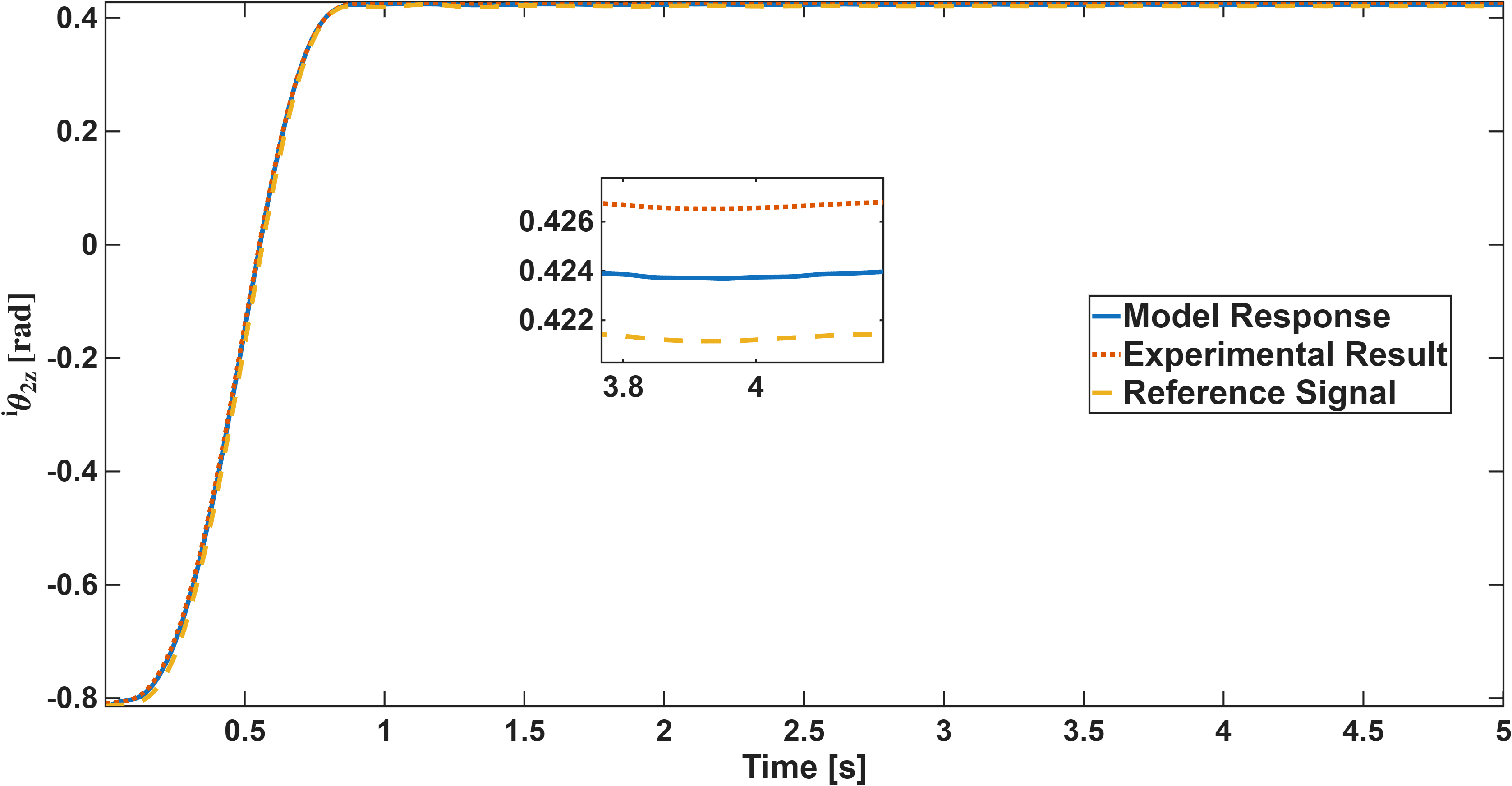}
\label{fig:3c}}
\caption{Controlled joint angle responses under PD input: comparison between the dynamic model and experimental measurements (a) $^{i}\theta_{y_{1}}$ (b) $^{i}\theta_{z_{1}}$ (c)  $^{i}\theta_{z_{2}}$.}
\label{fig_3}
\end{figure*}

\begin{figure*}[htbp]
\centering

\subfloat[]{
\includegraphics[width=0.45\textwidth]{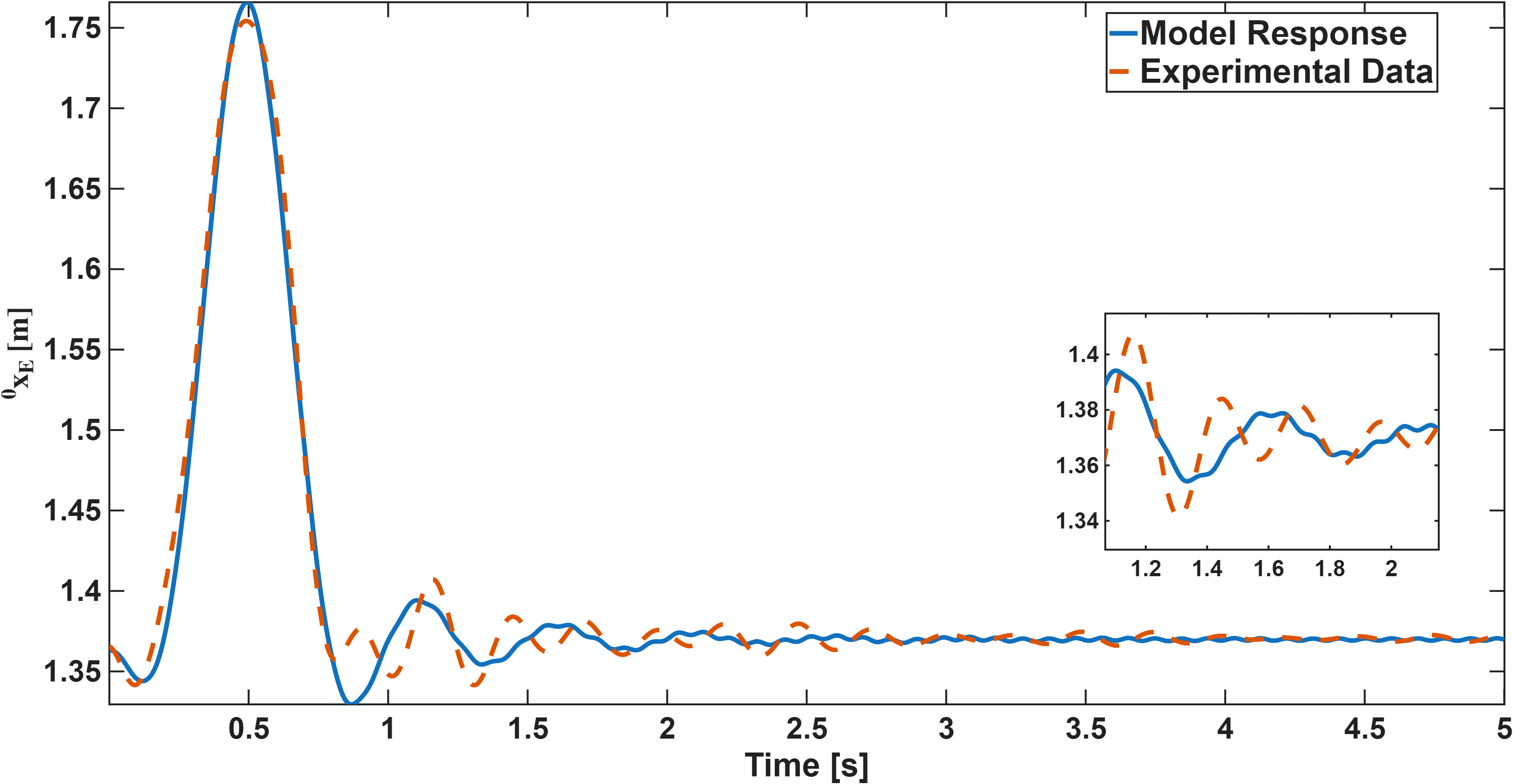}
\label{fig:7a}}
\hfill
\subfloat[]{
\includegraphics[width=0.45\textwidth]{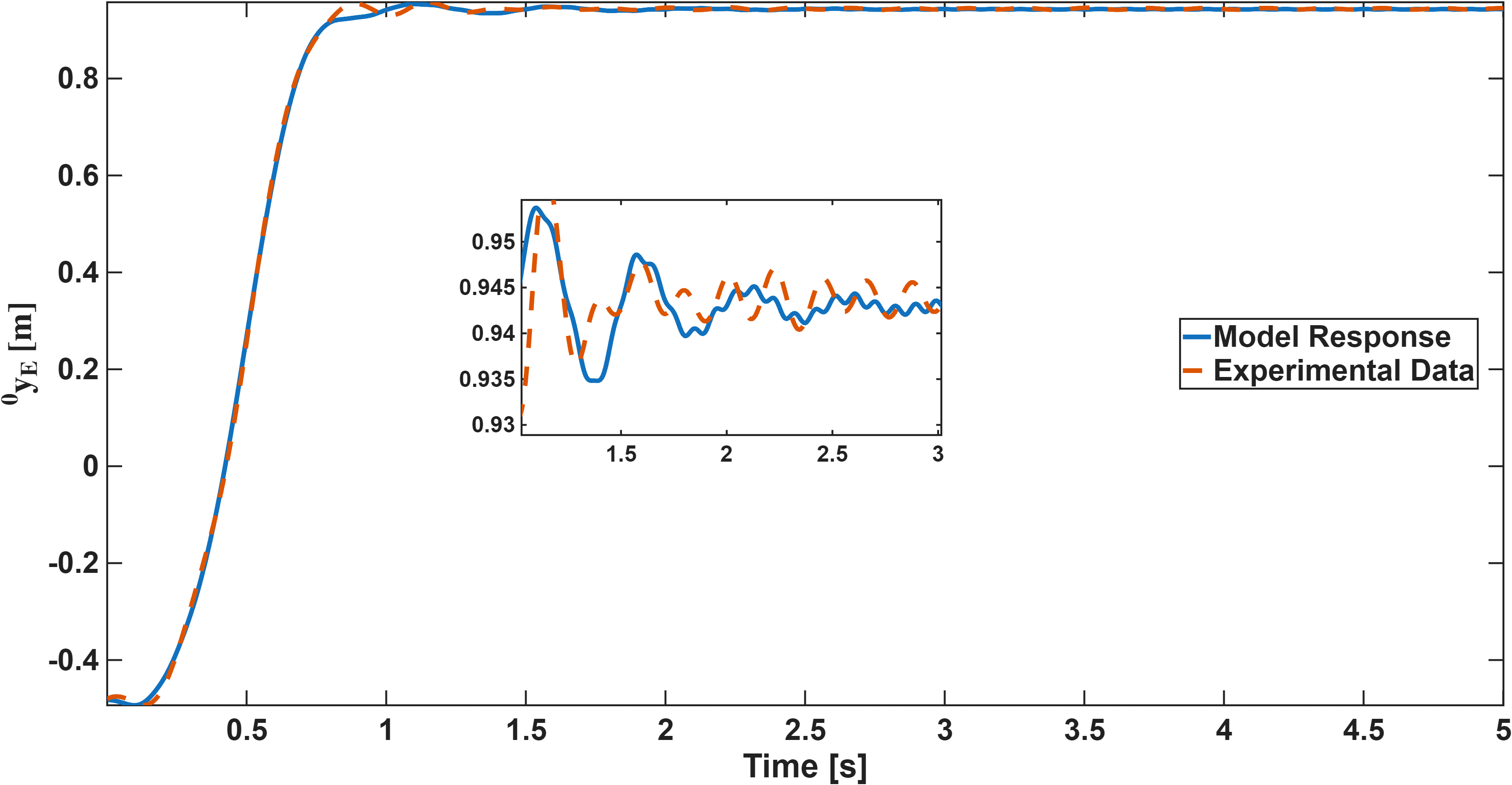}
\label{fig:7b}}

\vspace{1em}

\subfloat[]{
\includegraphics[width=0.45\textwidth]{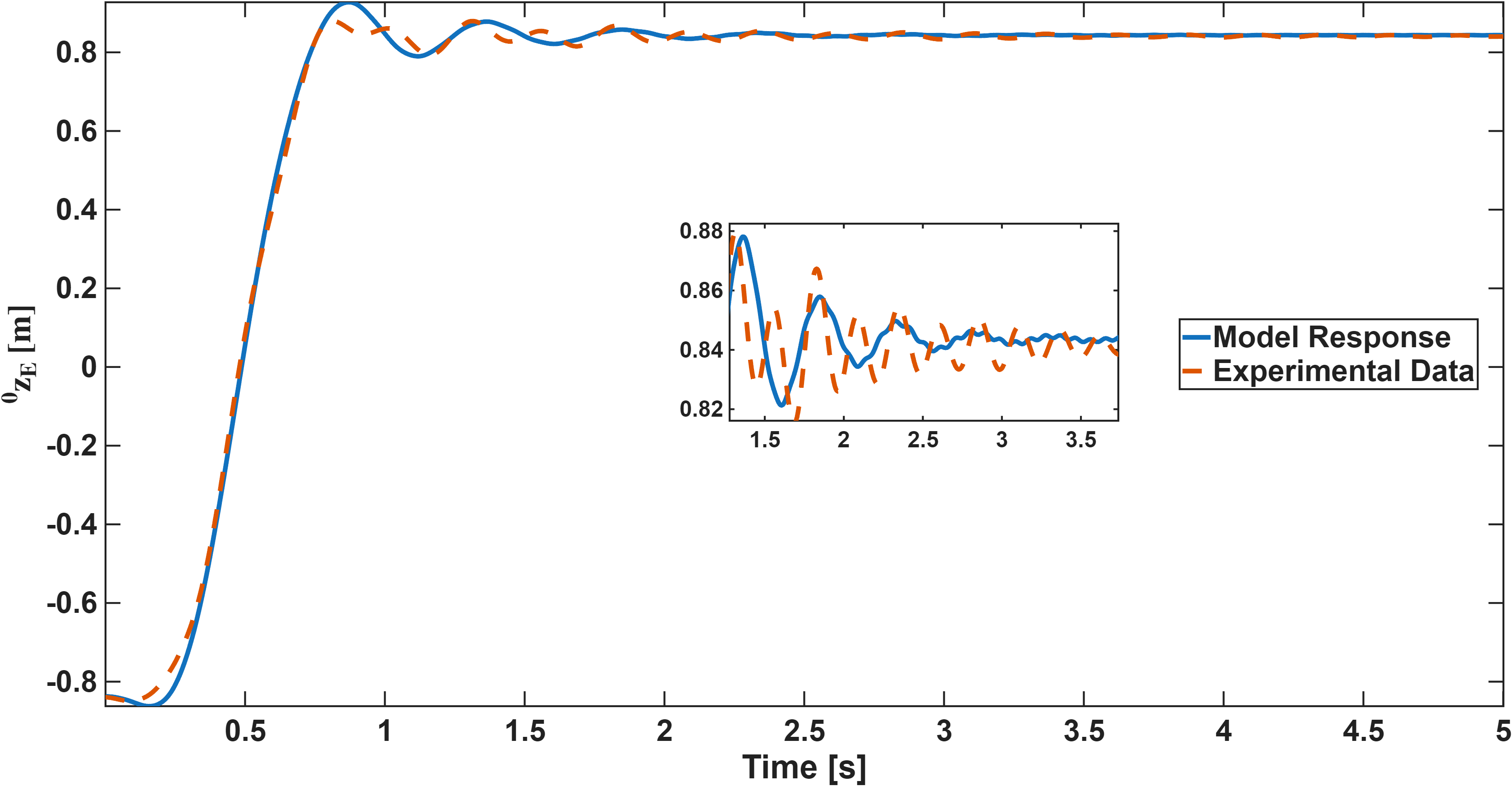}
\label{fig:7c}}
\caption{Endpoint position responses: comparison between the dynamic model and experimental measurements (a) $^{0}x_{E}$ (b) $^{0}y_{E}$ (c)  $^{0}z_{E}$.}
\label{fig_7}
\end{figure*}

\begin{figure*}[htbp]
\centering

\subfloat[]{
\includegraphics[width=0.45\textwidth]{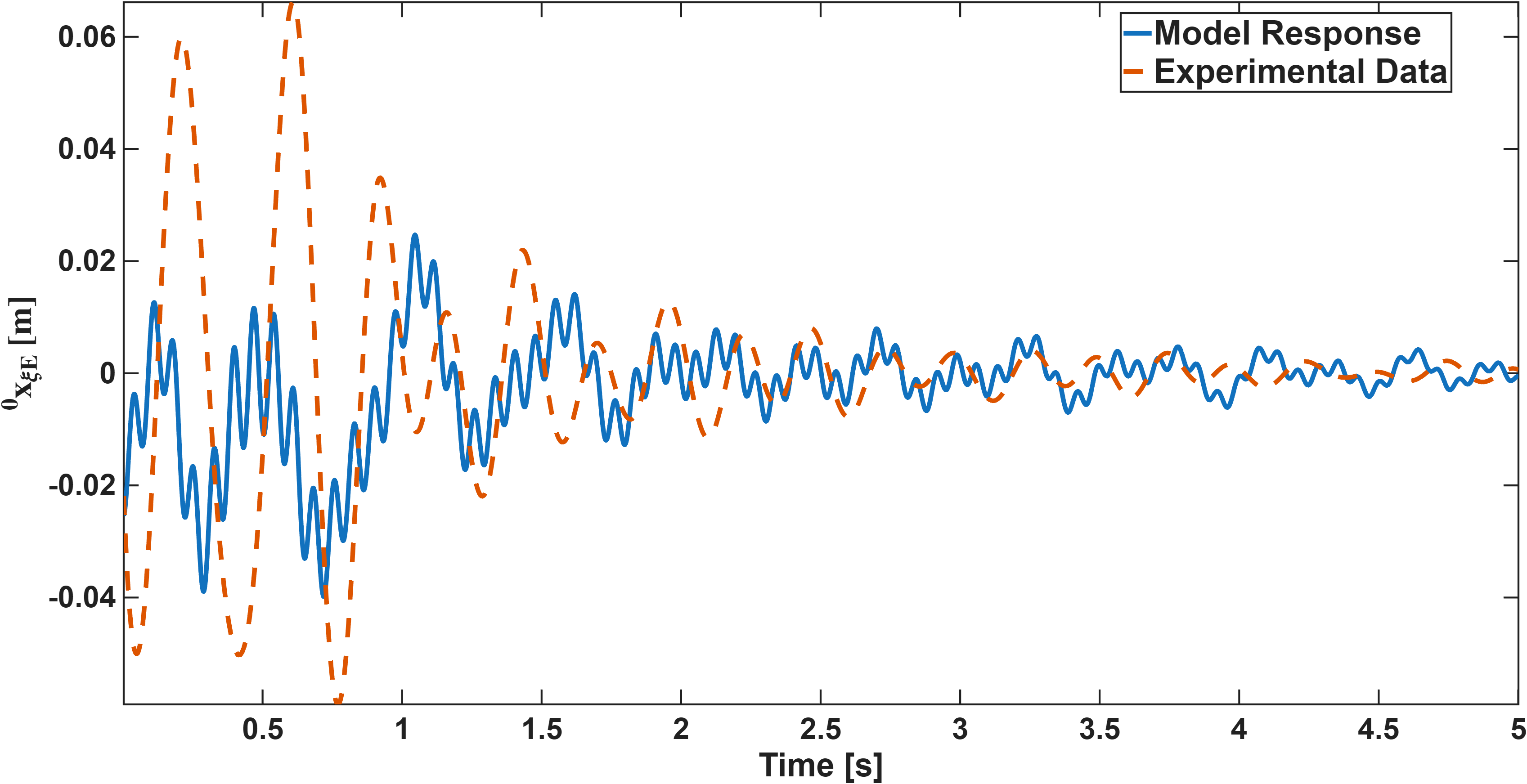}
\label{fig:8a}}
\hfill
\subfloat[]{
\includegraphics[width=0.45\textwidth]{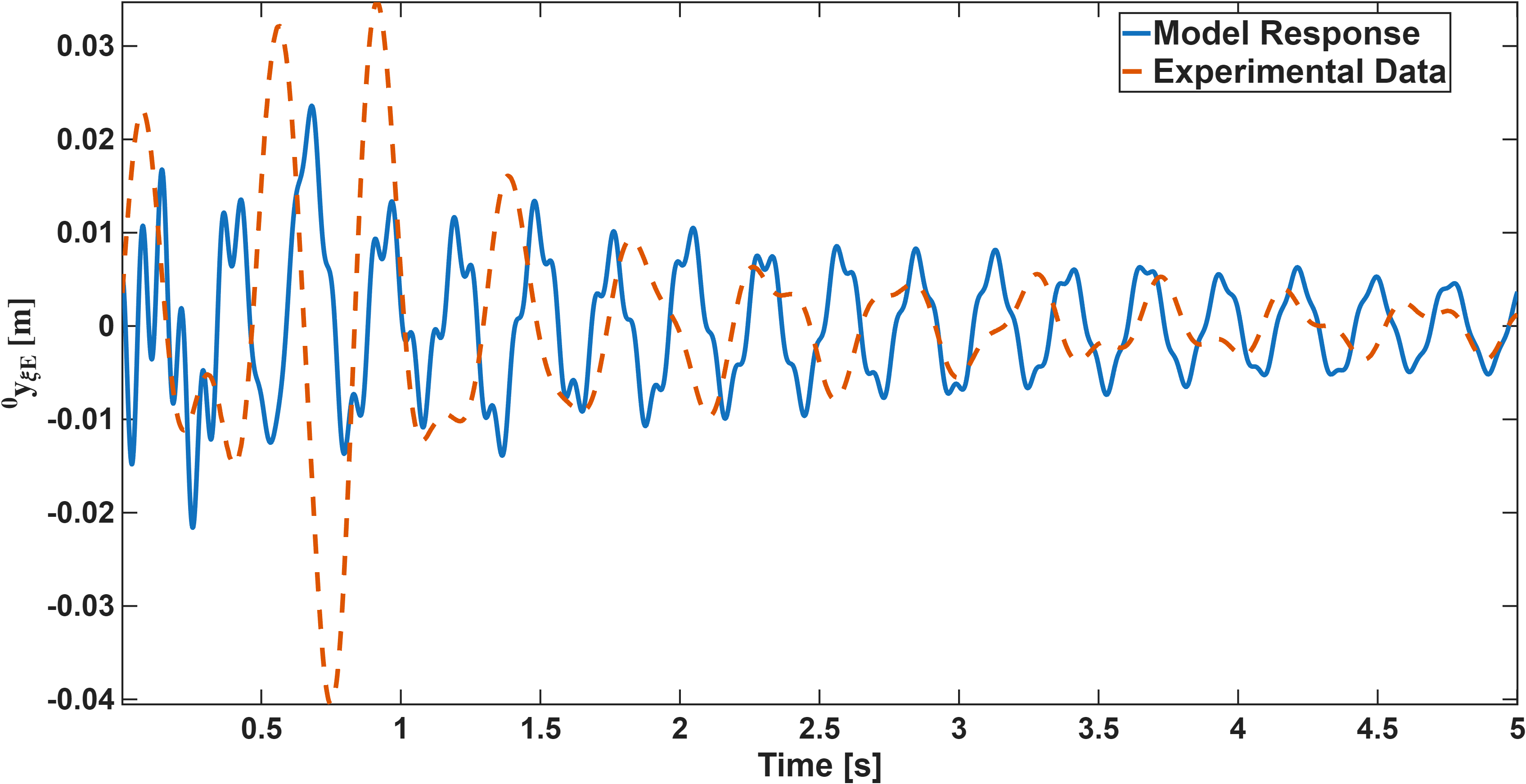}
\label{fig:8b}}

\vspace{1em}

\subfloat[]{
\includegraphics[width=0.45\textwidth]{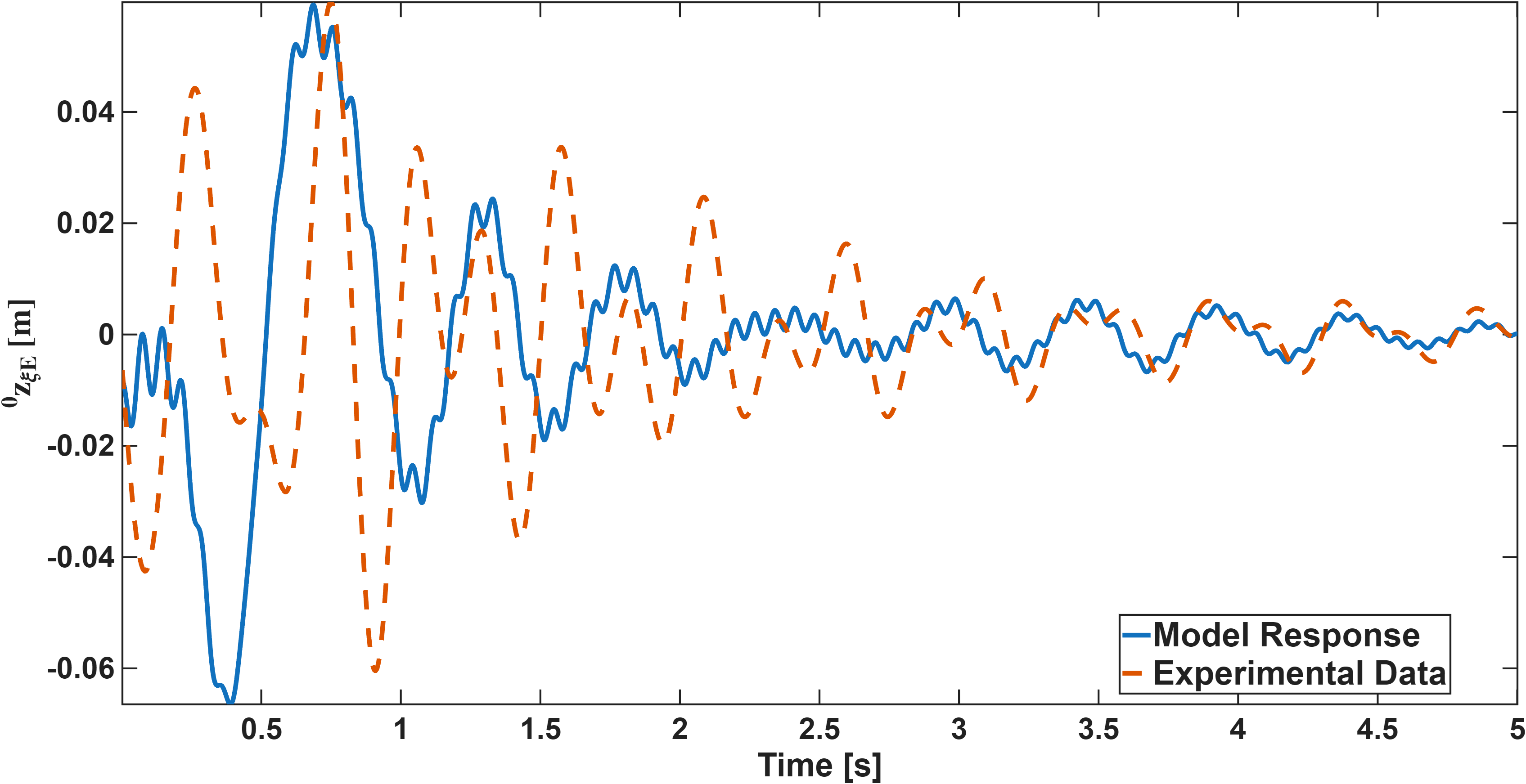}
\label{fig:8c}}
\caption{Endpoint displacement responses: comparison between the dynamic model and experimental measurements (a) $^{0}x_{\xi E}$ (b) $^{0}y_{\xi E}$ (c)  $^{0}z_{\xi E}$.}
\label{fig_8}
\end{figure*}

\begin{figure*}[htbp]
\centering

\subfloat[]{
\includegraphics[width=0.45\textwidth]{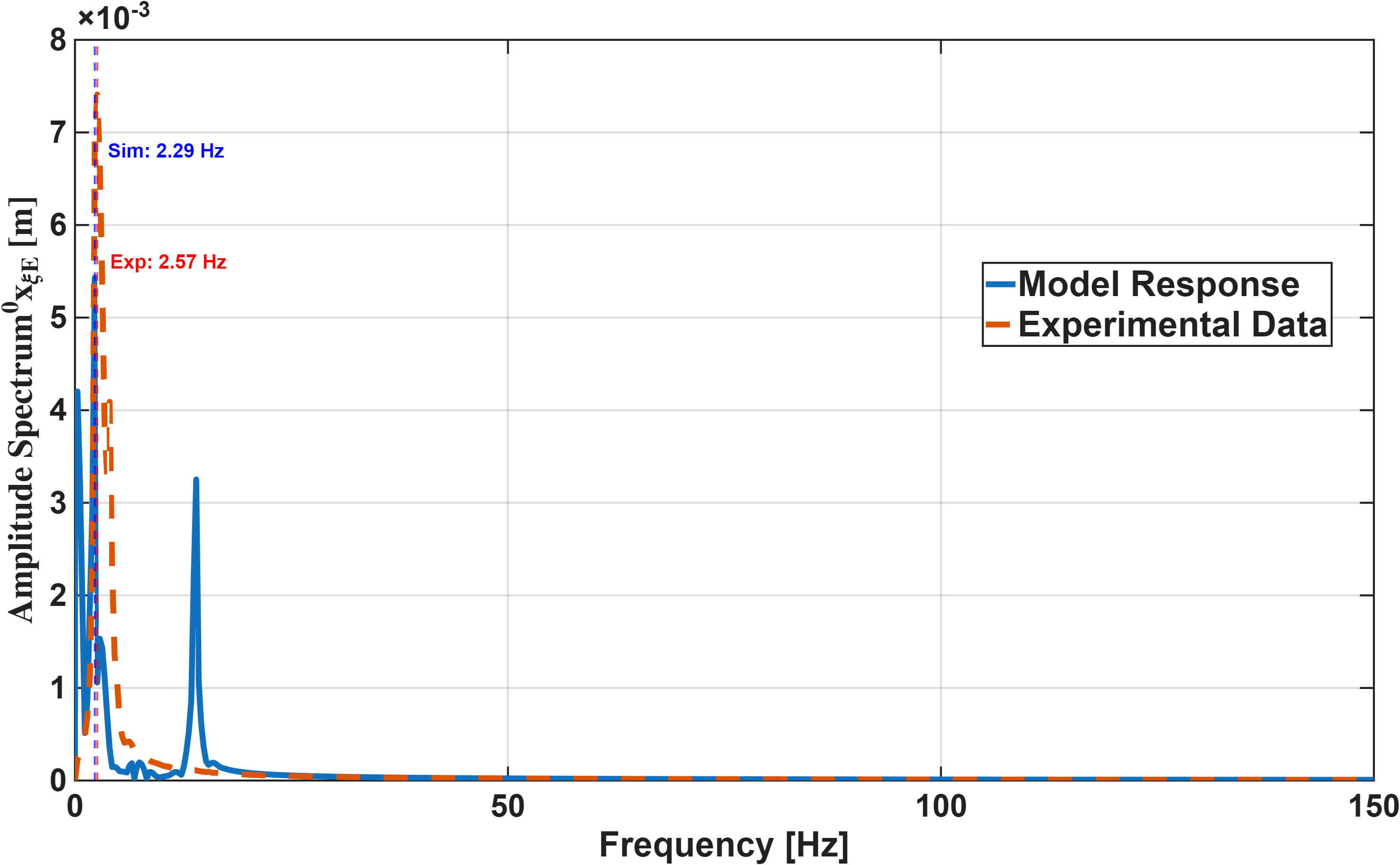}
\label{fig:speca}}
\hfill
\subfloat[]{
\includegraphics[width=0.45\textwidth]{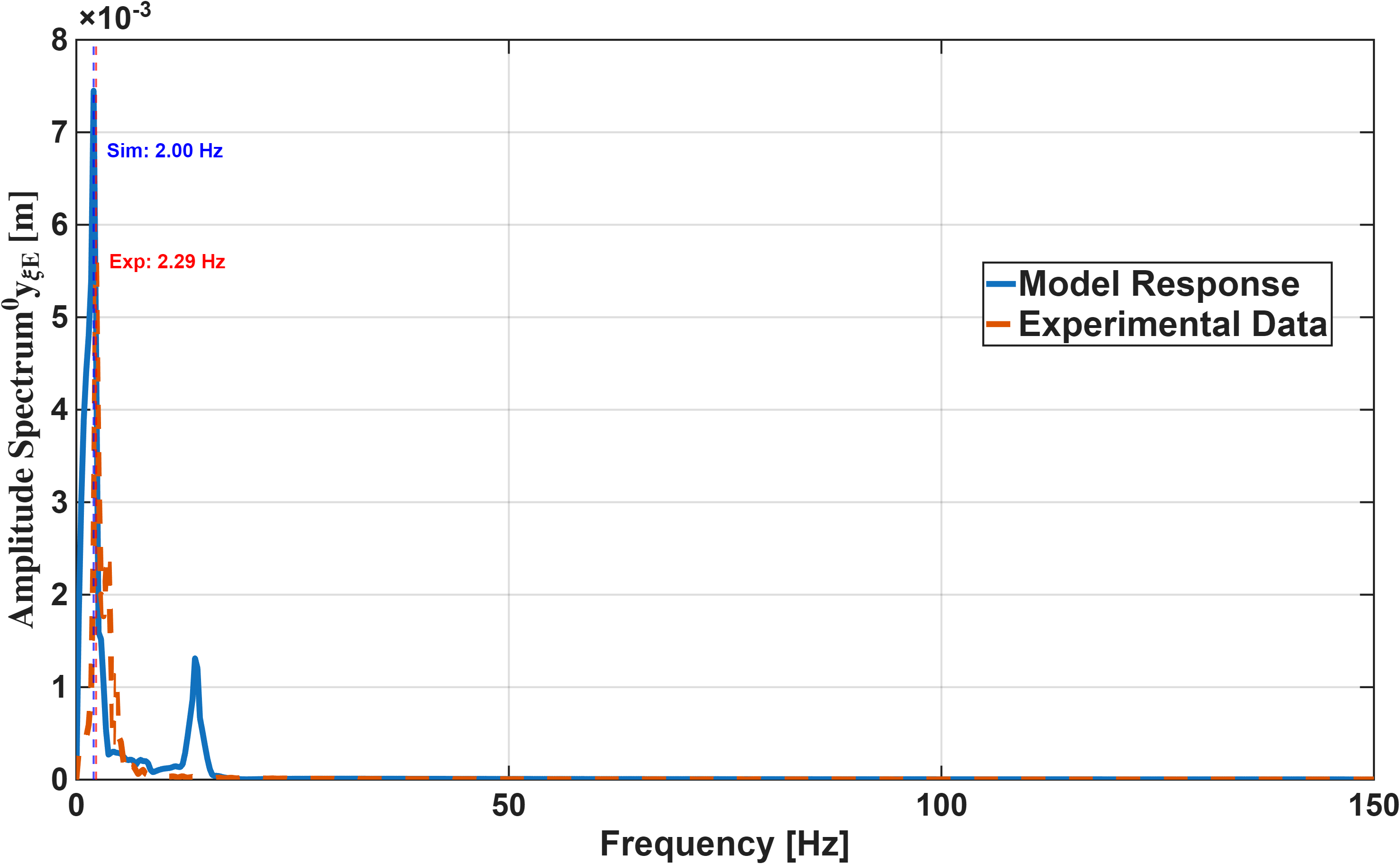}
\label{fig:specb}}

\vspace{1em}

\subfloat[]{
\includegraphics[width=0.45\textwidth]{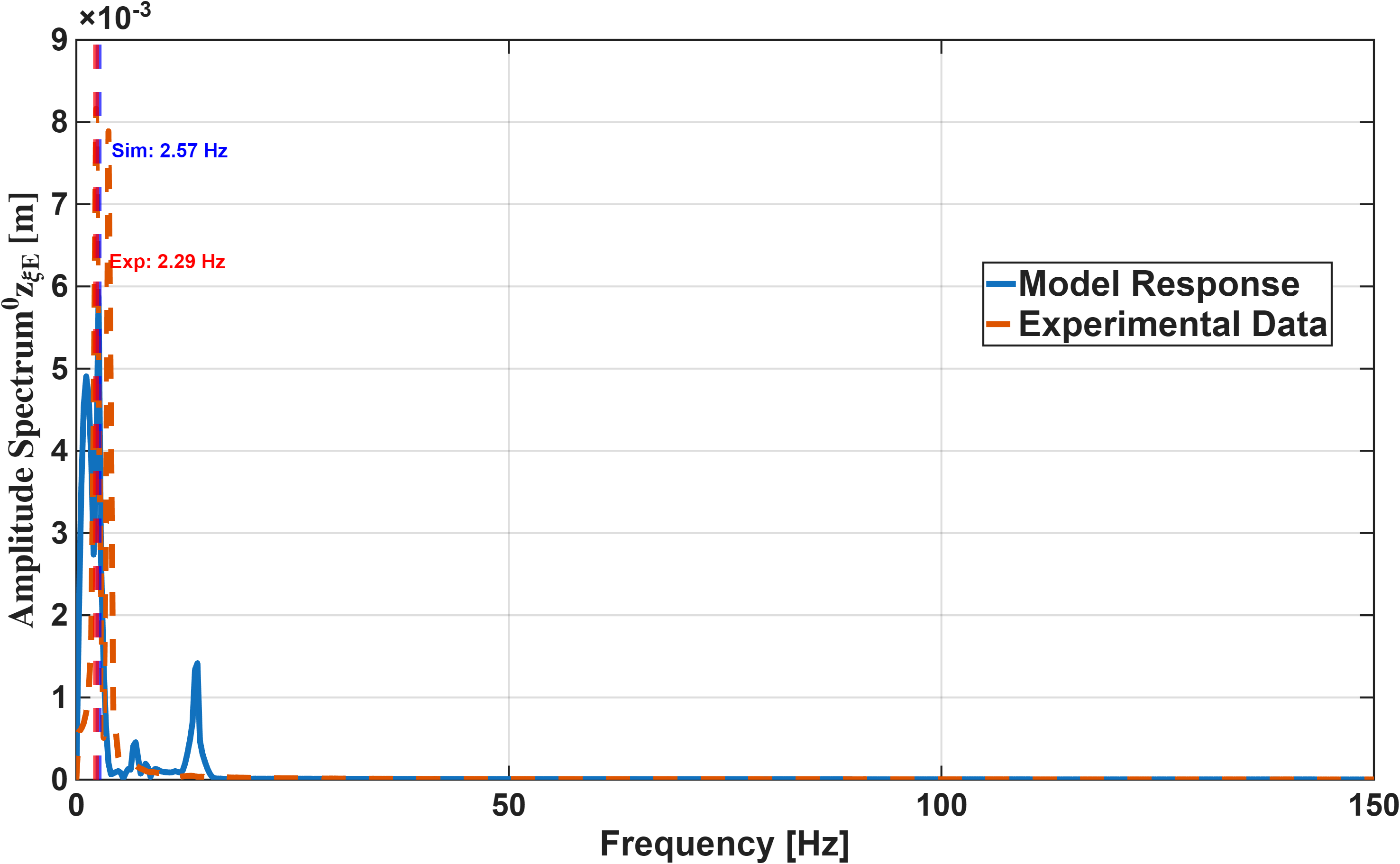}
\label{fig:specc}}
\caption{Frequency spectrum response for endpoint displacement: comparison between the dynamic model and experimental measurements: Amplitude spectrum for (a) $^{0}x_{\xi E}$ (b) $^{0}y_{\xi E}$ (c)  $^{0}z_{\xi E}$.}
\label{fig_spec}
\end{figure*}

\begin{figure*}[htbp]
\centering

\subfloat[]{
\includegraphics[width=0.45\textwidth]{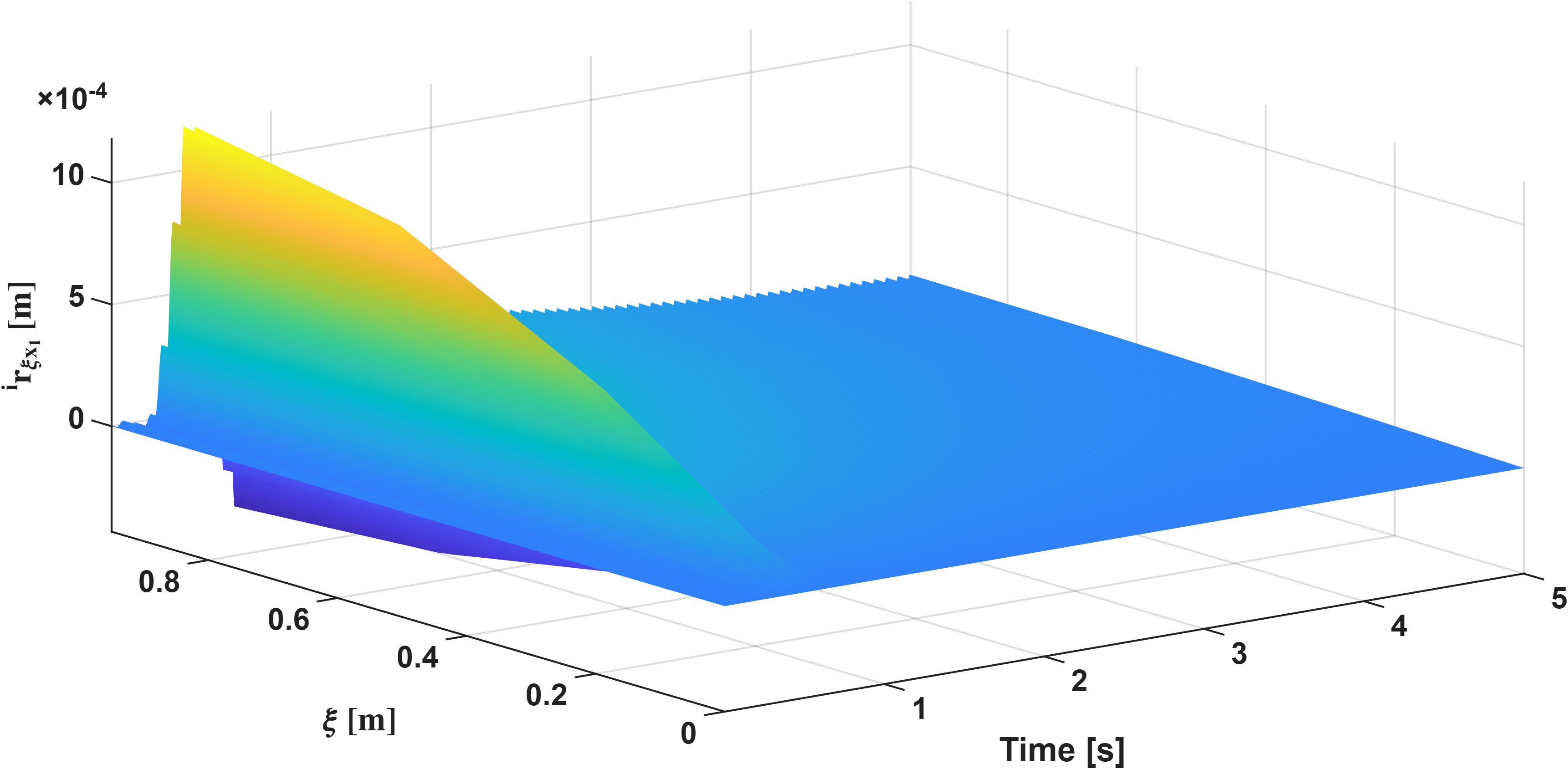}
\label{fig:9a}}
\hfill
\subfloat[]{
\includegraphics[width=0.45\textwidth]{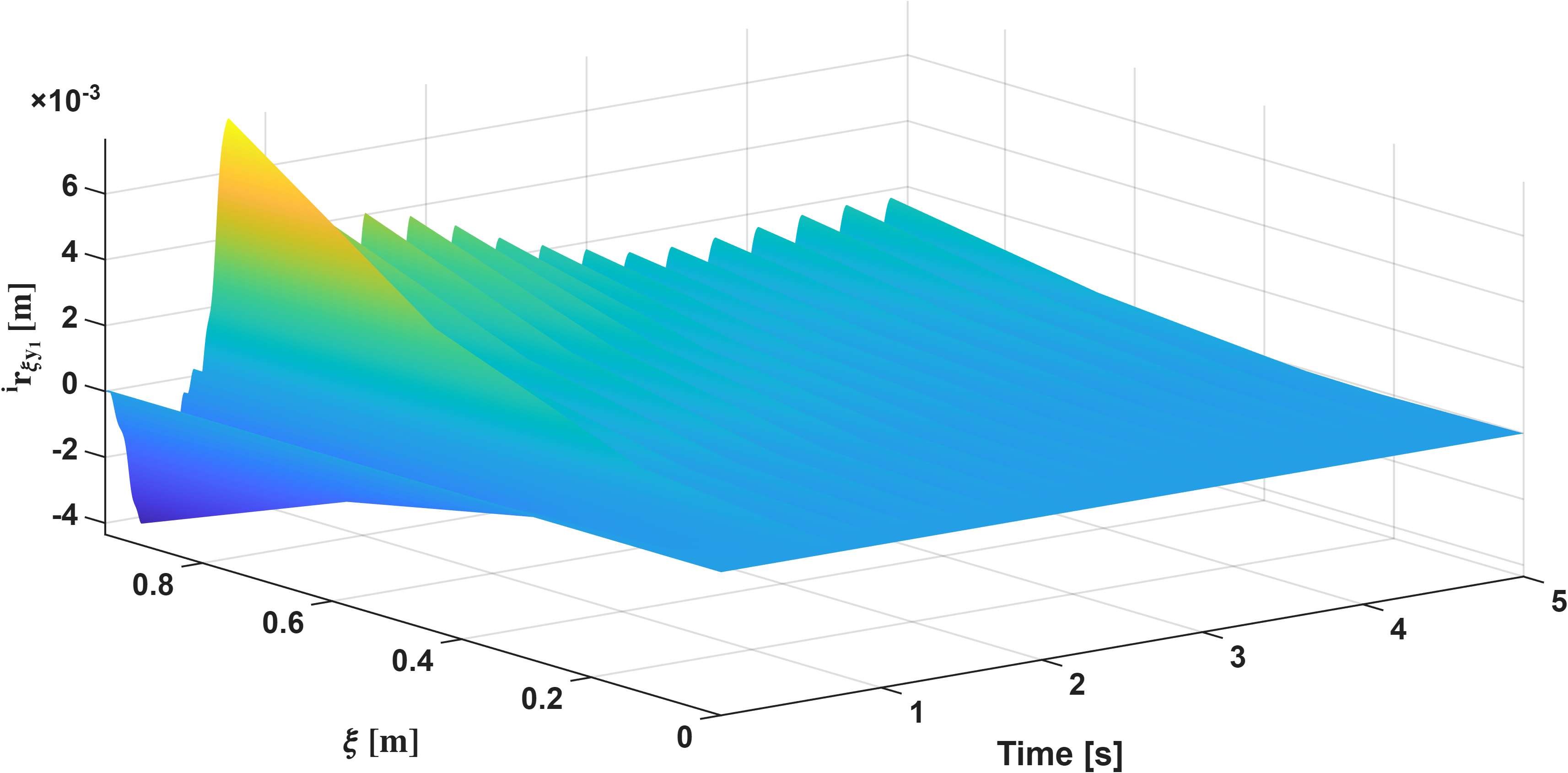}
\label{fig:9b}}

\vspace{1em}

\subfloat[]{
\includegraphics[width=0.45\textwidth]{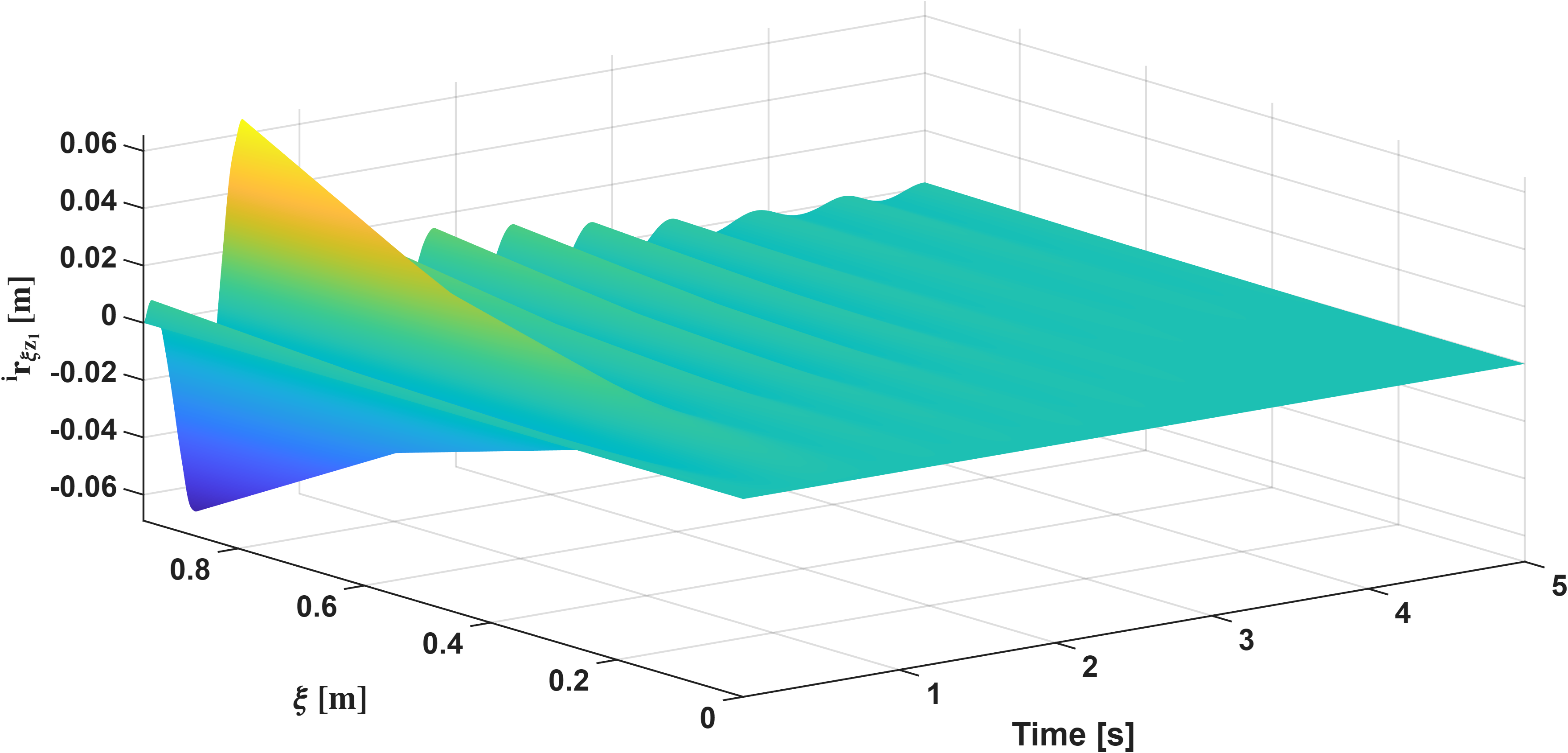}
\label{fig:9c}}
\caption{Displacement field for first link based on dynamic model (a) $^{i}r_{\xi x_1}$ (b) $^{i}r_{\xi y_1}$ (c)  $^{i}r_{\xi z_1}$.}
\label{fig_9}
\end{figure*}

\begin{figure*}[htbp]
\centering

\subfloat[]{
\includegraphics[width=0.45\textwidth]{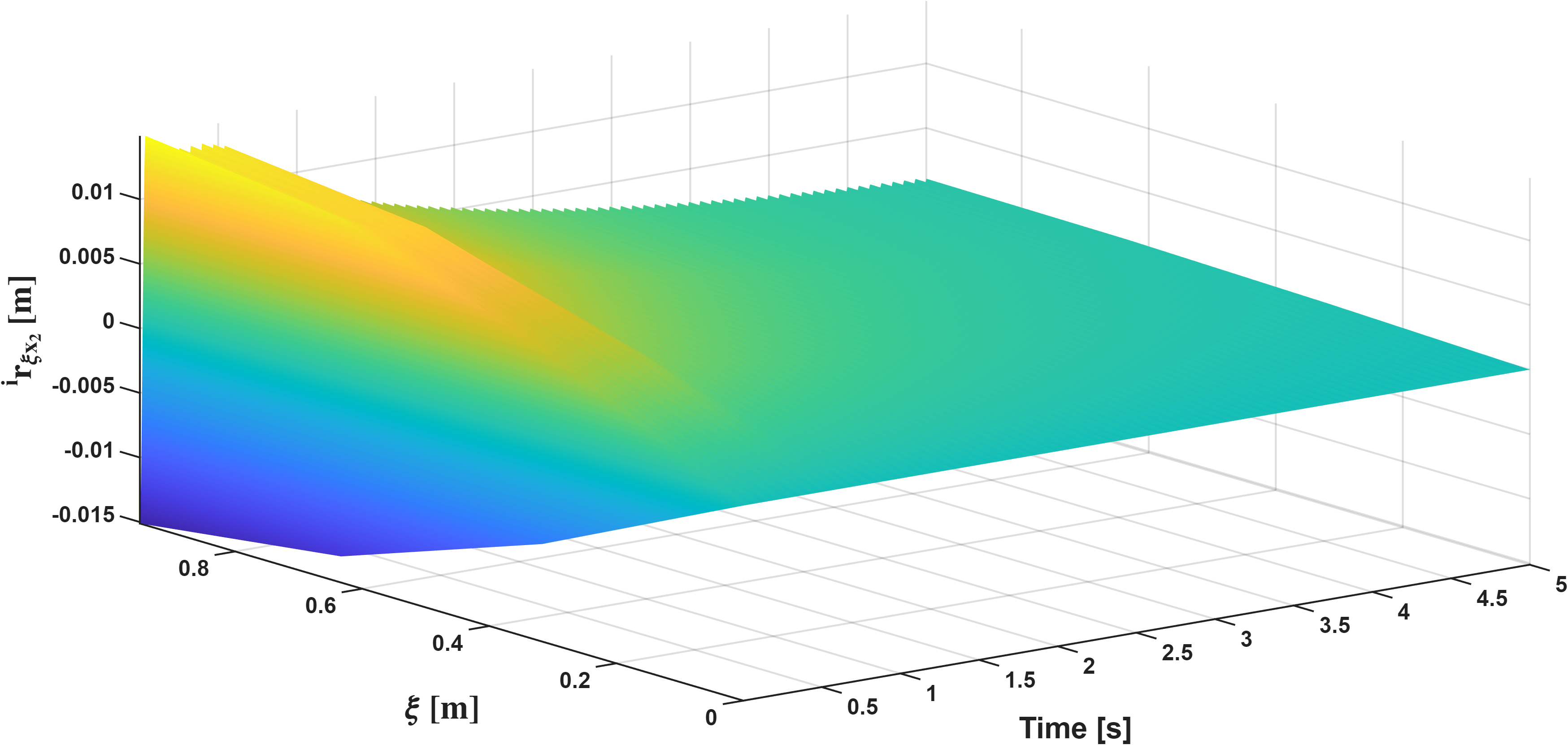}
\label{fig:10a}}
\hfill
\subfloat[]{
\includegraphics[width=0.45\textwidth]{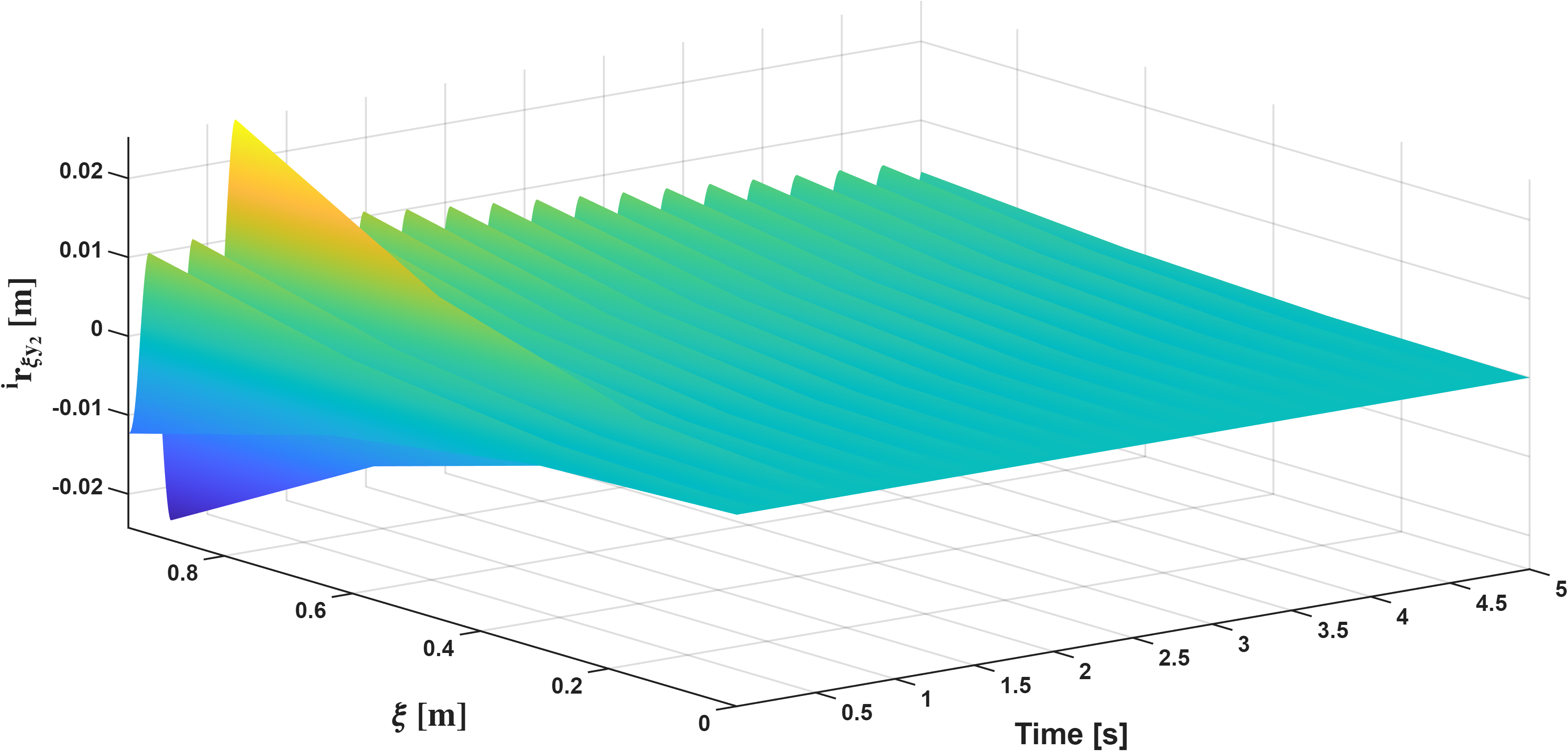}
\label{fig:10b}}

\vspace{1em}

\subfloat[]{
\includegraphics[width=0.45\textwidth]{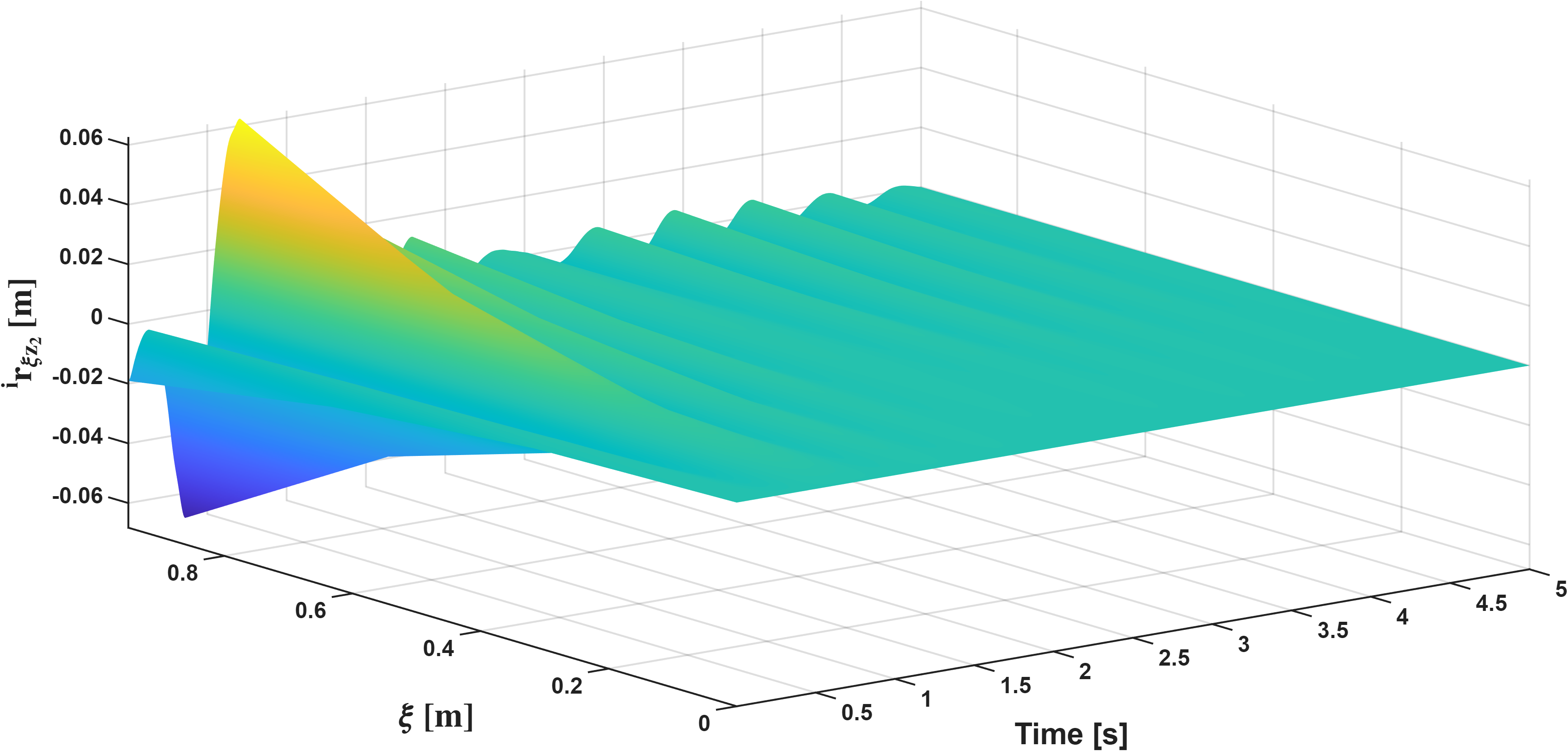}
\label{fig:10c}}
\caption{Displacement field for second link based on dynamic model (a) $^{i}r_{\xi x_2}$ (b) $^{i}r_{\xi y_2}$ (c)  $^{i}r_{\xi z_2}$.}
\label{fig_10}
\end{figure*}

\begin{figure*}[htbp]
\centering

\subfloat[]{
\includegraphics[width=0.45\textwidth]{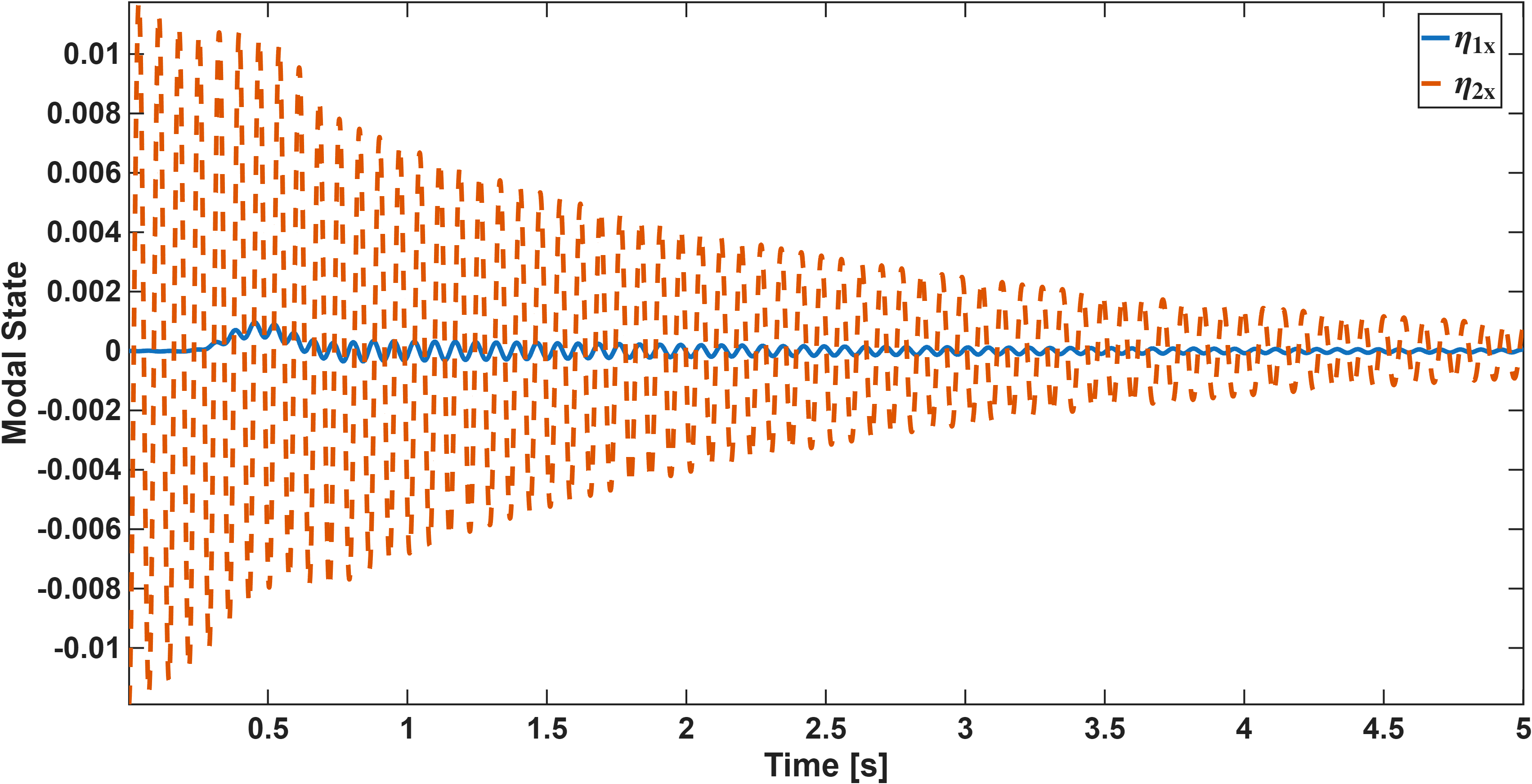}
\label{fig:11a}}
\hfill
\subfloat[]{
\includegraphics[width=0.45\textwidth]{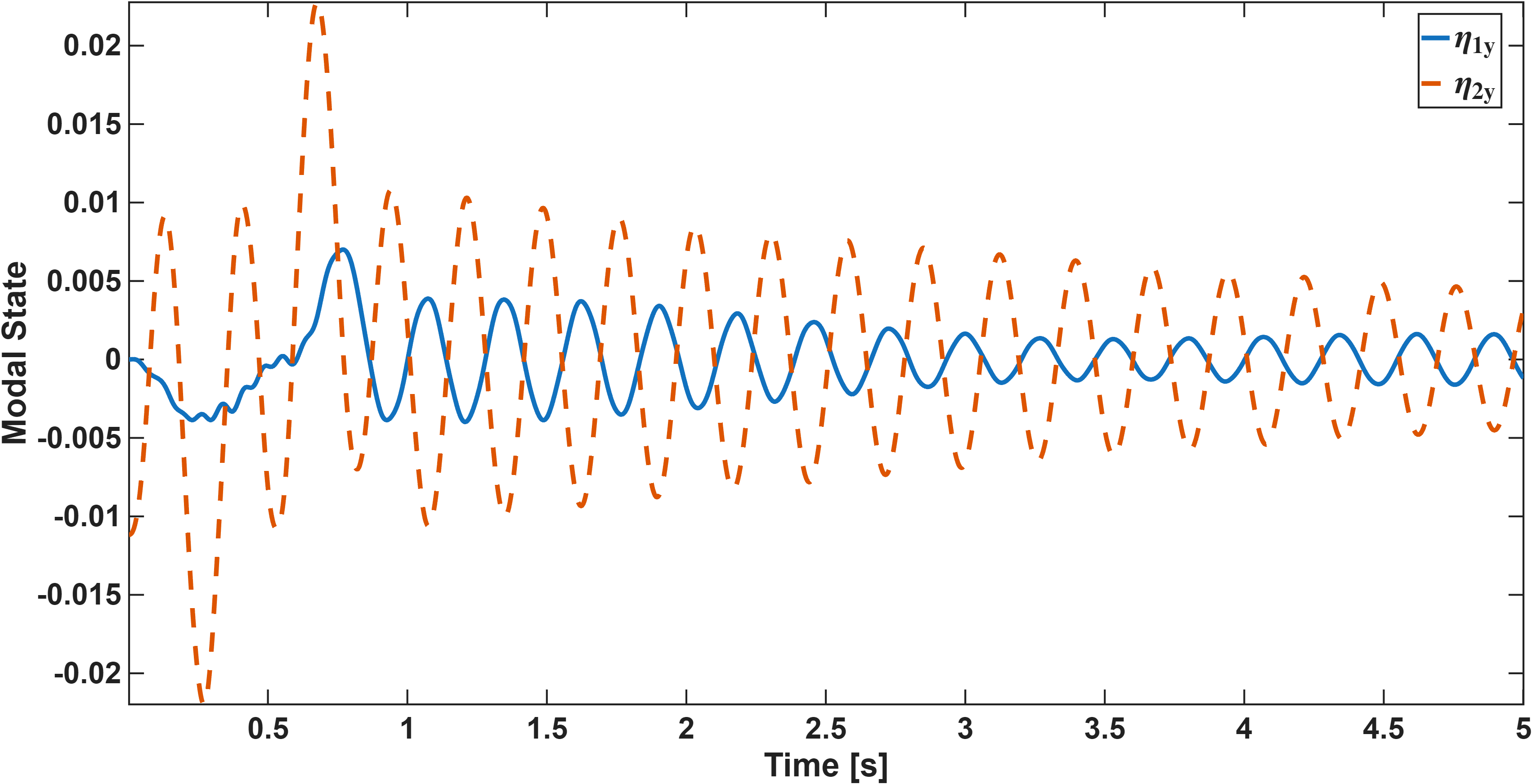}
\label{fig:11b}}

\vspace{1em}

\subfloat[]{
\includegraphics[width=0.45\textwidth]{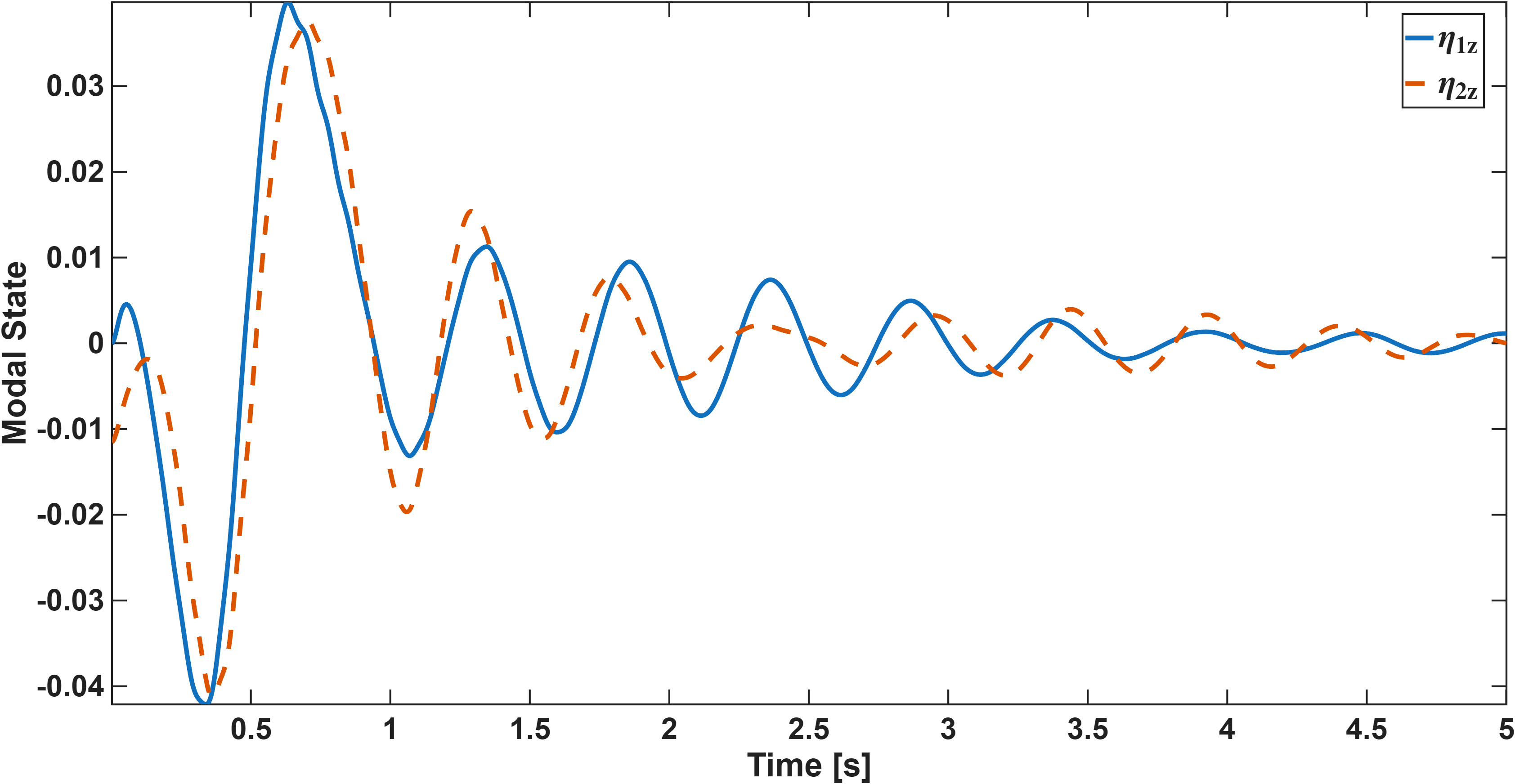}
\label{fig:11c}}
\caption{Modal states comparison between the two links (a) $\eta_{x}$ (b) $\eta_{y}$  (c)  $\eta_{z}$.}
\label{fig_11}
\end{figure*}

\begin{figure*}[htbp]
\centering

\subfloat[]{
\includegraphics[width=0.45\textwidth]{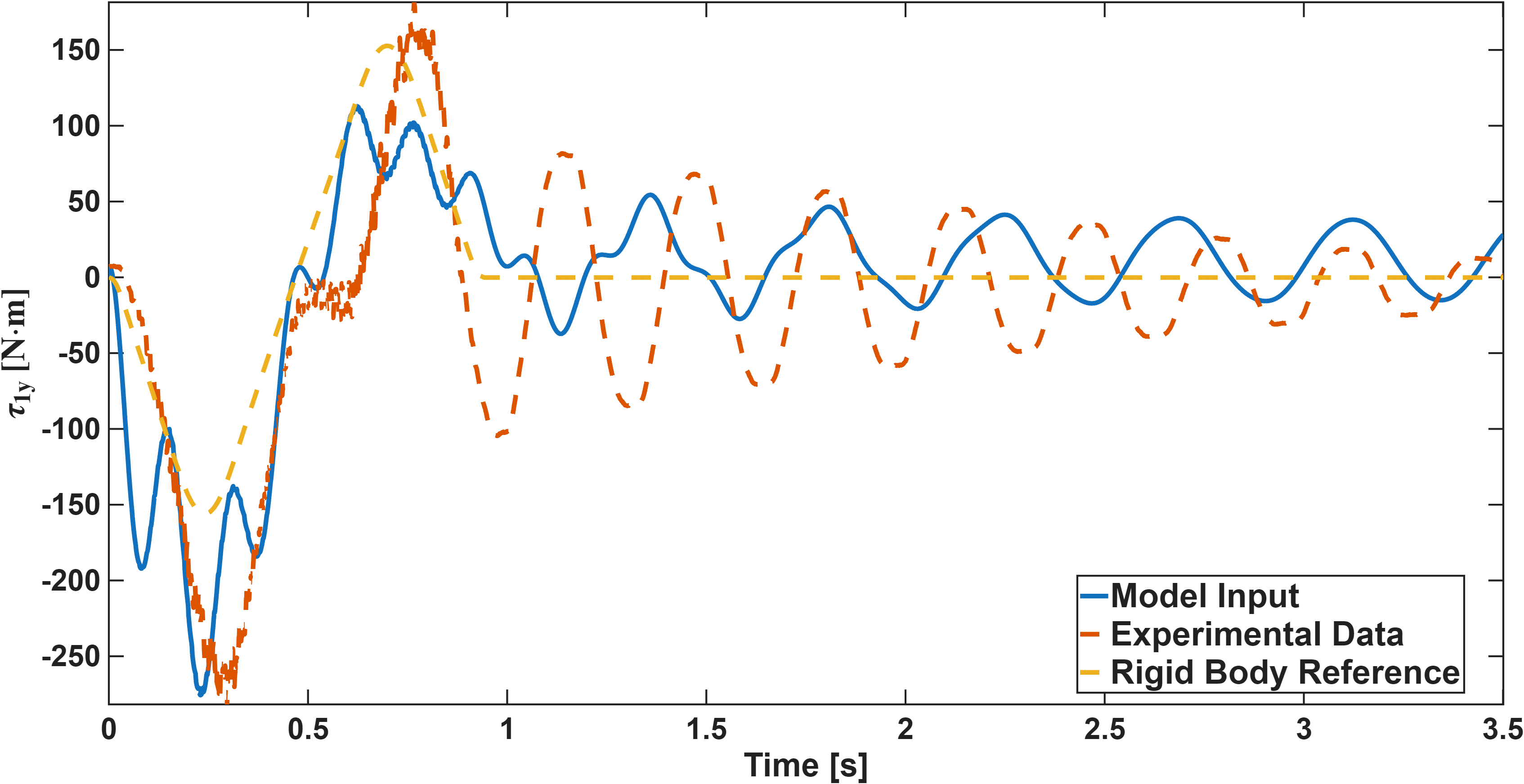}
\label{fig:12a}}
\hfill
\subfloat[]{
\includegraphics[width=0.45\textwidth]{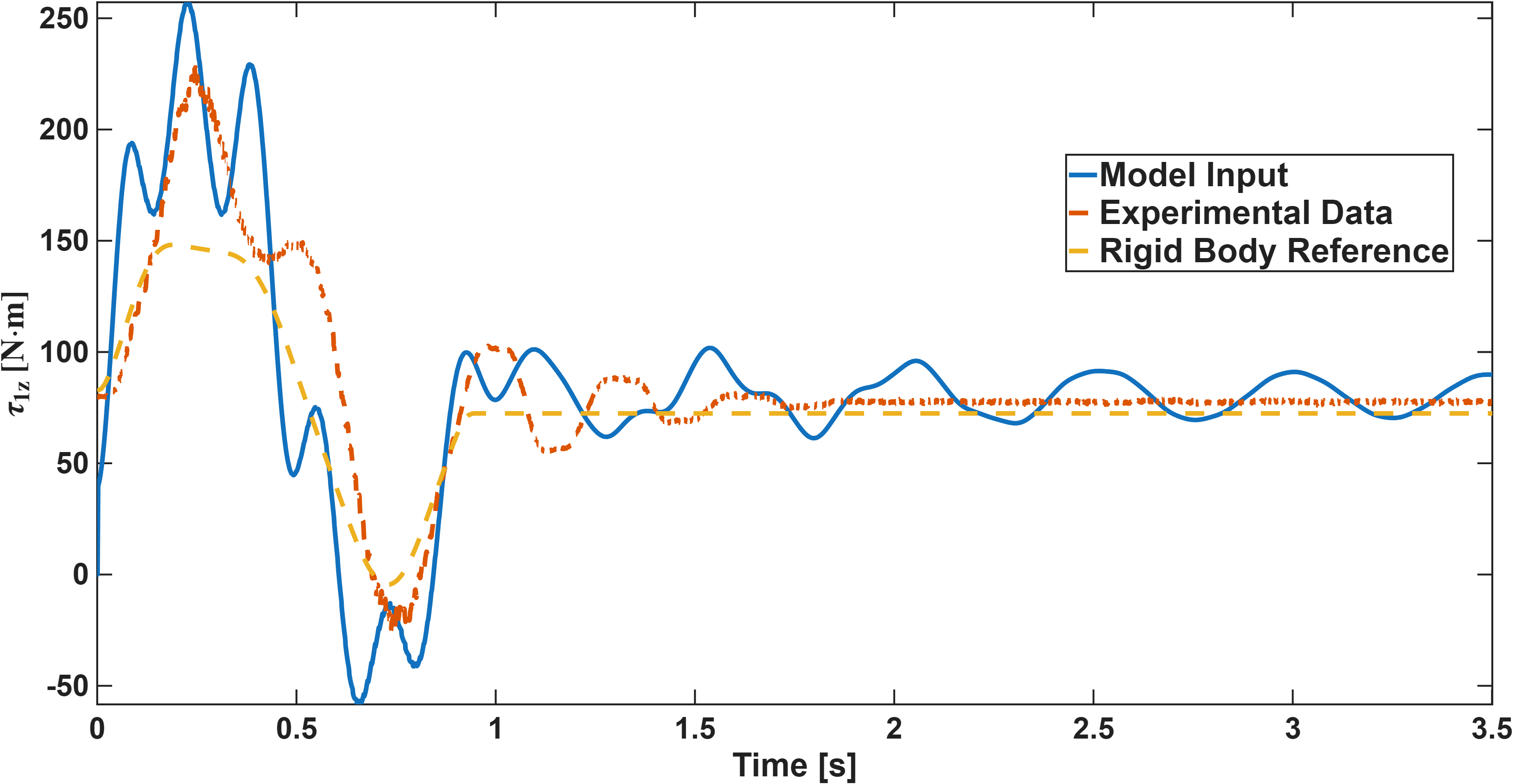}
\label{fig:12b}}

\vspace{1em}

\subfloat[]{
\includegraphics[width=0.45\textwidth]{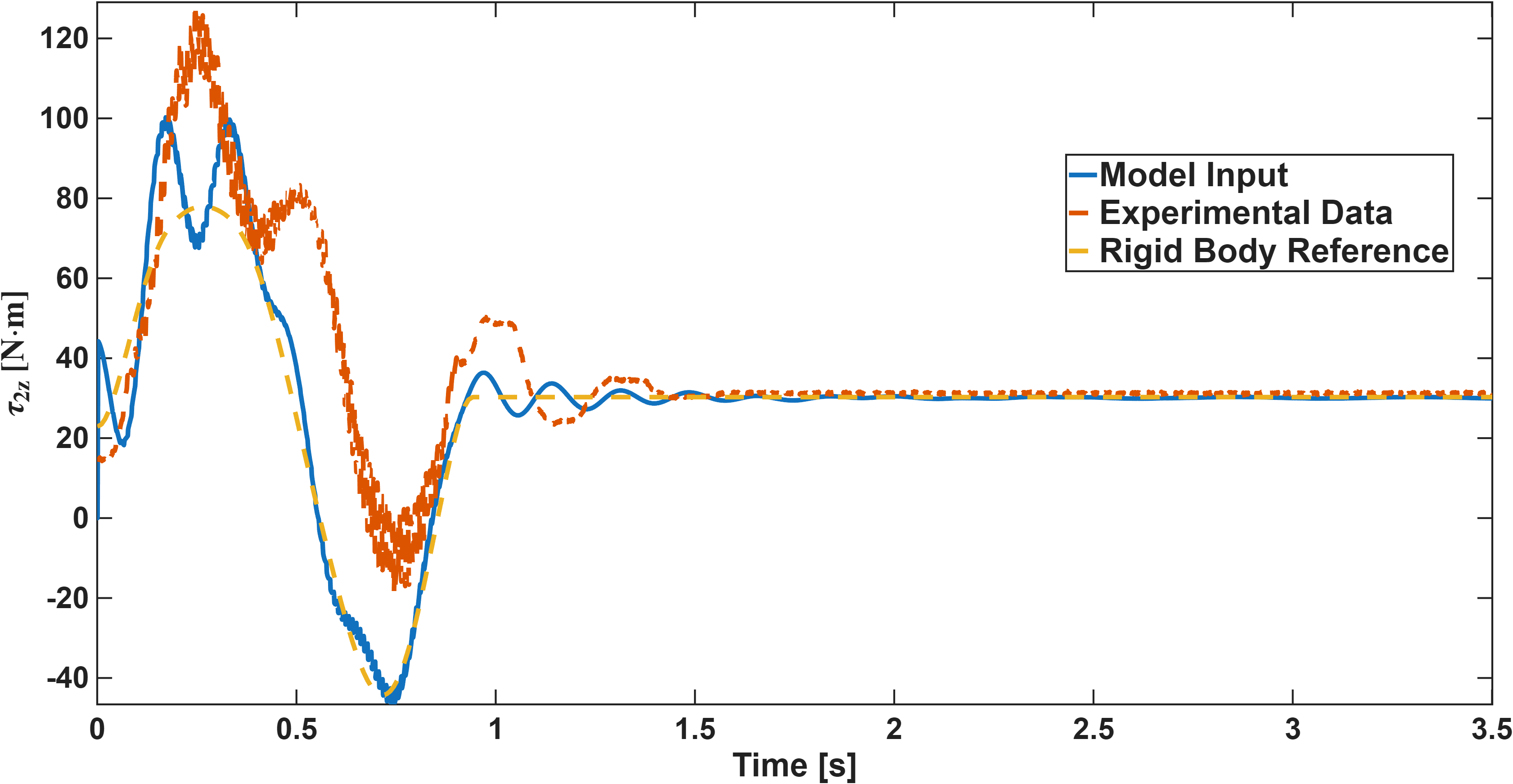}
\label{fig:12c}}
\caption{Control input torque generation: comparison between the dynamic model and experimental measurements (a) $^{i}\tau_{y_{1}}$ (b) $^{i}\tau_{z_{1}}$ (c)  $^{i}\tau_{z_{2}}$.}
\label{fig_12}
\end{figure*}

\begin{figure*}[htbp]
\centering

\subfloat[]{
\includegraphics[width=0.48\textwidth]{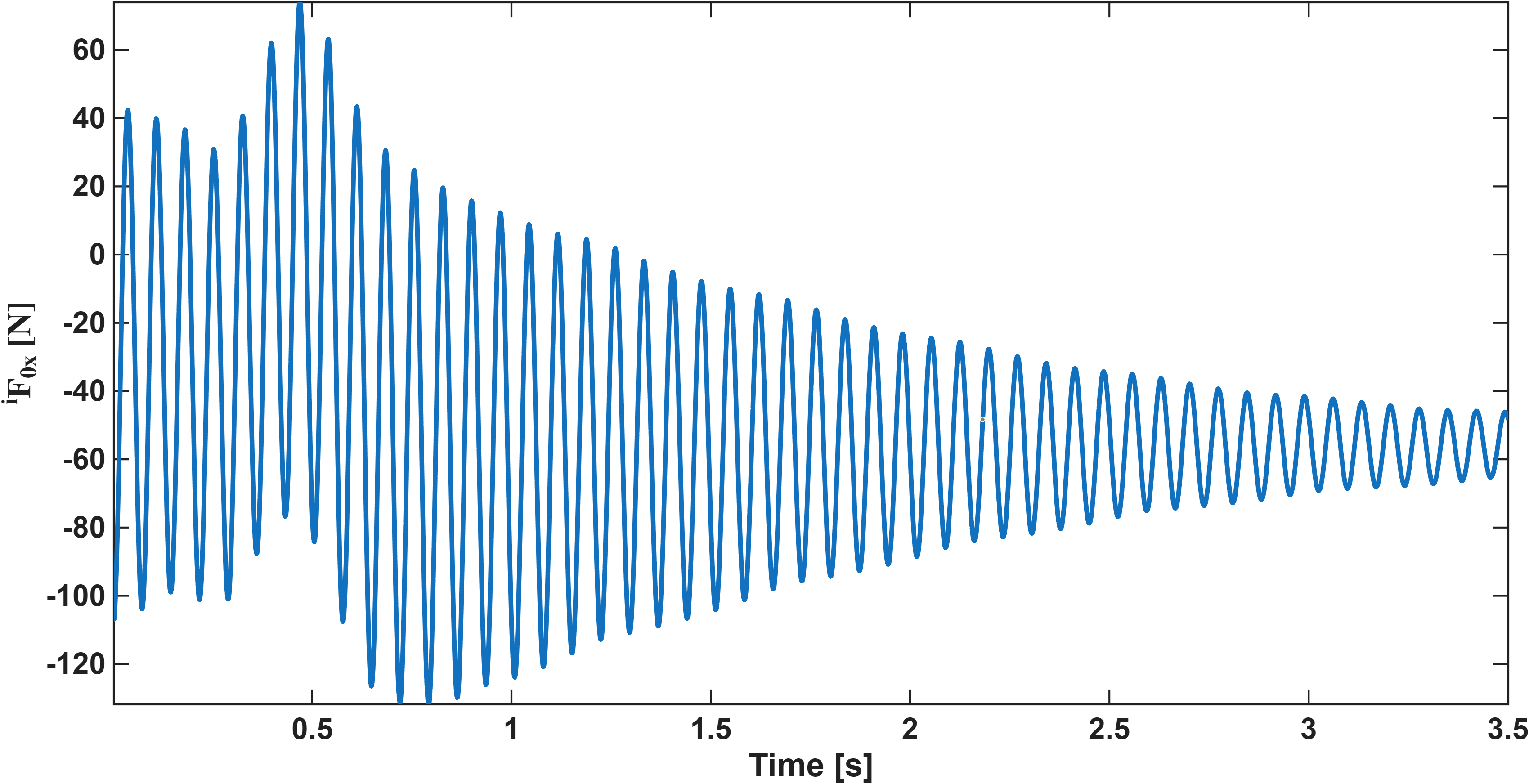}
\label{fig:13a}}
\hfill
\subfloat[]{
\includegraphics[width=0.48\textwidth]{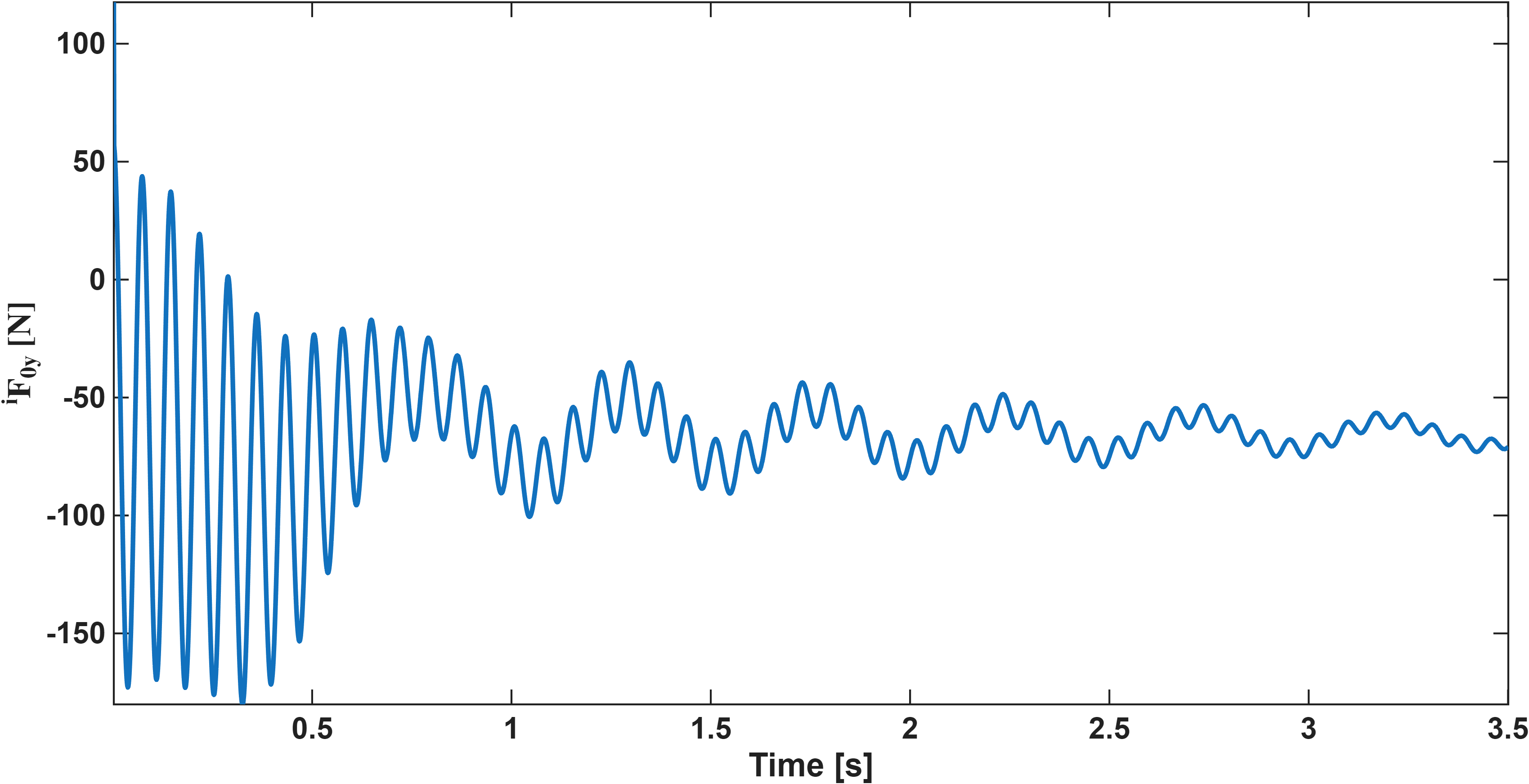}
\label{fig:13b}}

\vspace{1em}

\subfloat[]{
\includegraphics[width=0.48\textwidth]{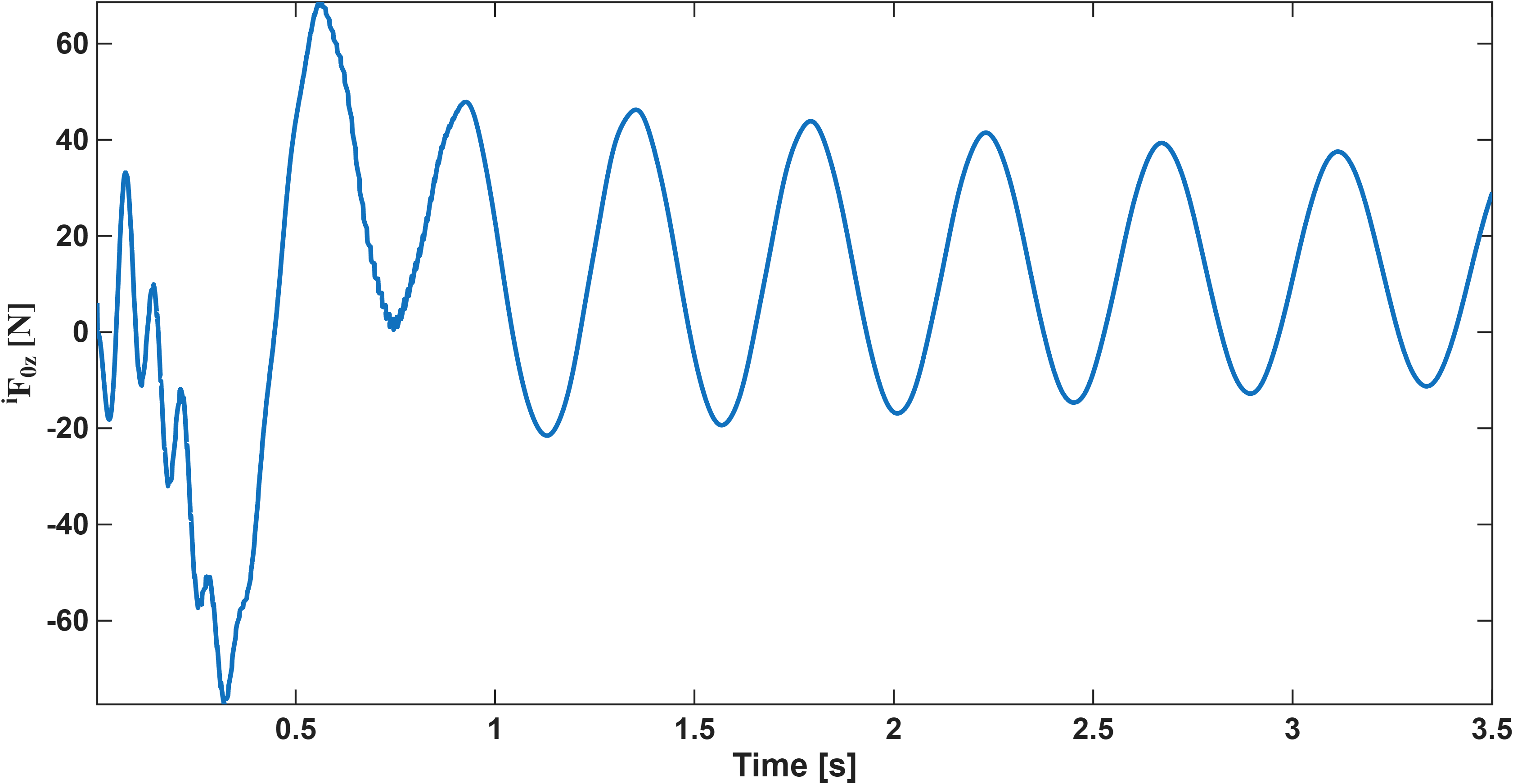}
\label{fig:13c}}
\hfill
\subfloat[]{
\includegraphics[width=0.48\textwidth]{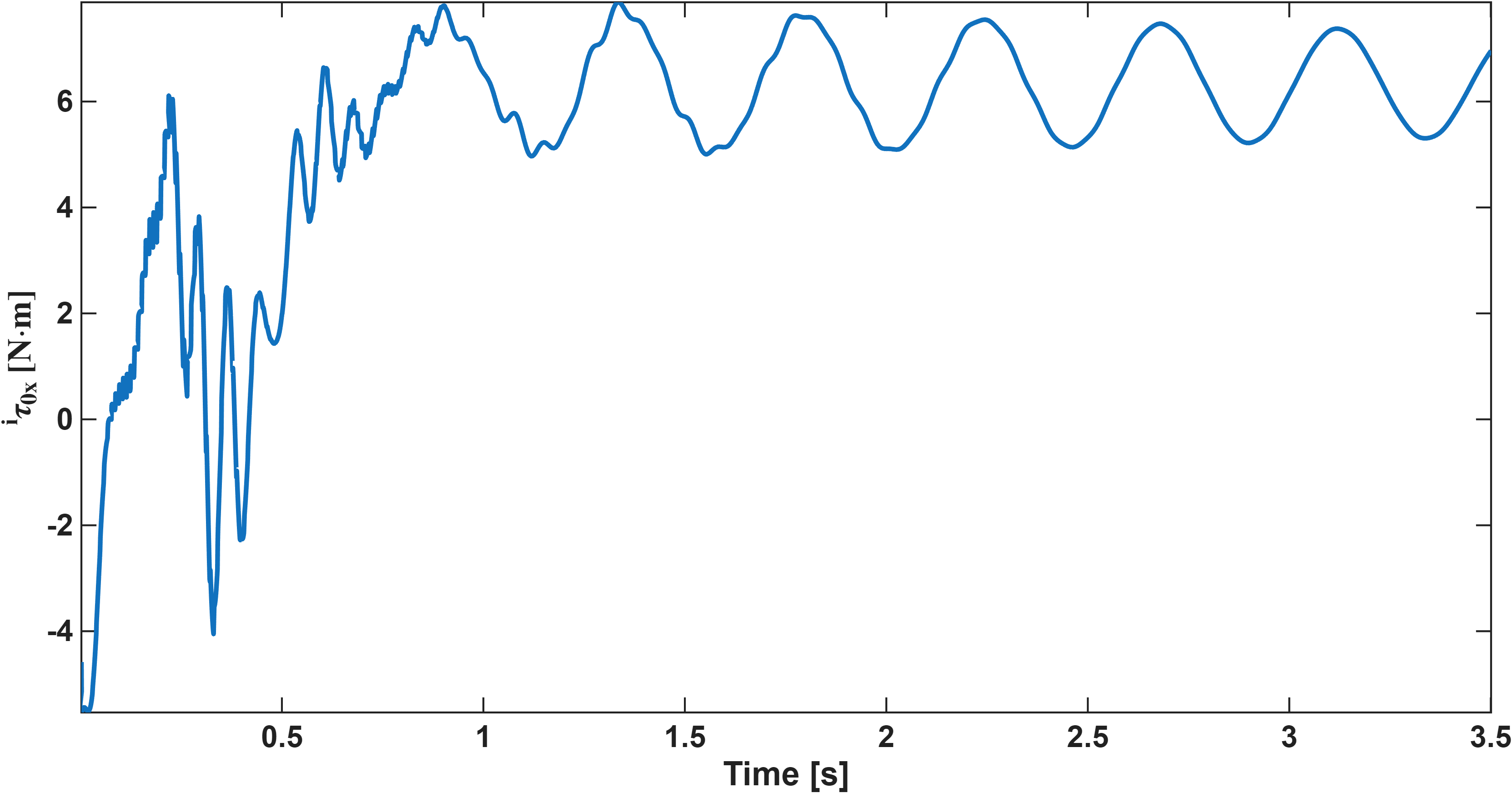}
\label{fig:13d}}

\vspace{1em}

\subfloat[]{
\includegraphics[width=0.48\textwidth]{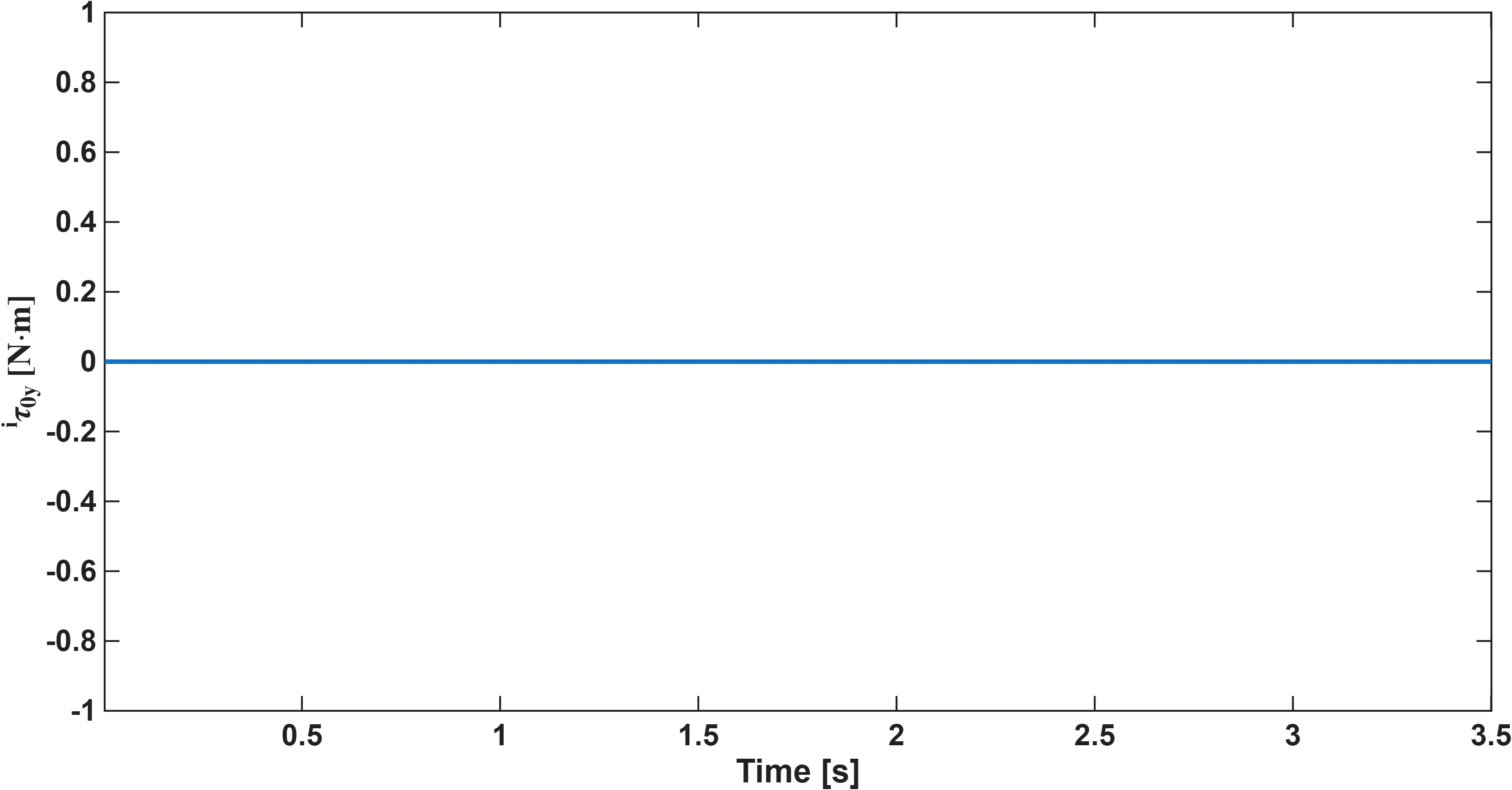}
\label{fig:13e}}
\hfill
\subfloat[]{
\includegraphics[width=0.48\textwidth]{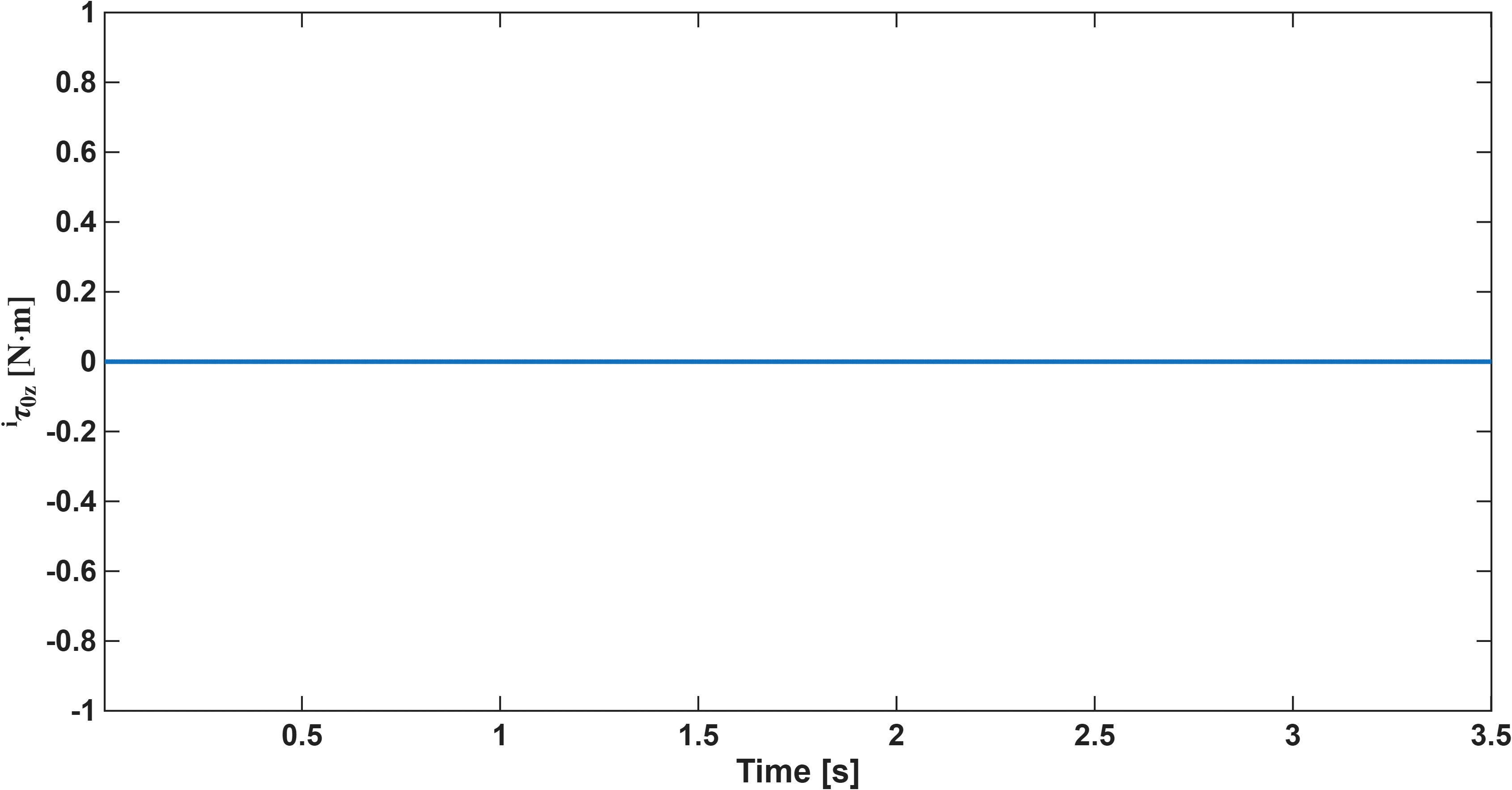}
\label{fig:13f}}

\caption{Interaction wrench elements of base joint as the algebraic state of multibody model: 
(a) $^{i}F_{J0x}$, (b) $^{i}F_{0y}$, (c) $^{i}F_{0z}$, (d) $^{i}\tau_{0x}$, (e) $^{i}\tau_{0y}$ (f) $^{i}\tau_{0z}$.}
\label{fig_13}
\end{figure*}

\begin{figure*}[htbp]
\centering

\subfloat[]{
\includegraphics[width=0.48\textwidth]{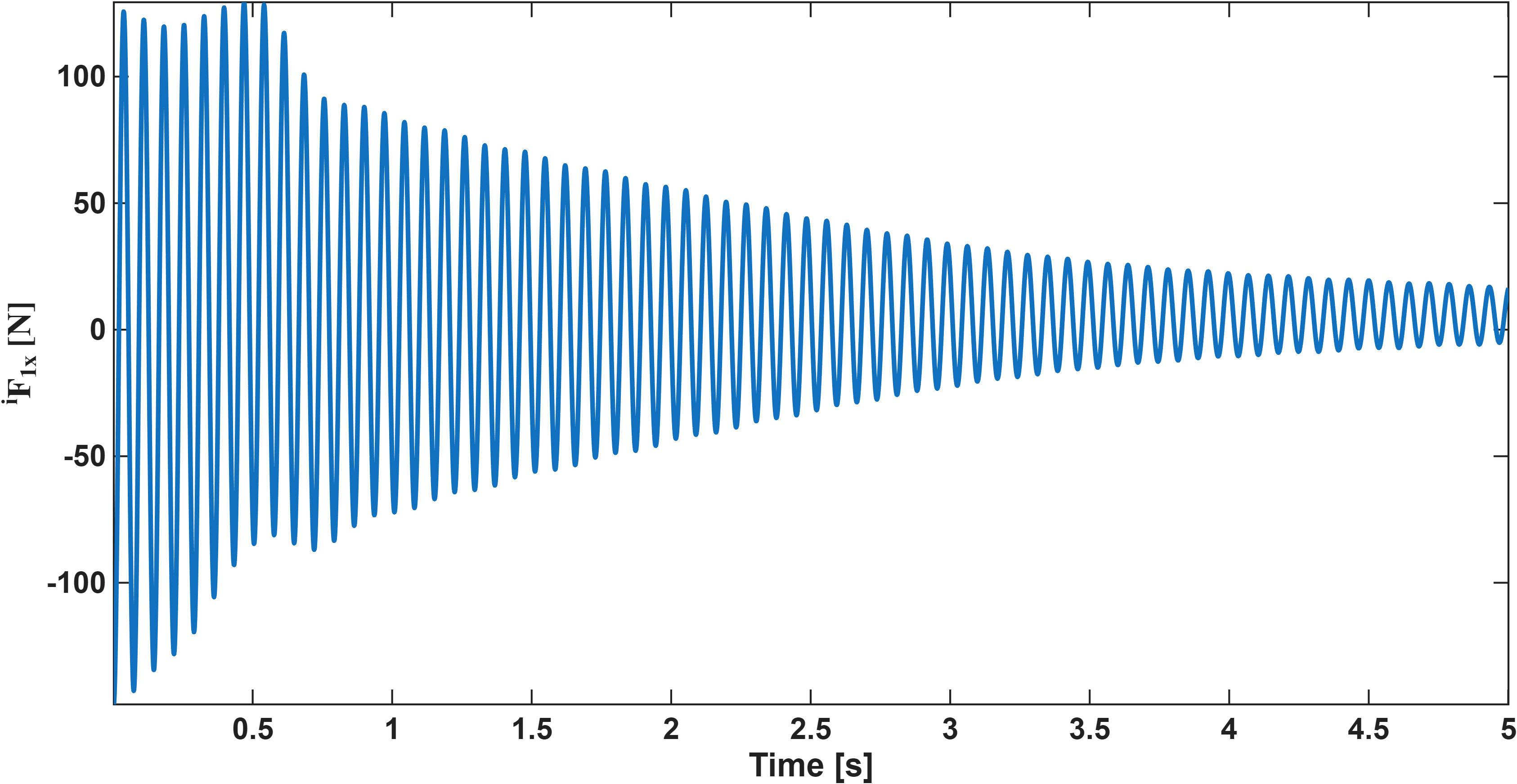}
\label{fig:14a}}
\hfill
\subfloat[]{
\includegraphics[width=0.48\textwidth]{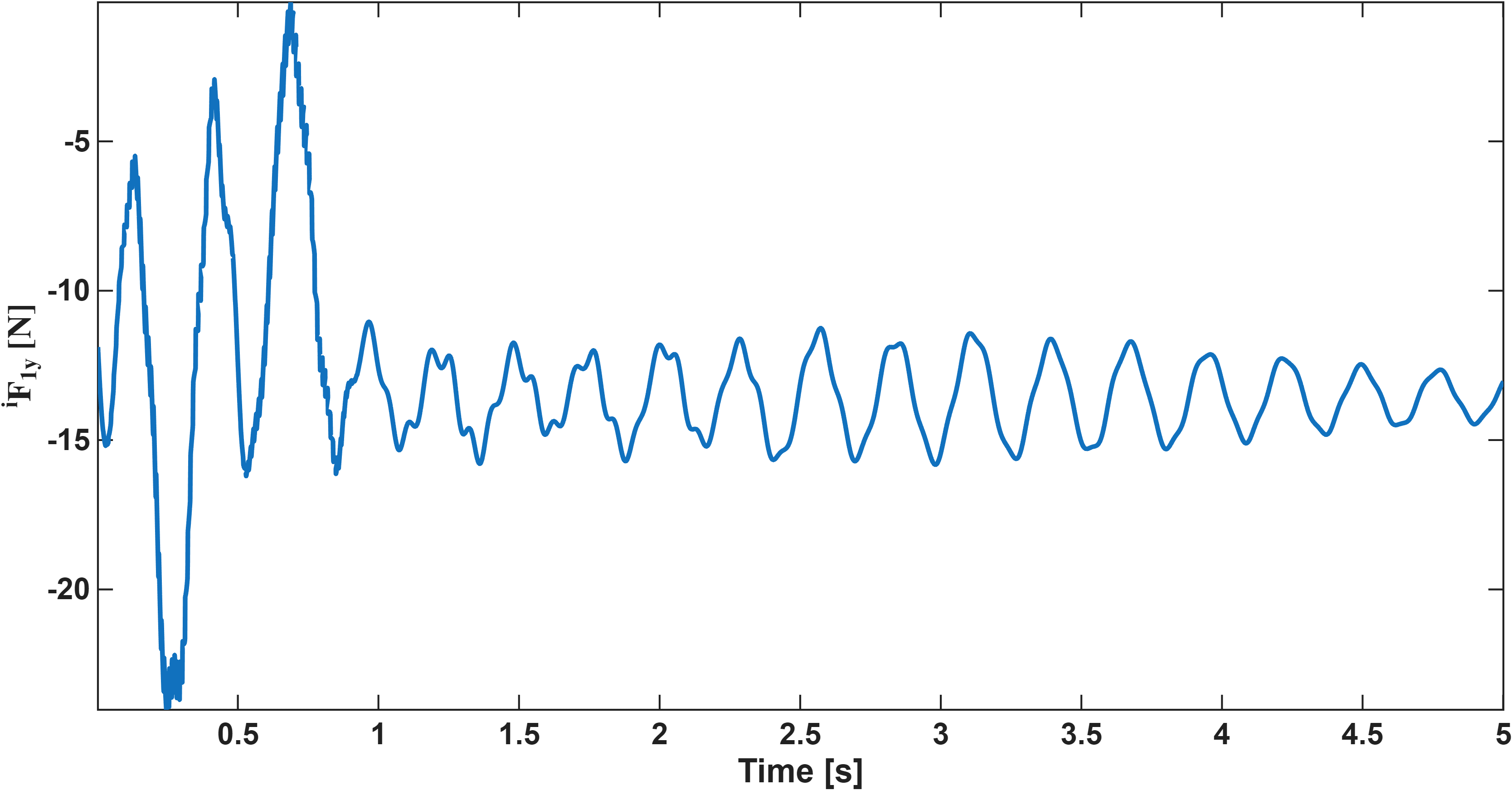}
\label{fig:14b}}

\vspace{1em}

\subfloat[]{
\includegraphics[width=0.48\textwidth]{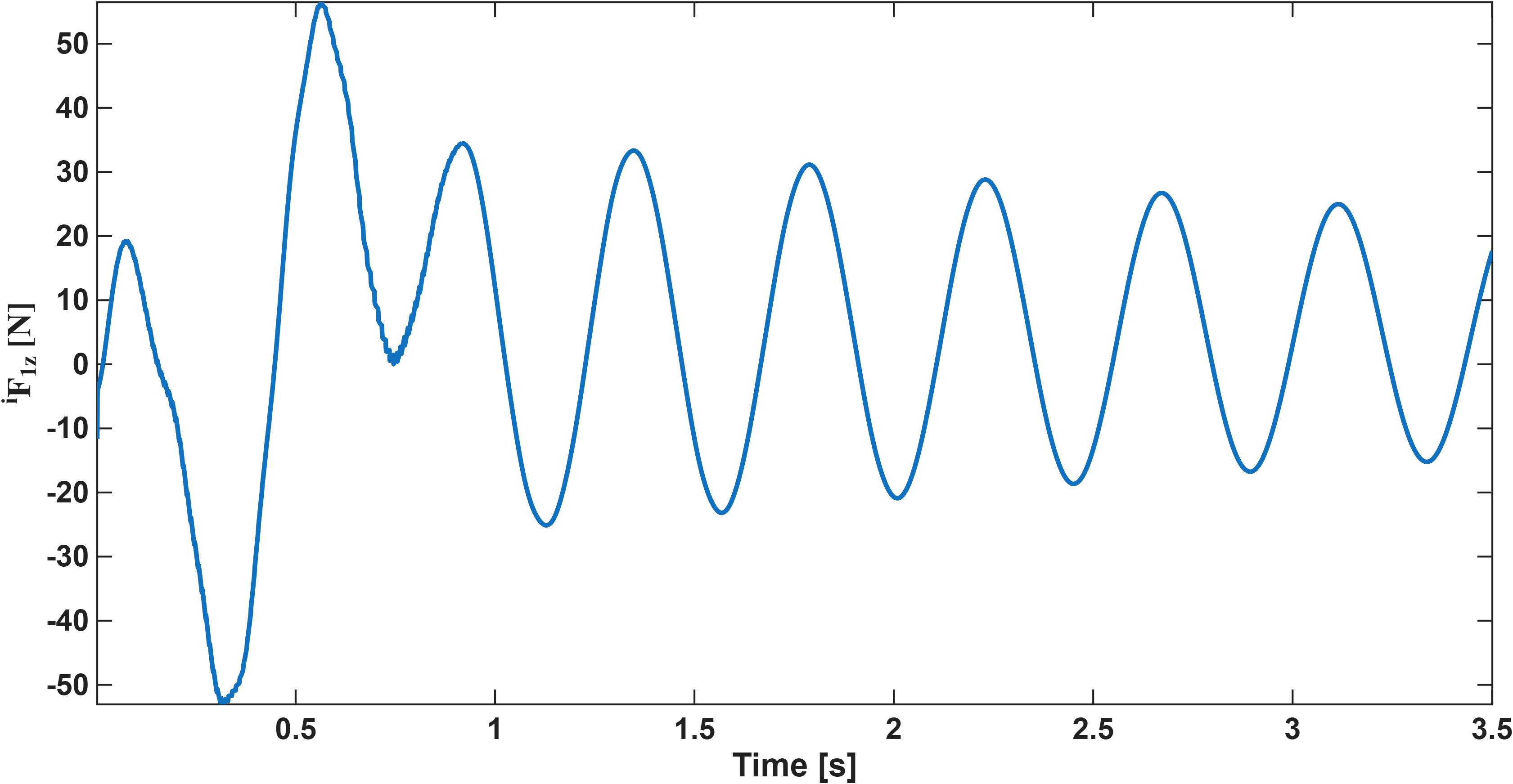}
\label{fig:14c}}
\hfill
\subfloat[]{
\includegraphics[width=0.48\textwidth]{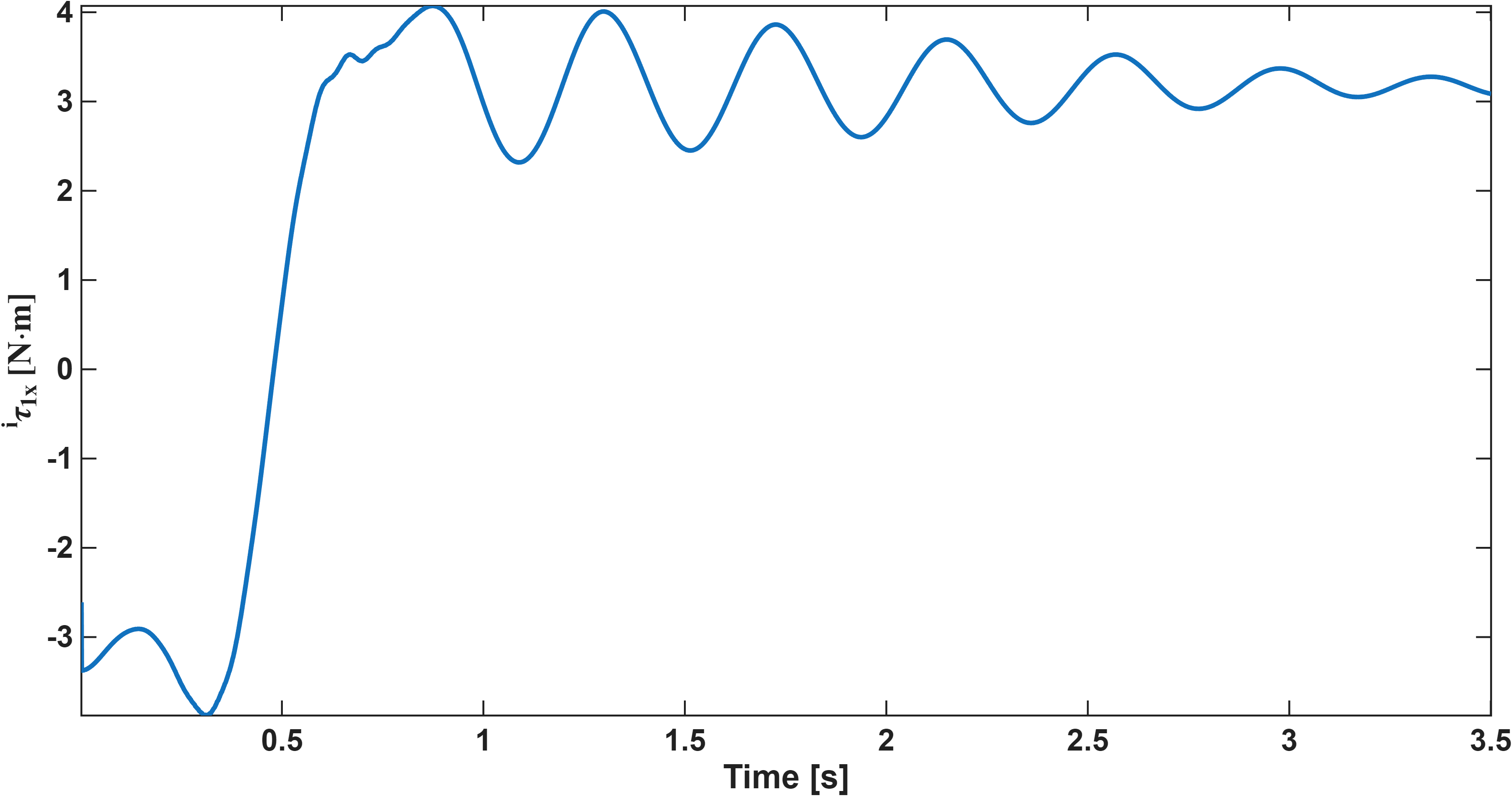}
\label{fig:14d}}

\vspace{1em}

\subfloat[]{
\includegraphics[width=0.48\textwidth]{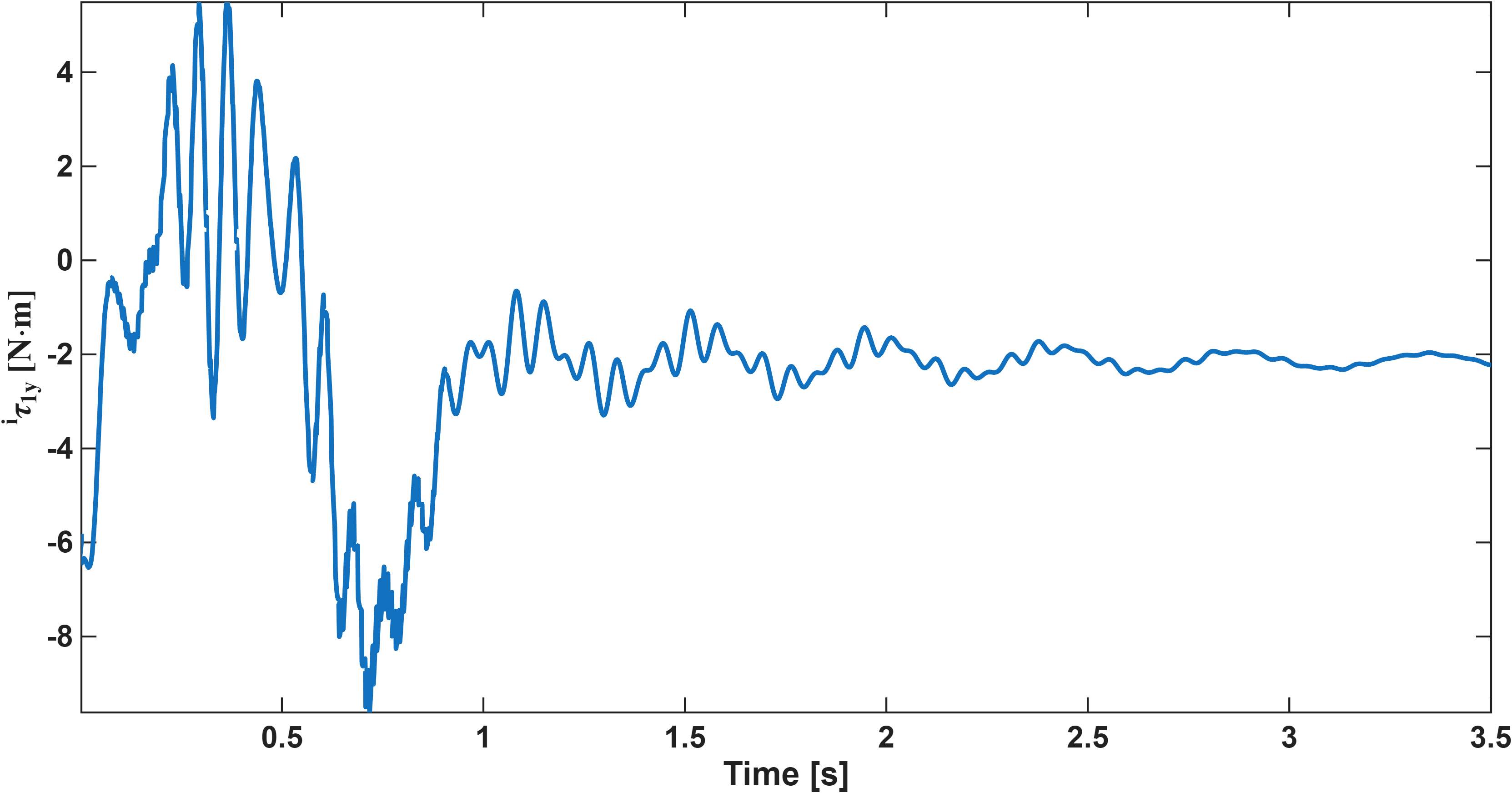}
\label{fig:14e}}
\hfill
\subfloat[]{
\includegraphics[width=0.48\textwidth]{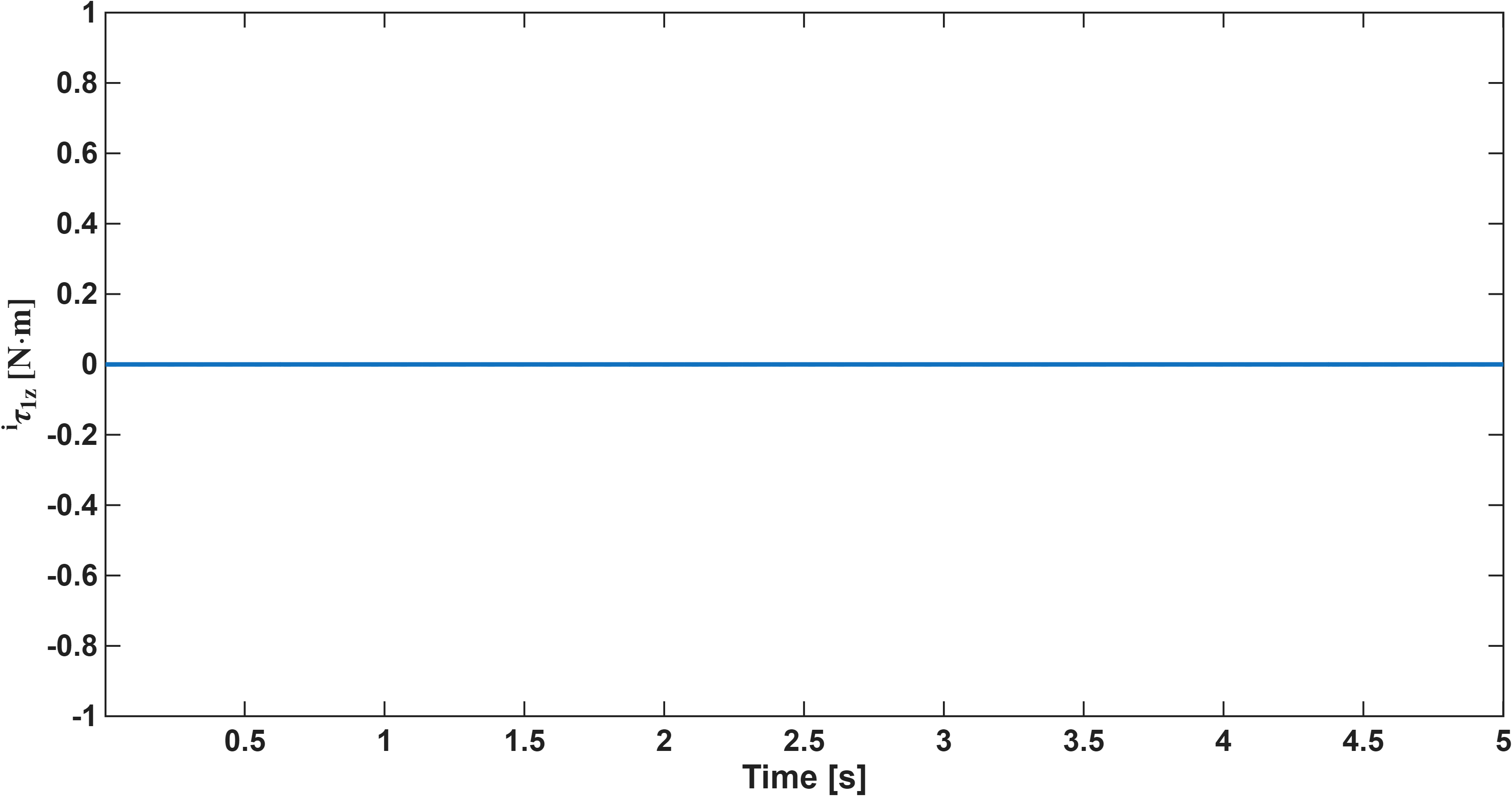}
\label{fig:14f}}

\caption{Interaction wrench elements of connecting joint as the algebraic state of multibody model: 
(a) $^{i}F_{1x}$, (b) $^{i}F_{1y}$, (c) $^{i}F_{1z}$, (d) $^{i}\tau_{1x}$, (e) $^{i}\tau_{1y}$ (f) $^{i}\tau_{1z}$.}
\label{fig_14}
\end{figure*}

The model response for the aforementioned configuration, properties, and simulation settings, and the corresponding comparative analysis, is described in Figs.~\ref{fig_4}--\ref{fig_14}. The comparative analysis of the model with alternate multibody modeling schemes of FFR and COFEM, is presented in the companion modeling paper~\cite{Yaqubi2026}.

The coordinates vector and twist elements describing the inertial-to-body frame motion are shown in Figs.~\ref{fig_4} and~\ref{fig_6} respectively, confirming the stability of the rigid body response and the bounded evolution of both position and velocity states across the simulation horizon. The controlled joint angle responses under PD control are presented in Fig.~\ref{fig_3}, comparing the dynamic model against experimental measurements for $^{i}\theta_{y_1}$, $^{i}\theta_{z_1}$, and $^{i}\theta_{z_2}$. The model responses converge to the reference trajectories and remain stable throughout the simulation interval, consistent with the experimental measurements.

The endpoint position and deformation responses are presented in 
Figs.~\ref{fig_7} and~\ref{fig_8} respectively. The endpoint position 
$^{0}\mathbf{r}_{\xi E}$ is directly obtained from the model as the inertial 
position of the tip of the second link, comprising both the rigid body contribution 
from the joint configuration and the elastic deformation field. In the experimental 
setting, isolating the deformation component has been conducted by high-pass 
filtering of the raw tip displacement signal to retain only the oscillatory elastic 
content. The simulated endpoint deformation remains within the experimentally observed range throughout the motion, with both signals exhibiting oscillations of comparable 
frequency and amplitude across all three spatial directions. Exact phase 
correspondence is not expected given the single-mode truncation of the flexibility 
model and the sensitivity of the deformation response to initial modal conditions, as well as the sheer multitude of involved factors in constrained three-dimensional elastic motion measurement and modeling uncertainty. The result confirms that the screw-theoretic formulation captures a stable response of deformation field along all deformation axes throughout the length of all links.

The frequency spectra of the endpoint deformation are presented in 
Fig.~\ref{fig_spec}, showing the amplitude spectrum for each spatial component. The 
dominant bending frequency identified from the model is in close agreement with 
the experimental spectrum in all three directions, directly validating the modal 
wavenumber identification procedure and the screw-theoretic modal formulation. The 
simulated spectra exhibit minor side lobes adjacent to the dominant peak, 
attributable to spectral leakage arising from the non-periodic nature of the 
simulated response over the finite analysis window, and initial measurements which have been directly implemented in the model, with no error correction. Additionally, the simulated response exhibits a secondary frequency component near 13~[Hz], consistent with the 
theoretical second bending mode of the clamped-free beam at $\left(4.6941/1.8751 
\right)^2 \approx 6.27$ times the fundamental frequency. The displacement fields along the length of each link are shown in Figs.~\ref{fig_9} and~\ref{fig_10}, and the corresponding modal coordinate evolution is presented in Fig.~\ref{fig_11}. These results confirm that the elastic deformation remains small relative to the link length throughout the motion, 
consistent with the small-deformation assumption underlying the Euler-Bernoulli foundation, and that the modal states of both links evolve in a bounded and 
physically consistent manner. Additionally, the comprehensive information regarding the deformation field would enable analysis and design of any corresponding system--for example, distributed control schemes--that require availability of this data, in a model-based manner, with no numerically intensive FEM-based meshing involved.

The control input torques are presented in Fig.~\ref{fig_12}, comparing the 
the experimental torque measurements and the model response for $^{i}\tau_{y_1}$, $^{i}\tau_{z_1}$, and $^{i}\tau_{z_2}$. Additionally, the PD input for a rigid-link of identical properties, calculated online as a part of experimental implementation code simultaneously, is included as another reference baseline to observe the dynamic effects caused by link flexibility. All three signals exhibit consistent trends and remain stable throughout the simulation interval. Discrepancies between the theoretical command and the experimental measurement arise from initial condition mismatches, unmodeled actuator dynamics, and the inherent difference between the discrete numerical controller and the physical motor driver implementation. Nevertheless, the overall torque pattern is reproduced by the model, confirming that the closed-loop dynamic response is physically consistent with the experimental setup.

The choice of PD control for validation carries specific methodological significance 
beyond its practical simplicity. Since a PD controller generates torque commands 
purely as a function of instantaneous state error without relying on any internal 
representation of the plant dynamics, it cannot compensate for structural errors in 
the model the way a model-based controller such as computed torque or feedback 
linearization would---the latter explicitly inverts the plant model and can achieve 
correct tracking even when applied to an incorrect one, validating the controller 
rather than the model. Consequently, the fact that the PD-controlled simulation 
simultaneously reproduces the joint angle trajectories, endpoint displacement 
profile, and torque signals consistent with experimental measurements constitutes 
a meaningful validation criterion: it indicates that the simulated system dissipates 
and stores energy in a manner consistent with the physical plant, a necessary 
condition for the closed-loop trajectories to remain bounded and phase-consistent under a controller with no knowledge of the plant dynamics. The bounded evolution of all states and control inputs throughout the simulation horizon therefore confirms the dynamic consistency of the screw-theoretic multibody model and supports its suitability as a basis for future model-based control design.

The interaction wrench elements at the base joint and connecting joint are presented 
in Figs.~\ref{fig_13} and~\ref{fig_14} respectively. These wrenches are obtained 
directly as algebraic states of the multibody model, culminating in expression (\ref*{eq:DAESOL}) of DAE solution, arising naturally from the 
constraint equations without requiring post-processing or additional computation. 
This is a direct consequence of the closed-form screw-theoretic synthesis, in which the interaction wrenches at each joint appear explicitly in the system equations. The base joint wrench $[^{i}F_{0}, {}^{i}\boldsymbol{\tau}_{0}]$ and the connecting joint wrench $[^{i}F_{1}, {}^{i}\boldsymbol{\tau}_{1}]$ both evolve in a bounded and 
physically consistent manner, confirming that the joint constraints are satisfied throughout the simulation. The magnitudes of the interaction forces and moments are consistent with the inertial and gravitational loading of the two-link configuration, providing an additional validation of the dynamic model alongside the trajectory, input, and displacement-level comparisons. This results in ensuring that the physical constraints of the system are satisfied throughout, which includes zero translational velocity of the base link, synchronous motion of endpoint of first link and base of the second link, rotation constraints along axes of revolute joint for base joint, and the sole rotation axis for the connecting joint. The constraint satisfaction is confirmed based on the reference screw response previously depicted in Figs.~\ref{fig_4} and~\ref{fig_6}, eliminating the risk of model divergence and numerical drift of constraint terms.

\section{Conclusion}

This paper has presented a screw-theoretic PDE-based framework for dynamic 
modeling of multi-link flexible robotic systems in three-dimensional space. 
All dynamic states — rigid-body motion, elastic deformation, and inter-link 
interaction forces — are expressed in screw theoretic framework, retaining the infinite-dimensional PDE structure of 
the deformation field exactly while admitting a linear-algebraic formulation 
that makes multibody assembly and formal well-posedness analysis 
simultaneously tractable. Joint constraints are enforced as 
screw-compatibility equations, and the per-link models are assembled via a 
closed-form stacking procedure yielding a semi-explicit index-1 
differential-algebraic system with near-tridiagonal block structure. 
Well-posedness is established by recasting the assembled system in abstract 
Cauchy form through modal projection, whose regularity follows directly from the energy-consistent dynamic model and corresponding synthesis structure. The framework was validated experimentally on a two-link flexible manipulator under PD feedback control, demonstrating close agreement in joint trajectories, endpoint deformation, and dominant bending frequencies, while interaction wrenches at both joints were recovered directly as algebraic 
states and confirmed to satisfy the kinematic constraints throughout the 
simulation.

Several directions for future work emerge naturally from this framework. 
The closed-form stacking procedure extends in principle to non-serial 
topologies: parallel robots and branched kinematic chains could be 
accommodated by appending the corresponding dynamic and constraint rows to 
the system matrix, without restructuring the existing link equations. A 
recursive representation of the dynamic model may be pursued as an alternative to the closed-form stacking, recovering the direct $O(n)$ complexity of recursive Newton-Euler methods while retaining 
the screw-theoretic structure. The current formulation integrates the 
rigid-body configuration via coordinate derivatives in the implementation process, and replacing this with 
direct exponential integration on $\mathrm{SE}(3)$ could eliminate 
representation singularities and improve long-horizon numerical fidelity. 
Finally, the Euler-Bernoulli beam model adopted here could be replaced by 
higher-fidelity flexibility models — including Timoshenko beam theory for 
shear-deformable links or nonlinear elasticity models for large-deformation 
regimes — without undermining the screw-theoretic assembly structure, since the 
flexibility operator $\mathbf{D}^*_i$ is the primary affected component, rather than screw theoretic representation itself, which permits altering deformation twist as ${}^{i}\mathbf{q}_\xi = 
[{}^{i}\boldsymbol{\theta}_\xi^\top,\, {}^{i}\mathbf{r}_\xi^\top]^\top$ on the structural level. The explicit recovery of interaction wrenches and the linear-algebraic system structure also provide a natural foundation for subsystem-based control architectures exploiting the detailed representation of link flexibility on a per-link level, potentially enabling precise trajectory tracking and active vibration suppression in multi-link configurations.

\section*{Declarations}

\subsection*{Funding}
This work was supported by the Research Council of Finland under the project “Nonlinear PDE-model-based control of flexible manipulators” [Grant No. 355664], and by the “LCM-K2 Center for Symbiotic Mechatronics” within the framework of the Austrian COMET-K2 program.

\subsection*{Conflicts of interest}
The authors declare that they have no conflict of interest.

\subsection*{Author Contributions}
S.~Yaqubi: Conceptualization, methodology, investigation, formal analysis, funding acquisition, software, data curation, writing—original draft, and writing – review \& editing.  \\
A.~Kitzinger: Experimental work, investigation, software, data curation.\\
A.~Müller: Methodology, formal analysis, supervision, funding acquisition, writing—original draft, and writing – review \& editing.\\
H.~Gattringer: Formal analysis, supervision, funding acquisition, experimental work, investigation, software, and writing – review \& editing.\\
J.~Mattila: Supervision, funding acquisition, project administration, and writing—review \& editing.

\appendix 
\section{Kinematic Identities and Frame Derivatives} \label{app:kinematics}

The inertial time-derivative of any vector $\mathbf{r}$ expressed in the 
body-fixed frame is related to its body-fixed derivative by the transport 
theorem~\cite{Baruh1999}

\begin{align}
\frac{d}{dt}\mathbf{r} &= \dot{\mathbf{r}} + \boldsymbol{\omega}_i \times \mathbf{r} 
= \dot{\mathbf{r}} + \tilde{\boldsymbol{\omega}}_i\mathbf{r} \label{eq:14} 
\end{align}

Applying ~(\ref*{eq:14}) to the position vectors defined in 
Section~\ref{sec:2}

\begin{align}
\frac{d}{dt}\mathbf{r}_i &= \dot{\mathbf{r}}_i + \tilde{\boldsymbol{\omega}}_i\mathbf{r}_i \label{eq:15} \\
\frac{d}{dt}\mathbf{r}_{\xi_i} &= \dot{\mathbf{r}}_{\xi_i} + \tilde{\boldsymbol{\omega}}_i\mathbf{r}_{\xi_i} \label{eq:16} \\
\frac{d}{dt}\mathbf{r}_{b_i} &= \tilde{\boldsymbol{\omega}}_i\mathbf{r}_{b_i} \label{eq:17}
\end{align}

where $\dot{\mathbf{r}}_{b_i} = \mathbf{0}$ since the reference geometry is 
time-invariant in the body-fixed frame. The inertial velocity of element 
$d\Omega_i$ follows from substituting (\ref*{eq:15})--(\ref*{eq:17}) 
in~(\ref*{eq:11})

\begin{align}
\frac{d}{dt}\mathbf{r}_{ob_i} &= \dot{\mathbf{r}}_i + \dot{\mathbf{r}}_{\xi_i} 
+ \tilde{\boldsymbol{\omega}}_i\mathbf{r}_{ob_i} \label{eq:18}
\end{align}

The second inertial time-derivative is calculated by applying 
~(\ref*{eq:14}) iteratively, noting the antisymmetry of the 
skew-symmetric operator

\begin{align}
\frac{d^2}{dt^2}(\mathbf{r}_i) 
&= \frac{d}{dt}\left(\dot{\mathbf{r}}_i + \tilde{\boldsymbol{\omega}}_i\mathbf{r}_i\right) \notag \\
&= \ddot{\mathbf{r}}_i 
- \tilde{\mathbf{r}}_i\dot{\boldsymbol{\omega}}_i 
+ 2\tilde{\boldsymbol{\omega}}_i\dot{\mathbf{r}}_i 
+ \tilde{\boldsymbol{\omega}}_i^2\mathbf{r}_i \label{eq:19}
\end{align}

On this basis, the second time-derivatives of position vectors connecting 
the inertial and body-fixed frames to element $d\Omega_i$ are

\begin{align}
\frac{d^2}{dt^2}(\mathbf{r}_{i {b_i}}) 
&= \ddot{\mathbf{r}}_{\xi_i} 
- (\tilde{\mathbf{r}}_{\xi_i} + \tilde{\mathbf{r}}_{b_i})\dot{\boldsymbol{\omega}}_i \notag \\
&\quad + 2\tilde{\boldsymbol{\omega}}_i\dot{\mathbf{r}}_{\xi_i} 
+ \tilde{\boldsymbol{\omega}}_i^2(\mathbf{r}_{\xi_i} + \mathbf{r}_{b_i}) \label{eq:20} \\
\frac{d^2}{dt^2}(\mathbf{r}_{ob_i}) 
&= \ddot{\mathbf{r}}_i + \ddot{\mathbf{r}}_{\xi_i} 
- (\tilde{\mathbf{r}}_i + \tilde{\mathbf{r}}_{\xi_i} + \tilde{\mathbf{r}}_{b_i})\dot{\boldsymbol{\omega}}_i \notag \\
&\quad + 2\tilde{\boldsymbol{\omega}}_i(\dot{\mathbf{r}}_i + \dot{\mathbf{r}}_{\xi_i}) 
+ \tilde{\boldsymbol{\omega}}_i^2(\mathbf{r}_i + \mathbf{r}_{b_i} + \mathbf{r}_{\xi_i}) \notag \\
&= \ddot{\mathbf{r}}_i + \ddot{\mathbf{r}}_{\xi_i}  
- \tilde{\mathbf{r}}_{ob_i}\dot{\boldsymbol{\omega}}_i 
+ 2\tilde{\boldsymbol{\omega}}_i\dot{\mathbf{r}}_{ob_i} 
+ \tilde{\boldsymbol{\omega}}_i^2\mathbf{r}_{ob_i} \label{eq:21}
\end{align}

The body-fixed twists $\mathbf{V}_i$, $\mathbf{V}_{\xi_i}$, $\mathbf{V}_{b_i} 
\in \mathfrak{se}(3)$, consistently with ~(\ref*{eq:twist}), are assembled 
from $\boldsymbol{\omega}_i$ and the body-fixed frame expressions of the twist 
derivatives from~(\ref*{eq:15})--(\ref*{eq:17})

\begin{align}
{}^{i}\mathbf{V}_i &= 
\begin{bmatrix}
\boldsymbol{\omega}_i \\
\mathbf{v}_i
\end{bmatrix}
=
\begin{bmatrix}
\boldsymbol{\omega}_i \\
\dot{\mathbf{r}}_i + \tilde{\boldsymbol{\omega}}_i\mathbf{r}_i
\end{bmatrix}
\in \mathfrak{se}(3)
\label{eq:22} \\
{}^{i}\mathbf{V}_{\xi_i} &= 
\begin{bmatrix}
\mathbf{0} \\
\mathbf{v}_{\xi_i}
\end{bmatrix}
=
\begin{bmatrix}
\mathbf{0} \\
\dot{\mathbf{r}}_{\xi_i} + \tilde{\boldsymbol{\omega}}_i\mathbf{r}_{\xi_i}
\end{bmatrix}
\in \mathfrak{se}(3)
\label{eq:23} \\
{}^{i}\mathbf{V}_{b_i} &= 
\begin{bmatrix}
\mathbf{0} \\
\mathbf{v}_{b_i}
\end{bmatrix}
=
\begin{bmatrix}
\mathbf{0} \\
\tilde{\boldsymbol{\omega}}_i\mathbf{r}_{b_i}
\end{bmatrix}
\in \mathfrak{se}(3)
\label{eq:24}
\end{align}

The small adjoint $\mathrm{ad}_{\mathbf{V}_i} \in \mathbb{R}^{6\times6}$ is the 
matrix representation of the Lie bracket on $\mathfrak{se}(3)$, such that 
$\mathrm{ad}_{\mathbf{V}_i}\mathbf{u} = [\mathbf{V}_i, \mathbf{u}]_{\mathfrak{se}(3)}$ 
for any $\mathbf{u} \in \mathfrak{se}(3)$~\cite{murray1994}

\begin{equation}
\mathrm{ad}_{\mathbf{V}_i} =
\begin{bmatrix}
\tilde{\boldsymbol{\omega}}_i & \mathbf{0} \\
\tilde{\mathbf{v}}_i & \tilde{\boldsymbol{\omega}}_i
\end{bmatrix}
\label{eq:25}
\end{equation}

The body-fixed derivative $\dot{\mathbf{V}}_i$ required for numerical 
integration follows directly from differentiating ~(\ref*{eq:twist}) 
in the body-fixed frame

\begin{align}
\dot{\mathbf{V}}_i &= 
\begin{bmatrix}
\dot{\boldsymbol{\omega}}_i \\
\ddot{\mathbf{r}}_i - \tilde{\mathbf{r}}_i\dot{\boldsymbol{\omega}}_i 
+ \tilde{\boldsymbol{\omega}}_i\dot{\mathbf{r}}_i
\end{bmatrix}
\label{eq:Vdot_coord}
\end{align}

The inertial time-derivative of the twist, where $\dot{\mathbf{v}}_i$ denotes 
the translational row of ~(\ref*{eq:Vdot_coord}), is given by the Lie 
derivative on $\mathfrak{se}(3)$~\cite{murray1994}

\begin{align}
\frac{d\mathbf{V}_i}{dt} &= \dot{\mathbf{V}}_i + \mathrm{ad}_{\mathbf{V}_i}\mathbf{V}_i = 
\begin{bmatrix}
\dot{\boldsymbol{\omega}}_i \\
\dot{\mathbf{v}}_i
\end{bmatrix}
\label{eq:dvi} \\
\frac{d\mathbf{V}_{\xi_i}}{dt} &= \dot{\mathbf{V}}_{\xi_i} + \mathrm{ad}_{\mathbf{V}_i}\mathbf{V}_{\xi_i} = 
\begin{bmatrix}
\mathbf{0} \\
\dot{\mathbf{v}}_{\xi_i} + \tilde{\boldsymbol{\omega}}_i\mathbf{v}_{\xi_i}
\end{bmatrix}
\label{eq:dvxi} \\
\frac{d\mathbf{V}_{b_i}}{dt} &= \dot{\mathbf{V}}_{b_i} + \mathrm{ad}_{\mathbf{V}_i}\mathbf{V}_{b_i} = 
\begin{bmatrix}
\mathbf{0} \\
\dot{\mathbf{v}}_{b_i} + \tilde{\boldsymbol{\omega}}_i\mathbf{v}_{b_i}
\end{bmatrix}
\label{eq:dvb}
\end{align}

The Lie derivative relation $\frac{d\mathbf{V}_i}{dt} = 
\dot{\mathbf{V}}_i + \mathrm{ad}_{\mathbf{V}_i}\mathbf{V}_i$ is verified 
by substituting~(\ref*{eq:Vdot_coord}): the translational row of 
$\mathrm{ad}_{\mathbf{V}_i}\mathbf{V}_i$ contributes 
$\tilde{\mathbf{v}}_i\boldsymbol{\omega}_i + \tilde{\boldsymbol{\omega}}_i\mathbf{v}_i$, 
which vanishes identically since $\tilde{\mathbf{v}}_i\boldsymbol{\omega}_i = 
-\tilde{\boldsymbol{\omega}}_i\mathbf{v}_i$, leaving $\dot{\mathbf{v}}_i$ 
as the sole translational component, consistent with ~(\ref*{eq:dvi}).

\enlargethispage{2\baselineskip} 
\section{Coordinate-explicit forms of modal projection matrices}
\label{app:modal}

This appendix provides the coordinate-explicit forms of the distributed
flexibility matrix $\boldsymbol{M}_D$, angular coupling matrix
$\boldsymbol{M}_{D\Psi}$, and quadratic velocity vector $\mathbf{h}_{M_D}$
appearing in~(\ref*{eq:95}) of Section~\ref{sec:5}, suitable for direct
numerical implementation.

\newcommand{\Phieta}[1]{\widetilde{\boldsymbol{\phi}_{#1}\boldsymbol{\eta}_{#1}}}
\newcommand{\Rint}[2]{\int_{a_{#1}}^{c_{#1}} #2 \,d\xi_{#1}}

The row structure of all three quantities follows the DAE block pattern
of~(\ref*{eq:66}): for each link $i$, rows appear in the order
(i)~translational algebraic constraint, (ii)~rotational algebraic
constraint, (iii)~differential linear momentum,
(iv)~differential angular momentum, with junction rows between
adjacent links interleaving the pattern.

\begin{equation}
\setlength{\arraycolsep}{5pt} 
\boldsymbol{M}_D = \begin{bmatrix}
-\mathbf{R}_{o1}\boldsymbol{\phi}_1(a_1) & \mathbf{0} & \cdots & \mathbf{0}\\
\mathbf{0} & \cdots & \cdots & \mathbf{0}\\
\Rint{1}{\boldsymbol{\phi}_1} & \mathbf{0} & \cdots & \mathbf{0}\\
\Rint{1}{\tilde{\mathbf{r}}_{\xi_1}\boldsymbol{\phi}_1} & \mathbf{0} & \cdots & \mathbf{0}\\
\mathbf{R}_{o1}\boldsymbol{\phi}_1(c_1) & -\mathbf{R}_{o2}\boldsymbol{\phi}_2(a_2) & \mathbf{0} & \cdots\\
\vdots & & \ddots & \vdots\\
\mathbf{0} & \cdots & \mathbf{0} & \Rint{n}{\tilde{\mathbf{r}}_{\xi_n}\boldsymbol{\phi}_n}
\end{bmatrix} \label{eq:B1}
\end{equation}

\begin{equation}
\setlength{\arraycolsep}{3pt}
\boldsymbol{M}_{D\Psi} = \begin{bmatrix}
\mathbf{0} & -\mathbf{R}_{o1} & \mathbf{0} & \mathbf{0} & \cdots & \mathbf{0}\\
\mathbf{0} & \mathbf{0} & \cdots & \cdots & \cdots & \mathbf{0}\\
\mathbf{0} & -\Rint{1}{\Phieta{1}} & \mathbf{0} & \mathbf{0} & \cdots & \mathbf{0}\\
\mathbf{0} & -\Rint{1}{\tilde{\mathbf{r}}_{\xi_1}\Phieta{1}} & \mathbf{0} & \mathbf{0} & \cdots & \mathbf{0}\\
\mathbf{0} & \mathbf{R}_{o1} & \mathbf{0} & -\mathbf{R}_{o2} & \mathbf{0} & \cdots\\
\vdots & & & & \ddots & \vdots\\
\mathbf{0} & \cdots & \mathbf{0} & \mathbf{0} & -\Rint{n}{\Phieta{n}} & \mathbf{0}\\
\mathbf{0} & \cdots & \mathbf{0} & \mathbf{0} &
-\Rint{n}{\tilde{\mathbf{r}}_{\xi_n}\Phieta{n}} & \mathbf{0}
\end{bmatrix} \label{eq:B2}
\end{equation}

\begin{equation}
\mathbf{h}_{M_D} = \begin{bmatrix}
-\mathbf{R}_{o1}\tilde{\boldsymbol{\omega}}_1\boldsymbol{\phi}_1(a_1)\dot{\boldsymbol{\eta}}_1\\
\mathbf{0}\\
\tilde{\boldsymbol{\omega}}_1\Rint{1}{\boldsymbol{\phi}_1}\dot{\boldsymbol{\eta}}_1\\
\Rint{1}{\tilde{\mathbf{r}}_{\xi_1}\tilde{\boldsymbol{\omega}}_1\boldsymbol{\phi}_1}\dot{\boldsymbol{\eta}}_1\\
\mathbf{R}_{o1}\tilde{\boldsymbol{\omega}}_1\boldsymbol{\phi}_1(c_1)\dot{\boldsymbol{\eta}}_1 - \mathbf{R}_{o2}\tilde{\boldsymbol{\omega}}_2\boldsymbol{\phi}_2(a_2)\dot{\boldsymbol{\eta}}_2\\
\vdots\\
\tilde{\boldsymbol{\omega}}_n\Rint{n}{\boldsymbol{\phi}_n}\dot{\boldsymbol{\eta}}_n\\
\Rint{n}{\tilde{\mathbf{r}}_{\xi_n}\tilde{\boldsymbol{\omega}}_n\boldsymbol{\phi}_n}
\dot{\boldsymbol{\eta}}_n
\end{bmatrix} \label{eq:B3}
\end{equation}

Column $j$ of $\boldsymbol{M}_D$ corresponds to modal coordinate
$\boldsymbol{\eta}_j$ of link $j$, with nonzero entries only in the
rows associated with link $j$ and its junctions to links $j-1$ and
$j+1$. Column pairs $(2j-1, 2j)$ of $\boldsymbol{M}_{D\Psi}$
correspond to $(\boldsymbol{\mathcal{W}}_{j-1}, \dot{\mathbf{Z}}_j)$
of $\boldsymbol{\Psi}_{Z_j}$; the zero columns at
$\boldsymbol{\mathcal{W}}_{j-1}$ slots confirm that wrench variables
do not enter $\mathbf{D}_s$ directly, and the zero columns at the
$\dot{\mathbf{r}}_j$ part of $\dot{\mathbf{Z}}_j$ confirm that only
$\dot{\boldsymbol{\omega}}_j$ contributes the angular coupling term
$-\widetilde{\boldsymbol{\phi}_j\boldsymbol{\eta}_j}\,\dot{\boldsymbol{\omega}}_j$
arising from the twist linear velocity definition~(\ref*{eq:twist}).
The entries of $\mathbf{h}_{M_D}$ are quadratic in the rates
$(\boldsymbol{\omega}_i, \dot{\boldsymbol{\eta}}_i)$ and vanish
identically when all flexible deformation rates $\dot{\boldsymbol{\eta}}_i
= \mathbf{0}$, consistent with the rigid-body limit.

\bibliographystyle{unsrt} 
\bibliography{references}

\end{document}